%% file: NAACL2021.tex
\documentclass[11pt,a4paper]{article}
\usepackage[hyperref]{emnlp2020}
\usepackage{times}
\usepackage{latexsym}
\usepackage{booktabs}
\usepackage{multirow, multicol}
\usepackage{adjustbox}
\usepackage{graphicx}
\usepackage{pdflscape}
\usepackage{tipa} 

\usepackage{lscape}

\usepackage{url}

\graphicspath{{./images/}}

\def\Snospace~{\S{}}

\aclfinalcopy 

\title{ANLIzing the Adversarial Natural Language Inference Dataset}

\author{Adina Williams, Tristan Thrush, Douwe Kiela\\
 Facebook AI Research\\
  \texttt{\{adinawilliams, tthrush, dkiela\}@fb.com}}

\date{}

\begin{document}
\maketitle
\begin{abstract}

  We perform an in-depth error analysis of Adversarial NLI (ANLI), a recently introduced large-scale human-and-model-in-the-loop natural language inference dataset collected over multiple rounds. We propose a fine-grained  annotation scheme of the different aspects of inference that are responsible for the gold classification labels, and use it to hand-code all three of the ANLI development sets. 
  We use these annotations to answer a variety of interesting questions: which inference types are most common, which models have the highest performance on each reasoning type, and which types are the most challenging for state-of-the-art models? 
  We hope that our annotations will enable more fine-grained evaluation of models trained on ANLI, provide us with a deeper understanding of where models fail and succeed, and help us determine how to train better models in future. 
\end{abstract}

\section{Introduction}

Natural Language Inference (NLI) is one of the canonical benchmark tasks for research on Natural Language Understanding (NLU). NLI (also known as recognizing textual entailment; \citealt{dagan2006}) has several characteristics that are desirable from both practical and theoretical standpoints. Practically, NLI is easy to evaluate and intuitive even to non-linguists, enabling data to be collected at scale. Theoretically, entailment is, in the words of Richard Montague, ``the basic aim of semantics'' \citep[p. 223 fn.]{montague1970universal}, and indeed the whole notion of meaning in formal semantics is constructed following necessary and sufficient truth conditions, i.e., bidirectional entailment (``P'' \emph{if and only if} P). Hence, NLI is seen as a good proxy for measuring the overall NLU capabilities of NLP models.

Benchmark datasets are essential for driving progress in Artificial Intelligence, and in recent years, large-scale NLI benchmarks like SNLI~\citep{bowman-etal-2015-large} and MultiNLI~\citep{williams-etal-2018-broad} have played a crucial role in enabling 
a straightforward basis for comparison between trained models.
However, with the advent of huge transformer models, these benchmarks are now reaching saturation, which leads to the obvious question: have we solved NLI and, perhaps, NLU? Anyone working in the field will know that we are still far away from having models that can perform NLI in a  robust, generalizable, and dataset-independent way.
The recently collected ANLI~\citep[][]{nie-etal-2020-adversarial} dataset illustrated this by adversarially collecting difficult examples where current state-of-the-art models fail. Recently, \citet{brown2020gpt3} found that GPT-3 performs not much above chance on ANLI, noting that~``NLI is still a very difficult task for language models and [it is] only just beginning to show signs of progress'' \citep[p.20]{brown2020gpt3}. This raises the following question: where are we still falling short?

Crucially, if we want to improve towards general NLU, examples of failure cases alone are not sufficient. We also need a finer-grained understanding of \emph{which phenomena are responsible} for a model's failure or success. Since the adversarial set up of ANLI encouraged human annotators to exercise their creative faculties, the data contains a wide range of possible inferences (as we show). Because of this, ANLI is ideal for studying how current models fall short, and for characterizing what future models will have to do in order to make progress.

\input{tables/main_example_table}

Towards that end, we propose a domain-agnostic annotation scheme for NLI that breaks example pairs down into 40 
reasoning types. Our scheme is hierarchical, reaching a maximum of four layers deep, which makes it possible to analyze the dataset at a flexible level of granularity. 
A single linguist expert annotator hand-annotated all three rounds of the ANLI development set (3200 sentence pairs) according to our scheme. Another expert annotator hand-annotated a subset of the data to provide inter-annotator agreement---despite the difficulty of our annotation task, we found it to be relatively high.

This paper contributes an annotation of the ANLI development sets, not only in their entirety, but also by round, to uncover difficult inference types, and by genre, to uncover domain differences in the data. We also compare and contrast the performance of a variety of models, including the three original models used for the collection of ANLI 
and several other state-of-the art transformer architectures,
on the annotated ANLI dataset.
The annotations will be made available to the public, and we hope that they will be useful in the future, not only for benchmarking progress on different types of inference, but also to deepen our understanding of the current weaknesses of large transformer models trained to perform NLI.

\input{tables/main_ontology_table}

\section{Background}

This work proposes an inference type annotation scheme for the Adversarial NLI (ANLI) dataset. ANLI was collected via a gamified human-and-model-in-the-loop format with dynamic adversarial data collection. This means that human annotators were exposed to a ``target model'' trained on existing NLI data, and tasked with finding examples that fooled the model into predicting the wrong label. The ANLI data collection format mirrors that of SNLI and MultiNLI: na\"{\i}ve crowdworkers are given a context---and one of three classification labels, i.e., Entailment, Neutral and Contradiction---and asked to provide a hypothesis. \autoref{tab:examples} provides examples from the ANLI dataset.

The ANLI dataset was collected over multiple rounds, with different target model adversaries each round. The first round adversary was a BERT-Large~\cite{devlin2018bert} model trained on SNLI and MultiNLI. The second was a RoBERTa-Large~\cite{liu2019roberta} ensemble trained on SNLI and MultiNLI, as well as FEVER~\citep{thorne-etal-2018-fever} and the training data from the first round. The third round adversary was a RoBERTa-Large ensemble trained on all of the previous data plus the training data from the second round, with the additional difference that the contexts were sourced from multiple domains~(rather than just from Wikipedia, as in the preceding two rounds).

One hope of the ANLI dataset creators was that crowdworkers who participated in their gamified data collection effort gave free rein to their creativity \citep[p.8]{nie-etal-2020-adversarial}.\footnote{Gamification is known to generally result in datasets containing a wide variety of possible patterns \citep{joubert2018jeuxdemots, bernardy2019kind}.}~As rounds progress, ANLI annotators will attempt to explore this full range, then ultimately converge on reasoning types that are especially difficult for particular model adversaries. For example, if the target model in round 1 was susceptible to being fooled by numerical examples (which seems to be the case, see below \autoref{sec:experiments}), then the data from that round will end up containing a reasonably large amount of example pairs containing \textsc{Numerical} reasoning. If the adversary for later rounds is trained on that round 1 data (i.e., A1), it should improve on \textsc{Numerical} examples. For the next round then, crowdworkers would be less successful in employing \textsc{Numerical} examples to stump models trained on A1, and fewer example pairs containing that type of reasoning would make it in to the development sets for the later rounds.\footnote{Assuming that models trained on later rounds don't suffer from catastrophic forgetting.} In this way, understanding which types of reasoning are present in each of the rounds of ANLI gives us a window into the abilities of the target models used to collect them.

\section{A Scheme for Annotating Natural Language Inference Relation Types}\label{sec:scheme}

While isolating types of sentential relations is by no means a new endeavor (see e.g., Aristotle's doctrine of categories), 
the construction of a scheme should be, to some extent, sensitive to the particular task at hand. In this work, we propose an novel annotation scheme specific to the task of NLI. NLI's prominence as an NLU task has led to a variety of in-depth studies on the performance of NLI models and issues with existing NLI datasets, primarily focused on annotation artifacts ~\cite{gururangan2018,Geiger2018stress,poliak-etal-2018-hypothesis,Tsuchiya2018performance,Glockner2018breaking,Geva2019taskorannotator} and diagnostic datasets~\cite{mccoy2019right,naik-etal-2018-stress,nie2019,yanaka-etal-2019-help,warstadt-etal-2019-investigating,jeretic2020natural,warstadt-etal-2020-blimp}; see~\newcite{zhou2020curseofnlidatasets} for a critical examination. There has also been work in probing NLI models to see what they learn~\citep{richardson2019probing}, as well as specifically on the collection of NLI annotations~\citep{bowman2020collecting} and analyzing inherent disagreements between human annotators~\citep{pavlick-kwiatkowski-2019}. Taking inspiration from these works, as well as \citet{fracas}, \citet{sammons-etal-2010-ask}, \citet{lobue-yates-2011-types}, \citet{jurgens-etal-2012-semeval}, \citet{jia-liang-2017-adversarial}, \citet{white-etal-2017-inference}, \citet{naik-etal-2018-stress} and others, our goal here is to create a flexible and hierarchical annotation scheme specifically for NLI.

\input{tables/main_incidence_table}

Our scheme, provided in \autoref{tab:ontolog}, has 40 different tag types that can be combined to a depth of up to four. 
See \autoref{tab:fullontolog} in the Appendix for dataset examples for every type. The scheme 
was developed in response to reading random samples of the development set of ANLI Round 1. The top layer of the scheme was fixed by the original ANLI paper to five classes: \textsc{Numerical}, \textsc{Basic}, \textsc{Reference}, \textsc{Tricky} inferences, and \textsc{Reasoning}.\footnote{These top-level types were introduced for smaller subsets of the ANLI development set in \S~5 of \newcite{nie-etal-2020-adversarial}, 
which we drastically expand both in number and specificity of tag types, as well as in the scope of annotation.}

In solidifying our scheme, we necessarily walk a thin line between proliferating overly specific tags (and potentially being overly expressive), and limiting the number of tags to enable generalization (potentially not being expressive enough). A hierarchical tagset allows us to get the best of both worlds---since we can measure all our metrics both at a vague level and then more specifically as well---all while allowing for pairs to receive as many tags as are warranted~(see \autoref{tab:examples}). 

One unique contribution of our work is that our examples are \emph{only} tagged as belonging to a particular branch of the taxonomy when the tagged phenomenon  contributes to the target label assignment. Others label the \emph{presence} of linguistic phenomena in the sentences in either an automatic fashion, on in a way that is fairly easy for naive annotators to learn to perform. Since our annotations highlight only the phenomena present in each sentence pair that a human would (have to) use to perform NLI, automation is very difficult, making expert annotators crucial. We hope that our scheme will be for annotating other large NLI datasets to make even wider comparisons possible. Please see \autoref{tab:schemecompare} and \autoref{app:directcomparisons} for pairwise comparisons between our annotation scheme and several other popular existing semantic annotations schemes from which we drew inspiration. 

\paragraph{The Tags.}\textsc{Numerical} classes refer to examples where numerical reasoning is crucial for determining the correct label, and break down into \textsc{Cardinal}, \textsc{Ordinal}---along the lines of \citet{ravichander-etal-2019-equate}---\textsc{Counting} and \textsc{Nominal}; the first two break down further into \textsc{Ages} and \textsc{Dates} if they contain information about either of these topics. \textsc{Basic} consists of staple types of reasoning, such as lexical hyponymy and hypernymy (see also \citealt{glockner-etal-2018-breaking}), conjunction (see also \citealt{toledo2012semantic,saha2020conjnli}), and negation. \textsc{Reference} consists of pairs that require noun or event references to be resolved (either within or between context and hypothesis examples). \textsc{Tricky} examples require either complex linguistic knowledge, say of pragmatics or syntactic verb argument structure and reorderings, or word games. \textsc{Reasoning} examples require the application of reasoning outside of what is provided in the pair alone; it is divided into three levels. The first is \textsc{Plausibility}, which was loosely inspired by \newcite{AbductiveCommonSenseReasoning, chen-etal-2020-uncertain}, for which the annotator provided their subjective intuition on how likely the situation is to have genuinely occurred (for example `when computer games come out they are often buggy' and `lead actors get paid the most' are likely). The other two \textsc{Facts} and \textsc{Containment} refer to external facts about the world (e.g., `what year is it now?') and relationships between things (e.g., `Australia is in the southern hemisphere'), respectively, that were not clearly provided by the example pair itself.

There is also a catch-all class labeled \textsc{Imperfection} that catches not only label ``errors'' (i.e., rare cases of labels for which the expert annotator disagreed with the gold label from the crowdworker-annotator), but also  spelling errors (\textsc{Spelling}), event coreference examples\footnote{SNLI and MultiNLI annotation guidelines required annotators to assume that the premise and hypothesis refer to a single thing (i.e., entity or event). According to their guidelines, `a cat is sleeping on the bed' and `a dog is sleeping on the bed' should be a contradiction, because both sentences cannot describe one and the same animal (see \citealt[p.78--80]{bowman2016thesis}). The \textsc{Event Coreference} tag is for when annotators didn't make that assumption.}, foreign language content (\textsc{Translation}), and pairs that could reasonably be given multiple correct labels (\textsc{Ambiguous}). The latter are likely uniquely subject to human variation in entailment labels, \textit{\`a la} \citet{pavlick-kwiatkowski-2019}, \citet{ min2020ambigqa}, \citet{nie2020can}, since people might vary on which label they initially prefer, even though multiple labels might be possible.


\paragraph{Annotation.} Annotating NLI data for reasoning types requires various kinds of expert knowledge. One must not only be familiar with a range of complicated linguistic phenomena, such as pragmatic reasoning and syntactic argument structure, but also have knowledge of the particularities of task formats and dataset collection decisions (e.g., 2-\ vs.\ 3-way textual inference). Often, trained expert annotators achieve higher performance on linguistically sophisticated tasks than na\"{i}ve crowdworkers, e.g., for the CoLA subtask \citep{warstadt2018corpus} of the GLUE benchmark \citep[p. 4569]{nangia-bowman-2019-human}. 
This suggests that one ideally wants expert annotators for difficult annotation tasks like this one (see also \citealt{basile-etal-2012-developing,bos2017groningen} for other NLU annotation projects that benefit from experts). Because of this, we chose to rely on a single annotator with a decade's expertise in NLI and linguistics to both devise our scheme and to apply it to annotating the ANLI development set. 

Annotation was a laborious process. It took the expert on the order of several hundred hours. To our knowledge, our expert hand-annotation of the 3200 textual entailment sentences in the ANLI development set constitutes one of the largest single expert annotation projects for a complex NLU task, approximating the number of annotations on all five rounds of RTE \citep{dagan2006}, and exceeding other NLU expert annotation efforts (e.g., \citealt{snow-etal-2008-cheap,toledo2012semantic, mirkin-etal-2018-listening,raghavan2018annotating}) in total number of expert annotated pairs.

\paragraph{Inter-annotator Agreement.} Employing a single annotator may have downsides, since they could inadvertently introduce personal idiosyncrasies into their annotations. Recent work indicates that there is substantial variation in human judgements for NLI \citep{pavlick-kwiatkowski-2019, min2020ambigqa, nie2020can}. Given that our annotation task is also likely more difficult than NLI, we were especially keen to determine inter-annotator agreement. To understand the extent to which our tags that are very individual to the main annotator, we employed a second expert annotator (with 5 years of linguistic training) to annotate a subportion of the development datasets. We randomly selected 
200 examples across the three development sets for the second expert annotator to provide tags for. This task took the second annotator roughly 20 hours (excluding training time). Further details on the scheme, annotation guidelines, and our annotation process are provided in \autoref{app:furtherdetails}.

We measure inter-annotator agreement across these examples for each tag independently. For each example, annotators are said to agree on a tag if they both used that tag or both did not use that tag; they are said to disagree otherwise. We report average percent agreement here (but see \autoref{app:IAA} for further details on agreement). 

Average percent agreement between our annotators is 92\% and 75\%, for top-level and lower-level tags respectively. Recall that 50\% would be chance (since we are measuring whether the tag was used or not and comparing between our two annotators). Our inter-annotator agreement is comparable to a similar semantic annotation effort on top of the original RTE data \citep{toledo2012semantic}, suggesting we have reached an acceptable level of agreement for our setting. To provide additional NLI-internal context for these results, percent agreement on both top and lower level tags exceeds the percent agreement of non-experts on the task of NLI as reported in \citet{bowman-etal-2015-large} and \citet{williams-etal-2018-broad}. Since our annotation scheme incorporated some subjectivity---i.e., to fully but subjectively annotate as many phenomena as you think contribute to the label decision---annotators are likely to have different blindspots. For this reason, we will release the union of the two annotators tags, for examples where that is available.

\section{Experiments}\label{sec:experiments}

We conduct a variety of experiments using our annotations. The goal of these experiments is two-fold: to shed light on the dataset and existing methods used on it, as well as to illustrate how the annotations extend the usefulness of the dataset by making it possible to analyze future model performance with more granularity.

\subsection{Tag Distribution}

In this section, we ask whether the incidence of tags in the ANLI development sets differ by rounds and gold label. The results for top-level tags are presented in \autoref{tab:incidence}, while those for lower-level tags are presented in the Appendix in \autoref{tab:fullincidence}. \textsc{Reasoning} tags are the most common in the dataset, followed by \textsc{Numerical}, \textsc{Tricky}, \textsc{Basic} and \textsc{Reference} and then \textsc{Imperfections}.

We find that \textsc{Numerical} pairs appear at the highest rate in A1, which makes sense since it was collected using the first few lines of Wikipedia entries---which often have numbers in them%
---as contexts. A2, despite also using Wikipedia contexts, has a lower percentage of \textsc{Numerical} examples, possibly because its target model---also trained on A1---improved on that category. In A3, the percentage of \textsc{Numerical} pairs has dropped even lower.
Between A1/A2 and A3, the drop in top level \textsc{Numerical} tags is accompanied by a drop in the use of second level \textsc{Cardinal} tags, which results in a corresponding drop of third level \textsc{Dates} and \textsc{Ages} tags as well (in the Appendix). 
Overall, \textsc{Numerical} pairs are more likely to have the gold label contradiction or entailment than neutral.

\textsc{Basic} pairs are relatively common, with increasing rates as rounds progress.  Second level tags \textsc{Lexical} and \textsc{Negation} rise sharply in incidence between A1 and A3, \textsc{Implications} and \textsc{Idioms} also rise in incidence---though they rise less sharply and are only present in trace levels (i.e., $<10\%$ of examples)---and the incidence of \textsc{Coordinations} and \textsc{Comparatives \& Superlatives} stays roughly constant. Overall, \textsc{Basic} examples tend to be gold labeled as entailment more often than as contradiction or neutral.

\textsc{Reference} tags are the least prevalent main tag type, with the lowest incidence of 24.5\% in A1 rising to the upper 20s in A2 and A3. The most common second level tag for \textsc{Reference} is \textsc{Coreference} with incidences ranging from roughly 16\% in A1 to 26\% in A3. Second level tags \textsc{Names} and \textsc{Family} maintain roughly constant low levels of incidence across all rounds (although there is a precipitous drop in \textsc{Names} tags for A3, likely reflecting genre differences).  Examples tagged as \textsc{Reference} are more commonly entailment examples across all rounds.

\textsc{Tricky} reasoning types occur at relatively constant rates across rounds. A1 contains more examples with syntactic reorderings than the others. For both A1 and A3, \textsc{Pragmatic} examples are more prevalent. A2 is unique in having slightly higher incidence of \textsc{Exhaustification} tags, and \textsc{Wordplay} examples increase in A2 and A3 compared to A1. On the whole, there are fewer neutral \textsc{Tricky} pairs than contradictions or entailments, with contradiction being more common than entailment.

\textsc{Reasoning} examples are very common across the rounds, with 50--60\% of pairs receiving at least one. Second level \textsc{Facts} pairs are also common, rising from 19\% in A1 to roughly $1/4$ of A2 and A3 examples; \textsc{Containment} shows the opposite pattern, and halves its incidence between A1 and A3.  The incidence of third level \textsc{Likely} examples remains roughly constant whereas third level \textsc{Unlikely} and \textsc{Debatable} examples become more common over the rounds. In particular, \textsc{Debatable} tags rise to 3 times their rate in A3 as in A1, perhaps reflecting the contribution of different domains of text. On average, \textsc{Reasoning} tags are more common for examples with neutral as the gold label. 

\textsc{Imperfection} tags are rare across rounds ($\approx 14$\% of example pairs receive that tag on average), and are slightly more common for neutral pairs. \textsc{Spelling} imperfections are the most common second level tag type, at approximately $\approx 5--6$\% of examples, followed by examples marked as \textsc{Ambiguous} and \textsc{Translation} and \textsc{Error}, which were each at $\approx 3$\%. There were very few examples of \textsc{Event Coreference} ($\approx 2\%$). 

\begin{figure}[t]
    \centering
    \includegraphics[width=\columnwidth]{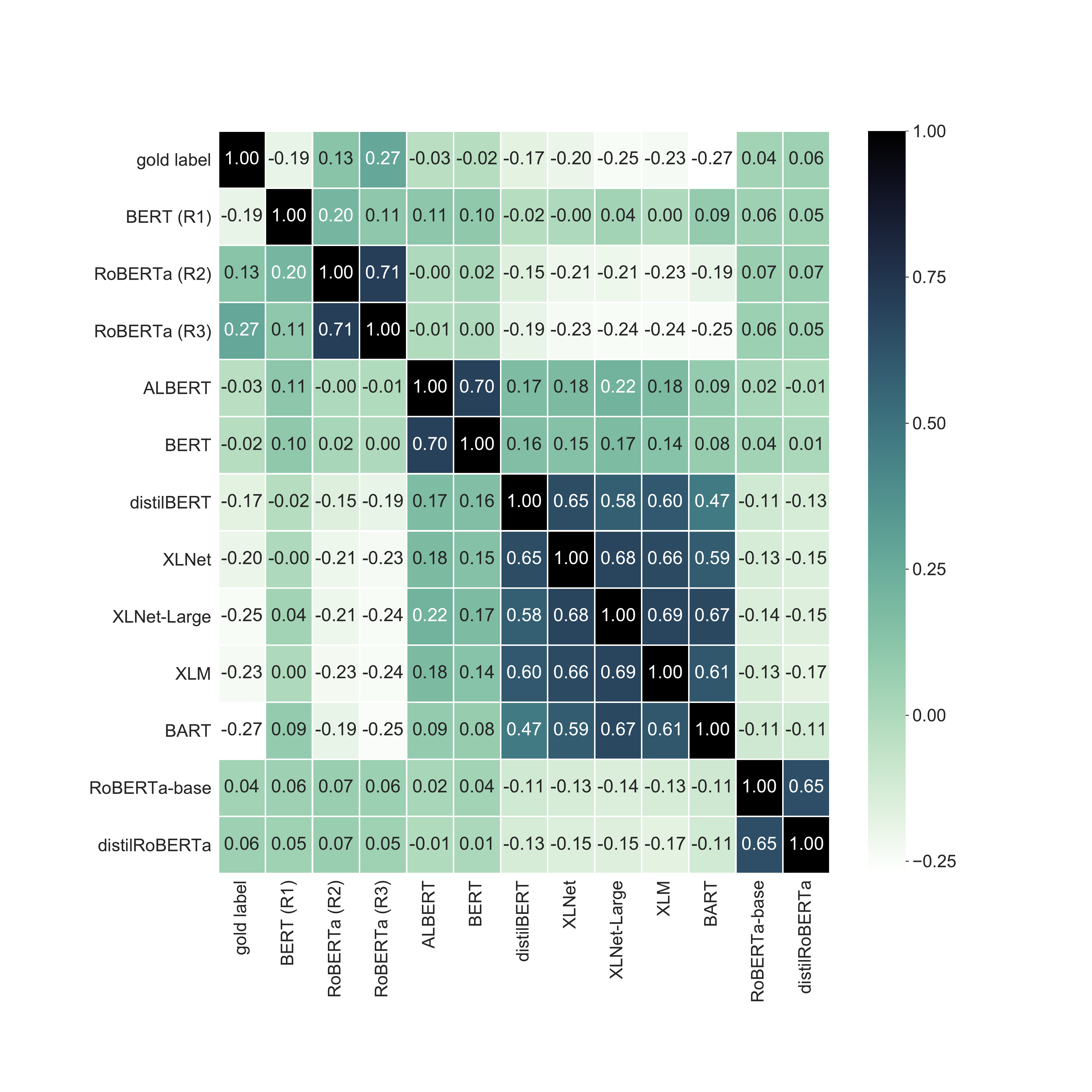}
    \caption{Pearson's correlation between all models and the gold labels. $p$-values are in \autoref{tab:heatmapPearsonP}.}
    \label{fig:overlap}
\end{figure}

\subsection{Analyzing Model Predictions}

\input{tables/main_model_preds_table}

We compare the performance of a variety of transformer models, trained on a combination of datasets, namely SNLI~\citep{bowman-etal-2015-large}, MultiNLI~\citep{williams-etal-2018-broad}, FEVER~\citep{thorne-etal-2018-fever}, and all the rounds of ANLI. Specifically, we include two RoBERTa-type models---RoBERTa-base~\cite{liu2019roberta} and a distilled version called DistilRoBERTa~\cite{sanh2019distilbert}---three BERT-type models---BERT-base~\cite{devlin2018bert} (uncased), ALBERT-base~\cite{lan2019albert} and DistilBert~\cite{sanh2019distilbert} (uncased)---two XLNet models~(\citealt{yang2019xlnet}; base and large, both cased), XLM~\cite{XLM}, and BART-Large~\cite{BART}.
We also include a comparison to the original ANLI target models. For A2 and A3, which were ensembles, we randomly select a single RoBERTa-Large model as the representative.



We examine how much these models overlap by measuring the pairwise Pearson's correlations between their predictions.
\autoref{fig:overlap} presents the result. We enable comparison with the gold labels by representing the predictions as one-hot vectors.
We find that the RoBERTa model used to collect round 3 of ANLI has the highest correlation with the gold labels from the full ANLI dataset. Positive correlations with the gold label are also found for the other three RoBERTa-type architectures.

\paragraph{Different Models Make Similar Mistakes.} We find that the predictions of many of the architectures are correlated. For example, distilBERT, XLNet-base, XLNet-Large, XLM, and BART are all significantly correlated with each other with Pearson's coefficients exceeding $0.50$; while having low negative correlations with the RoBERTa-type models, and even lower scores for the gold labels. This suggests that these models are performing comparably to each other, but perform poorly overall. We also see that ALBERT and BERT-base are highly correlated, with a Pearson's coefficient of $0.70$, as one might expect. We find that predictions from the RoBERTa-Large models used to collect A2 and A3 are correlated, with a Pearson's coefficient of $0.70$. Finally, RoBERTa-base and distilRoBERTa are also highly correlated with a Pearson's coefficient of $0.65$, while the correlation between BERT and distilBERT is much lower.

\subsection{Model Predictions by Tag}\label{subsec:tags}

Given these pairwise model correlations, we next analyze the correct label probability and entropy of predictions for an informative subset of the models in \autoref{tab:analysis}, while providing these metrics for the remainder of the models in \autoref{appendix:modelpreds} in  \autoref{tab:modelpredsbasic}--\autoref{tab:modelpredstricky}. We find that Large models do better (as measured by higher correct label probability and lower entropy) than their base counterparts on the entirety of the ANLI development set. On the whole, RoBERTa models do better than BERT, echoing their positive correlation with gold labels reported above. XLNet models, XLM, and BART have the lowest performance. ALBERT and BERT are very similar, with distilBERT being markedly worse than either (having both low probabilities and low entropies, suggesting relatively high certainty on incorrect answers). On the other hand. RoBERTa and distilRoBERTa are close, with a much smaller performance drop from RoBERTa to distilRoBERTa than from BERT or ALBERT to distilBERT.

The best-performing model overall is RoBERTa-Large from round 3; it has the highest label probability for the full ANLI development set, coupled with the lowest entropy for each tag type. RoBERTa-Large (R2) is a somewhat distant second. On A1 and A2, RoBERTa-Large (R3) also has the highest average label probability and lowest entropy (except for A2, Imperfections, where distilRoBERTa-Base exceeds its prediction probability, suggesting a higher tolerance for noise). For A3, RoBERTa-Large (R3) was one of the models in the ensemble, meaning that its average prediction probability on this round should be low. With RoBERTa-Large (R3) out of the running, there is no clear winner for A3, although disilRoBERTa-Base, and BERT-Large~(R1) come in as possible contenders.

\paragraph{Analyzing Tag Difficulty.}

The hardest overall category for all models appears to be \textsc{Reference} with \textsc{Tricky} and \textsc{Reasoning} being the next most difficult categories, which indicates that models struggle with the fact that ANLI often requires external knowledge. RoBERTa-type models beat BERT-type models especially on \textsc{Numerical} and \textsc{Tricky} examples. RoBERTa-Large (R3) has the most trouble with \textsc{Imperfections} out of all tag types, suggesting that spelling errors and other sources of noise do affect its performance to some extent; BERT-base seems to have reasonable resilience to \textsc{Imperfections}, at least for A2 and A3, as do RoBERTa-Base and distilRoBERTa-base. This suggests that model robustness plays a part in the examples that annotators learn to exploit and it might make sense to use more diverse sets of target model ensembles to increase robustness.

\input{tables/main_model_genre}

We analyze the performance of the RoBERTa-Large model used to collect A3 on the hardest tag \textsc{Reasoning}, by decomposing performance on second and third level tags. \textsc{Reasoning} examples with the \textsc{Likely} tag are easiest for the model, followed by \textsc{Containment}, for example that New York  City is on the East Coast of the United States, suggesting that some amount of world knowledge has been acquired. On the other hand, \textsc{Facts} examples are usually more difficult than \textsc{Likely}, \textsc{Containment} and \textsc{Unlikely}, perhaps because they are predicated on knowledge not included in the context or hypothesis (e.g., that this year is the year 2020). Finally, \textsc{Debatable} examples---recall that these examples often contain opinions---are the most difficult ones under the \textsc{Reasoning} top-level tag. Further description of the results for all second and third level tags is provided in \autoref{appendix:modelpreds}.


\subsection{Multi-Domain NLI}

ANLI Round 3 was collected using contexts from a variety of domains. \autoref{tab:genre-r3probs} shows the performance of the RoBERTa-Large model from Round~3 on the different genres. Wikipedia is the least difficult genre (as well as the most frequent in the overall dataset), and the others are about equally difficult (with News being somewhat harder, yet also lower entropy, than RTE, Legal, Procedural and Fiction genres). Genres differ widely in how many of their examples have particular top-level tags. \autoref{tab:genreincidence} shows a breakdown of the tags by genre. Across all genres, \textsc{Tricky} and \textsc{Reasoning} examples occur at roughly the same rates---with \textsc{Reasoning} examples being very common than all other reasoning types across the board. A much higher proportion of News genre examples have \textsc{Basic} tags and Wikipedia has a much higher rate of \textsc{Numerical} tags when compared to the other genres. Procedural text has the lowest  rate of \textsc{Numerical} and \textsc{Reference} examples, but the highest rate of \textsc{Imperfection}.  Taken together, these results suggest that the text domains are very different, and that domain likely has a large impact on how we should understand our results.
 
\autoref{tab:genre-breakdown} shows a breakdown of the performance of the RoBERTa-Large model from Round~3, broken down by tags and genre (see \autoref{tab:modelpredsgenreall} in the Appendix for the other models' performances across genres). The RoBERTa-Large~(R3) model does best on all tags in Wikipedia data (the genre that a large part of its training data came from). It does somewhat better on \textsc{Numerical} examples from the Fiction and  Procedural genres, on \textsc{Reference} examples from the Fiction genre, and \textsc{Tricky} and \textsc{Imperfection} examples from the Procedural genre. This suggests that data from different genres could be differentially beneficial for training the skills needed for these top-level tags, suggesting that targeted upsampling could be beneficial in the future. 

\subsection{Other Analyses}

We also provide a detailed analysis of other word and sentence-level dataset properties (such as word and sentence length, most common words by round, gold label, and tag), available in \autoref{appendix:properties}, where we find that ANLI and MultiNLI are relatively similar, with SNLI having a rather different distribution. We also investigate the annotator-provided rationales more closely in \autoref{appendix:rationaleproperties}, \autoref{tab:words_round_gold}--\autoref{tab:words_top_annot_tags}.

\section{Conclusion}

We annotated the development set of the ANLI dataset \citep{nie-etal-2020-adversarial} according to a 
hierarchical reasoning scheme to determine which types of reasoning are responsible for model success and failure. We find that the percentage of examples with a given tag increases as ANLI rounds increase for most tags, and that inferences relying on common sense reasoning and numerical reasoning are the most prevalent, appearing in roughly 40\%--60\% of dataset examples respectively. 
We trained a variety of NLI models and compared their performance to the original target models used to adversarially collect the dataset. We find that RoBERTa-type models currently perform the best of our sample, but there is still a lot of room for improvement on every type.  When we compare types of reasoning, we find that examples requiring common sense reasoning, understanding of entity coreference, and linguistic knowledge are the most difficult for our models across the board.

ANLI was recently found to be extremely difficult for the huge 175B parameter GPT-3 model, suggesting that it may require radically new ideas. We hope that our annotations will enable a more fine-grained understanding of model strengths and weaknesses as ANLI matures and the field makes advances towards the end goal of natural language understanding.

\section*{Acknowledgements}

Special thanks to Naman Goyal for converting the R2 and R3 RoBERTa models to something modern fairseq can load. Thanks as well to Yixin Nie, Mohit Bansal, Emily Dinan, and Grusha Prasad for relevant discussion relating to ANLI and analysis for NLI in general.

\bibliography{acl2019, anthology}
\bibliographystyle{acl_natbib}

\appendix

\newpage \clearpage 

\input{appendix.tex} 

\end{document}

%% file: tables/main_example_table.tex
\begin{table*}
\tiny
	\begin{tabular}{p{28em}p{9em}p{17em}p{5em}p{6em}}
		\toprule
        \bf Context & \bf Hypothesis & \bf Rationale  & \bf Gold/Pred. (Valid.) & \bf Tags \\ \midrule
		Eduard Schulte (4 January 1891 in D\"usseldorf – 6 January 1966 in Z\"urich) was a prominent German industrialist. He was one of the first to warn the Allies and tell the world of the Holocaust and systematic exterminations of Jews in Nazi Germany occupied Europe. & Eduard Schulte is the only person to warn the Allies of the atrocities of the Nazis. & The context states that he is not the only person to warn the Allies about the atrocities committed by the Nazis. & C/N (CC) & Tricky, Prag., Numerical, Ordinal\\ \midrule
		Kota Ramakrishna Karanth (born May 1, 1894) was an Indian lawyer and politician who served as the Minister of Land Revenue for the Madras Presidency from March 1, 1946 to March 23, 1947. He was the elder brother of noted Kannada novelist K. Shivarama Karanth. &  Kota Ramakrishna Karanth has a brother who was a novelist and a politician & Although Kota Ramakrishna Karanth's brother is a novelist, we do not know if the brother is also a politician & N/E (NEN) & Basic, Coord., Reasoning, Plaus., Likely, Tricky, Syntactic\\  \midrule
	    Toolbox Murders is a 2004 horror film directed by Tobe Hooper, and written by Jace Anderson and Adam Gierasch. It is a remake of the 1978 film of the same name and was produced by the same people behind the original. The film centralizes on the occupants of an apartment who are stalked and murdered by a masked killer. & Toolbox Murders is both 41 years old and 15 years old. & Both films are named Toolbox Murders one was made in 1978, one in 2004. Since it is 2019 that would make the first 41 years old and the remake 15 years old.  & E/C (EE) & Reasoning, Facts, Numerical Cardinal, Age, Tricky, Wordplay\\ 
		\bottomrule
	\end{tabular}
	\caption{\label{tab:examples} Examples from development set. `corr.' is the original annotator's gold label, `pred.' is the model prediction, `valid.' is the validator label(s).
	} 
\end{table*}

%% file: tables/main_ontology_table.tex
\begin{table*}[t]
    \centering
    \tiny
    \begin{tabular}{ccp{57em}}
    \toprule
 \bf Top Level & \bf Second Level & \bf Description \\\midrule 
\multirow{4}{*}{Numeral} & Cardinal & basic cardinal numerals (e.g., 56, 57, 0, 952, etc.). \\
& Ordinal & basic ordinal numerals (e.g., 1$^{st}$, 4$^{th}$, 72$^{nd}$ etc.). \\
& Counting & counting references in the text, such as: \textit{Besides A and B, C is one of the monasteries located at Mt. Olympus.} $\Rightarrow$ \textit{C is one of three monasteries on Mount Olympus.}\\
& Nominal & numbers as names, such as: \textit{Player 37 scored the goal} $\Rightarrow$ \textit{a player was assigned jersey number 37}.\\ \midrule
\multirow{5}{*}{Basic} & Comp.\& Super. & degree expressions denoting relationships between things, such as: \textit{if X is faster than Y} $\Rightarrow$ \textit{Y is slower than X}  \\
& Implications & cause and effect, or logical conclusions that can be drawn from clear premises. Includes classical logic types such as Modus Ponens. \\
& Idioms & idioms or opaque multiword expressions, such as: \textit{Team A was losing but managed to beat the other team}  $\Rightarrow$ \textit{Team A rose to the occasion}  \\
& Negation & inferences relying on negating content from the context, with ``no'', ``not', ``never'', ``un-'' or other linguistic methods \\
& Coordinations &  inferences relying on ``and'', ``or'', ``but'', or other coordinating conjunctions. \\ \midrule
\multirow{3}{*}{Ref.} & Coref. & accurately establishing multiple references to the same entity, often across sentences, such as: \textit{Sammy Gutierrez} is \textit{Guty} \\
& Names & content about names in particular (e.g., \textit{Ralph is a male name}, \textit{Fido is a dog's name}, \textit{companies go by acronyms})\\
& Family & content that is about families or kinship relations (e.g., \textit{if X is Y’s aunt, then Y is X’s nephew/niece and Y is X's parent's sibling})
\\\midrule
\multirow{4}{*}{Tricky} & Syntactic & argument structure alternations or changes in argument order (e.g., \textit{Bill bit John} $\Rightarrow$ \textit{John got bitten.}, \textit{Bill bit John} $\not \Rightarrow$ \textit{John bit Bill})
 \\
& Pragmatic & presuppositions, implicatures, and other kinds of reasoning about others’ mental states: \textit{It says `mostly positive’ so it stands to reason some were negative.}
 \\
& Exhaustification  & pragmatic reasoning where all options not made explicit are impossible, for example: \textit{a field involves X, Y, and Z} $\Rightarrow$ \textit{X, Y and Z are the only aspects of the field} \\
& Wordplay & puns, anangrams, and other fun language tricks, such as \textit{Margaret Astrid Lindholm Ogden's initials are MALO, which could be scrambled around to form the word 'loam'.} \\ \midrule
\multirow{3}{*}{Reasoning} & Plausibility & the annotators subjective impression of how plausible a described event is (e.g. \textit{Brofiscin Quarry is named so because a group of bros got together and had a kegger at it.} and \textit{Fetuses can't make software} are unlikely)\\
& Facts & common facts the average human would know (like that the year is 2020), but that the model might not (e.g., \textit{the land of koalas and kangaroos} $\Rightarrow$ \textit{Australia}), including statements that are clearly not facts (e.g., \textit{In Ireland, there's only one job.}) \\ 
& Containment & references to mereological part-whole relationships, temporal containment between entities (e.g., \textit{October is in Fall}), or physical containment between locations or entities (e.g., \textit{Germany is in Europe}). Includes examples of bridging (e.g., \textit{the car had a flat} $\Rightarrow$ \textit{The car's tire was broken}). \\ \midrule
\multirow{5}{*}{Imperfections} & Error & examples for which the expert annotator disagreed with the gold label, such as the gold label of neutral for the pair  \textit{How to limbo. Grab a long pole. Traditionally, people played limbo with a broom, but any long rod will work} $\Rightarrow$ \textit{limbo is a type of dance} \\
& Ambig. & example pairs for which multiple labels seem to the expert to be appropriate. For example, with the context \textit{Henry V is a 2012 British television film}, whether \textit{Henry V is 7 years old this year} should get a contradiction or neutral label depends on what year it is currently as well as on which month Henry V began to be broadcast and when exactly the hypothesis was written.  \\
& Spelling & examples with spelling errors.  \\
& Translation & examples with a large amount of text in a foreign language. \\
\bottomrule
\end{tabular}
    \caption{Summary of the Annotation Scheme. Toy examples are provided, $\Rightarrow$ denotes entailment, $\not \Rightarrow$ denotes contradiction. Only top and second level tags are provided, due to space considerations.
    }
    \label{tab:ontolog}
\end{table*}

%% file: tables/main_incidence_table.tex
\begin{table*}[t]
    \centering
    \small
    \begin{tabular}{clrrrrrr}
    \toprule
 \bf Dataset & \bf Subset & \bf Numerical & \bf Basic & \bf Reference & \bf Tricky & \bf Reasoning & \bf Error \\\midrule 
\multirow{4}{*}{\bf A1} 
& All & 40.8 & 31.4 & 24.5 & 29.5 & 58.4 & 3.3 \\ \cmidrule{2-8}
&   C & 18.6 & 8.2 & 7.8 & 13.7 & 11.9 & 0.7 \\ 
&	N & 7.0 & 9.8 & 7.1 & 6.4 & 31.3 & 1.0 \\ 
&	E & 15.2 & 13.4 & 9.6 & 9.4 & 15.2 & 1.6 \\ \midrule
\multirow{4}{*}{\bf A2}  
& All & 38.5 & 41.2 & 29.4 & 29.1 & 62.7 & 2.5 \\ \cmidrule{2-8}
&	C & 15.6 & 11.8 & 10.2 & 13.6 & 15.5 & 0.3 \\ 
&	N & 8.1 & 12.8 & 9.1 & 7.4 & 30.0 & 1.4 \\ 
&	E & 14.8 & 16.6 & 10.1 & 8.1 & 17.2 & 0.8 \\ \midrule
\multirow{4}{*}{\bf A3}  
& All & 20.3 & 50.2 & 27.5 & 25.6 & 63.9 & 2.2 \\ \cmidrule{2-8}
&	C & 8.7 & 17.2 & 8.6 & 12.7 & 14.9 & 0.3 \\ 
&	N & 4.9 & 13.1 & 8.2 & 4.6 & 30.1 & 1.0 \\ 
&	E & 6.7 & 19.9 & 10.7 & 8.3 & 18.9 & 0.8 \\ 
\bottomrule
\end{tabular}
    \caption{Percentages (of the total) of tags by gold label and subdataset. `All' refers to the total percentage of examples in that round that were annotated with that tag. `C', `N', and `E', refer to percentage of examples with that tag that receive each gold label.
    }
    \label{tab:incidence}
\end{table*}

%% file: tables/main_model_preds_table.tex
\begin{table*}[t]
    \centering
    \small
    \begin{adjustbox}{max width=\linewidth}
    \begin{tabular}{crccccccc}
    \toprule
    \bf Round & \bf Model & \bf Numerical & \bf Basic & \bf Ref. \& Names & \bf Tricky & \bf Reasoning & \bf Imperfections\\ \midrule
\multirow{7}{*}{\bf A1}  &                BERT-Large (R1) &  0.10 (0.57) &  0.13 (0.60) &  0.11 (0.56) &  0.10 (0.56) &  0.12 (0.59) &  0.13 (0.57) \\
 &    RoBERTa-Large (R2) &  0.68 (0.13) &  0.67 (0.13) &  0.69 (0.15) &  0.60 (0.18) &  0.66 (0.15) &  0.61 (0.14) \\
 &    RoBERTa-Large (R3) & \bf 0.72 (0.07) & \bf 0.73 (0.08) & \bf 0.72 (0.08) & \bf 0.65 (0.09) & \bf 0.70 (0.08) & \bf 0.68 (0.07) \\
 &        BERT-Base &  0.24 (0.92) &  0.39 (0.92) &  0.28 (0.88) &  0.26 (0.86) &  0.30 (0.92) &  0.30 (0.87) \\
 &  distilBERT-Base &  0.19 (0.35) &  0.21 (0.34) &  0.21 (0.31) &  0.22 (0.31) &  0.17 (0.34) &  0.24 (0.31) \\
 &             RoBERTa-Base &  0.32 (0.40) &  0.47 (0.33) &  0.31 (0.34) &  0.34 (0.40) &  0.38 (0.34) &  0.37 (0.36) \\
 &       distilRoBERTa-Base &  0.34 (0.39) &  0.42 (0.34) &  0.31 (0.31) &  0.37 (0.38) &  0.39 (0.36) &  0.40 (0.39) \\
 \midrule
 \multirow{7}{*}{\bf A2} &                BERT-Large (R1) &  0.29 (0.53) &  0.30 (0.47) &  0.29 (0.44) &  0.25 (0.48) &  0.31 (0.47) &  0.33 (0.48) \\
 &    RoBERTa-Large (R2) &  0.19 (0.28) &  0.21 (0.26) &  0.20 (0.25) &  0.16 (0.23) &  0.19 (0.24) &  0.19 (0.27) \\
 &    RoBERTa-Large (R3) & \bf 0.50 (0.18) & \bf 0.43 (0.16) & \bf 0.41 (0.14) & \bf 0.44 (0.14) & \bf 0.45 (0.14) &  0.33 (\textbf{0.14}) \\
 &        BERT-Base &  0.25 (0.91) &  0.39 (0.88) &  0.30 (0.84) &  0.25 (0.86) &  0.31 (0.94) &  0.39 (0.91) \\
 &  distilBERT-Base &  0.22 (0.36) &  0.27 (0.33) &  0.24 (0.34) &  0.25 (0.34) &  0.23 (0.38) &  0.25 (0.33) \\
 &             RoBERTa-Base &  0.39 (0.48) &  0.40 (0.41) &  0.35 (0.38) &  0.39 (0.41) &  0.36 (0.41) &  0.42 (0.38) \\
 &       distilRoBERTa-Base &  0.42 (0.44) &  0.40 (0.38) &  0.36 (0.38) &  0.41 (0.37) &  0.39 (0.41) &  \textbf{0.43} (0.34) \\
 \midrule
  \multirow{7}{*}{\bf A3} &                BERT-Large (R1) &  0.34 (0.53) &  0.34 (0.51) &  \textbf{0.32} (0.50) &  0.29 (0.55) &  0.32 (0.49) &  0.31 (0.54) \\
 &    RoBERTa-Large (R2) &  0.29 (0.47) &  0.26 (0.54) &  0.26 (0.57) &  0.24 (0.58) &  0.27 (0.55) &  0.23 (0.58) \\
 &    RoBERTa-Large (R3) &  0.20 (0.43) &  0.23 (0.50) &  0.24 (0.53) &  0.25 (0.54) &  0.25 (0.54) &  0.23 (0.52) \\
 &        BERT-Base &  0.28 (0.80) &  \textbf{0.42} (0.66) &  0.26 (0.64) &  0.21 (0.60) &  0.30 (0.65) &  \textbf{0.37} (0.64) \\
 &  distilBERT-Base &  0.23 (\textbf{0.41}) &  0.25 (\textbf{0.35}) &  0.26 (\textbf{0.36}) &  0.24 (\textbf{0.35}) &  0.22 (\textbf{0.34}) &  0.22 (\textbf{0.35}) \\
 &             RoBERTa-Base & \textbf{0.41} (0.48) &  0.36 (0.40) &  0.29 (0.38) &  0.29 (0.43) &  0.34 (0.43) &  0.34 (0.43) \\
 &       distilRoBERTa-Base &  0.39 (0.42) &  0.33 (0.37) &  0.30 (0.37) &  \textbf{0.33} (0.36) &  \textbf{0.35} (0.37) &  0.32 (0.37) \\
 \midrule
   \multirow{7}{*}{\bf ANLI} &               BERT-Large (R1) &  0.22 (0.54) &  0.26 (0.52) &  0.26 (0.50) &  0.21 (0.53) &  0.26 (0.51) &  0.27 (0.53) \\
 &    RoBERTa-Large (R2) &  0.41 (0.26) &  0.37 (0.33) &  0.34 (0.37) &  0.33 (0.34) &  0.35 (0.33) &  0.32 (0.37) \\
 &    RoBERTa-Large (R3) &  \bf 0.52 (0.20) &  \bf 0.44 (0.27) &  \bf 0.41 (0.30) & \bf 0.45 (0.26) & \bf 0.45 (0.28) & \bf 0.39 (0.28) \\
 &        BERT-Base &  0.25 (0.89) &  0.40 (0.80) &  0.28 (0.76) &  0.24 (0.77) &  0.31 (0.83) &  0.36 (0.78) \\
 &  distilBERT-Base &  0.21 (0.37) &  0.25 (0.34) &  0.24 (0.34) &  0.24 (0.33) &  0.21 (0.36) &  0.23 (0.33) \\
 &             RoBERTa-Base &  0.37 (0.45) &  0.40 (0.39) &  0.31 (0.37) &  0.34 (0.41) &  0.36 (0.40) &  0.37 (0.39) \\
 &       distilRoBERTa-Base &  0.38 (0.41) &  0.38 (0.36) &  0.32 (0.36) &  0.37 (0.37) &  0.37 (0.38) &  0.37 (0.37) \\
    \bottomrule
    \end{tabular}
    \end{adjustbox}
    \caption{Mean probability of the correct label (mean entropy of label predictions) for each model on each top level annotation tag. 
    Bolded numbers correspond to the highest correct label probability and lowest entropy respectively. Recall that the entropy for three equiprobable outcomes (i.e., random chance of three NLI labels) is upper bounded by $\approx 1.58$. See Appendix \ref{appendix:modelpreds}: \autoref{tab:modelpredstoplevel}--\autoref{tab:modelpredsimperfect} for full details on all models.}
    \label{tab:analysis}
\end{table*}

%% file: tables/main_model_genre.tex
\begin{table}[t]
    \centering
    \small
    \begin{adjustbox}{max width=\linewidth}
\begin{tabular}{cccccc}
\toprule
\bf  Wikipedia & \bf Fiction & \bf News& \bf Procedural & \bf Legal & \bf RTE\\ \midrule
 0.55 (0.12) & 0.26 (0.68) & 0.22 (0.37) & 0.25 (0.62) & 0.25 (0.69) & 0.23 (0.57) \\
\bottomrule
\end{tabular}
\end{adjustbox}
\caption{Mean probability of the correct label (mean entropy of label predictions) for RoBERTa Ensemble (R3) for each genre.}
    \label{tab:genre-r3probs}
\end{table}

\begin{table}[t]
    \centering
    \small
    \begin{adjustbox}{max width=\linewidth}
\begin{tabular}{lrrrr}
\toprule
\bf Tag & \bf Wikipedia & \bf Fiction & \bf News & \bf Procedural \\
\midrule
Numerical &     39.7\% &   3.5\% &  17.2\% &             10.5\% \\
Basic &     36.8\% &   41.0\% &  54.5\% &             48.5\% \\
Reference &     27.5\% &   21.0\% &  19.7\% &             14.5\% \\
Tricky &     29.0\% &   28.5\% &  25.3\% &             24.0\% \\
Reasoning &     61.7\% &   67.5\% &  59.6\% &             62.5\% \\
Error &     2.8\% &   3.5\% &  1.0\% &             2.5\% \\
\bottomrule
\end{tabular}  
\end{adjustbox}
\caption{Percentage of  examples in each genre that contain a particular tag.}
    \label{tab:genreincidence}
\end{table}

\begin{table}[t]
    \centering
    \small
    \begin{adjustbox}{max width=\linewidth}
\begin{tabular}{lrrrr}
\toprule
\bf Tag     &   \bf   Wikipedia                & \bf Fiction            & \bf News & \bf Procedural \\\midrule
Numerical &  \textbf{0.58} (0.13) & \bf 0.35 (0.55) & 0.19 (\textbf{0.30}) & 0.29 (0.58) \\
Basic     &  0.51 (\textbf{0.12}) & 0.26 (0.70) & 0.22 (0.38) & 0.22 (0.53) \\
Reference &  0.54 (\textbf{0.12}) & 0.29 (0.73) & 0.21 (0.34) & 0.22 (0.59) \\
Tricky    &  0.52 (\textbf{0.12}) & 0.26 (0.72) & \textbf{0.26} (0.40) & \textbf{0.32} (0.63) \\
Reasoning &  0.53 (0.13) & 0.27 (0.64) & 0.22 (0.39) & 0.24 (0.59) \\
Imperfection   &  0.46 (0.12) &  0.28 (0.73) &  0.23 (0.41) &  0.25 (\textbf{0.51}) \\ 
\bottomrule
\end{tabular}
\end{adjustbox}
\caption{Mean probability of the correct label (mean entropy of label predictions) for RoBERTa-Large (R3) on each top level annotation tag per genre.}
    \label{tab:genre-breakdown}
\end{table}


%% file: appendix.tex

\input{tables/appendix_annotation_comparisons}

\section{Further Details on the Annotation Scheme}\label{app:furtherdetails}

\input{tables/appendix_full_ontology_table}

A full ontology, comprising all three levels, is provided together with examples in \autoref{tab:fullontolog}. Annotation guidelines for each tag were discussed verbally between the two annotators. The main expert annotator trained the second expert annotator by first walking through the annotation guidelines (i.e., \autoref{tab:ontolog}), answering any questions, and providing additional examples taken from their experience as necessary. The second expert then  annotated 20 randomly sampled examples from the R1 training set as practice. The two annotators subsequently discussed their selections on these training examples. Of course, there is some subjectivity inherent in this annotation scheme, which crucially relies on expert opinions about what information in the premise or hypothesis could be used to determine the correct label. After satisfactorily coming to a conclusion (i.e., a consensus), the second annotator was provided with another set of 20 randomly sample examples, this time from the R3 training set (to account for genre differences across rounds), and the discussion was repeated until consensus was reached. Several further discussions took place; once both annotators were confident in the second expert annotator's understanding of the scheme, the secondary annotator was provided with 3 random selections of 100 examples (one from each development set) as the final set to calculate inter-annotator agreement from. 

Throughout this process, the secondary annotator was provided with an exhaustive list of the 40 possible combinations of the three level tags from the initial annotator's annotations. These include: \textsc{Basic CauseEffect, Basic ComparativeSuperlative, Basic Coordination, Basic Facts, Basic Idioms, Basic Lexical Dissimilar, Basic Lexical Similar,
Basic Modus, Basic Negation, EventCoref, Imperfection Ambiguity, Imperfection Error, Imperfection NonNative, Imperfection Spelling, Numerical Cardinal, Numerical Cardinal Age, Numerical Cardinal Counting, Numerical Cardinal Dates, Numerical Cardinal Nominal, Numerical Cardinal Nominal Age, Numerical Cardinal Nominal Dates, Numerical Ordinal
Numerical Ordinal Age, Numerical Ordinal Dates, Numerical Ordinal Nominal, Numerical Ordinal Nominal Dates, Reasoning CauseEffect, Reasoning Containment Location, Reasoning Containment Parts, Reasoning Containment Times, Reasoning Debatable, Reasoning Facts, Reasoning-Plausibility Likely, Reasoning Plausibility Unlikely, Reference Coreference, Reference Family, Reference Names, Tricky Exhaustification, Tricky Pragmatic, Tricky Syntactic, Tricky Translation, Tricky Wordplay}. In addition to these tags, there are also some top-level tags associated with a \textsc{-0} tag; these are very rare (less than 30 of these in the dataset). The zero-tag was associated with examples that didn't fall into any second or third level categories. Finally, for the purposes of this paper, we collapsed two second-level tags \textsc{Basic CauseEffect} and \textsc{Basic Modus}\footnote{The identifier \textsc{Modus} labels classical logical reasoning types from analytic philosophy and was selected from the first word of some popular logical reasoning types: e.g., Modus Ponens, Modus Tollens, etc.} into `Basic Implications' because we felts the two are very related. In the actual dataset, tags at different levels are dash-separated, as in \textsc{Reasoning-Plausibility-Likely}.

Some tags required sophisticated linguistic domain knowledge, so more examples were provided on the annotation guidelines (some will be provided here). For example, the \textsc{Tricky-Exhaustification} is wholly novel, i.e., not adopted from any other tagset or similar to any existing tag. This tag marks examples where the original crowdworker-annotator assumed that only one predicate holds of the topic, and that other predicated don’t. Often \textsc{Tricky-Exhaustification} examples have the word ``only'' in the hypothesis, but that’s only a tendency: 
observe the context \textit{Linguistics is the scientific study of language, and involves an analysis of language form, language meaning, and language in context} and the 
hypothesis \textit{Form and meaning are the only aspects of language linguistics is concerned with}, which gets labeled as a contradiction.\footnote{This example also receives \textsc{Basic-Coordination}, and \textsc{Basic-Lexical-Similar} for ``involves'' and ``aspects''/``concerned with''.} For this example, the crowdworker-annotator wrote a  hypothesis that excludes one of the core properties of linguistics provided in the context and claims that the remaining two they list are the only core linguistic properties. To take another example, also a contradiction: For the context \textit{The Sound and the Fury is an American drama film directed by James Franco. It is the second film version of the novel of the same name by William Faulkner} and hypothesis 
\textit{Two Chainz actually wrote The Sound and the Fury}, we have a \textsc{Tricky-Exhaustification} tag. The Gricean Maxims of Relation and Quantity \citep{grice1975logic} would require the writer of the original context to list all the authors of The Sound and Fury, had there been more than one. Thus, we assume that the book only had one author, Faulkner. Since the author only listed Faulkner, we conclude that Two Chainz is \emph{not} in fact one of the authors of The Sound and the Fury.\footnote{This pair also gets \textsc{Tricky-Pragmatic}, and \textsc{EventCoref} and \textsc{Basic-Lexical-Similar} tags.}

The annotation guidelines also provided examples to aid in disentangling \textsc{Reference-Names} from \textsc{Reference-Coreference}, as they often appear together. \textsc{Reference-Coreference} should be used when resolving reference between non-string matched noun phrases (i.e. DP) is necessary to get the label: 
\textit{\textbf{Mary Smith$_{i}$} was a prolific author.~\textbf{She wrote a lot$_{i}$} had a lot of published works by 2010.$\Rightarrow$\textbf{Smith$_{i}$} published many works of literature.} \textsc{Reference-Names} is used when the label is predicated on either (i) a discussion of names, or (ii) resolving multiple names given to a person, but the reference in the hypothesis is an exact string match: \textit{\textbf{La Cygne$_{i}$} (pronounced \textbf{``luh SEEN''}) is a city in the south of France.$\Rightarrow$\textbf{La Cygne$_{i}$} is in France.}
Some examples require both: \textit{\textbf{Mary Beauregard Smith, the fourth grand Princess of Winchester} was a prolific author.$\Rightarrow$\textbf{Princess Mary} wrote a lot.}

\subsection{Inter-Annotator Agreement: Cohen's Kappa}\label{app:IAA}

\input{tables/appendix_interannotator_agreement}

Descriptively, annotators differed slightly in the number of tags they assign on average: the original annotator assigns fewer tags per example (Mean $=5.61$, Std. $=2.31$) than the second expert (Mean $=6.46$, Std. $=3.35$). The number of tags in the intersection of the two was predictably lower (Mean $=2.76$, Std.= $1.81$) than either annotator's average or the union (Mean $=8.39$, Std. $=2.60$).

In addition to agreement percentages that are reported in \autoref{sec:scheme} in the main text, we also report Cohen's kappa \citep{cohen1960} for top level tags and for lower level tags. Cohen's kappa is a ``conservative'' measure of agreement \citep{strijbos2006} that is ideal for measuring agreement between two raters. It is often preferred to percent agreement, since it can better account for accidental agreement \citep{mchugh2012}. Average Cohen's kappa for all labels tested independently are 0.34 for top level tags and 0.29 for tags from other levels. Note that Cohen's kappa ranges from -1 to 1 and scores in the 0.2 to 0.4 range are typically considered fair agreement \citep{landis1977}, i.e., a level that is often acceptable for non-sensitive applications \citep{cohen1960, mchugh2012} such as semantic annotation. 

Cohen's kappa is relatively high for \textsc{Numerical} tags, but seems a bit  low for a few of the others, in particular the lower level ones. For example, both annotators employed \textsc{Basic-Modus} only very rarely and percent agreement for presence/absence of this tag is 98.99\%, but with a Cohen's kappa of nearly zero. 

We also report precision and recall for our annotations. For low level tags, average precision and recall were comparable and averaged 0.44, whereas average precision and recall for top level tags averaged 0.64. Assuming that the main annotator's label was the gold label, this suggests that the secondary annotator understood and correctly applied the annotation guidelines to an acceptable level, but that the task is difficult and is somewhat subjective.

\subsection{Direct Comparisons to other Annotation Schemes}\label{app:directcomparisons}

Our scheme derives its inspiration from the wealth of prior work on types of sentential inference both within and from outside NLP \citep{fracas, jia-liang-2017-adversarial, white-etal-2017-inference, naik-etal-2018-stress}. When one implements an annotation scheme, one must decide on the level of depth one wants to achieve. On the one hand, a small number of tags can allow for easy annotation (by non-experts or even automatically), whereas on the other, a more complicated and complete annotation scheme (like, e.g., \citealt{fracas, bejar2012cognitive}) can allow for a better understanding of the full range of possible phenomena that might be relevant. (Note: our tags are greater in tag number than \citealt{naik-etal-2018-stress} but smaller and more manageable than \citealt{fracas} and \citealt{bejar2012cognitive}). 
We wanted annotations that allow for an evaluation of model behavior on a phenomenon-by-phenomenon basis, in the spirit of \citet{weston2016towards,wang-etal-2018-glue, jeretic2020natural}---but unlike \citet{jia-liang-2017-adversarial}. We also wanted to be able to detect interactions between phenomena \citep{sammons-etal-2010-ask}. Thus, we implemented our hierarchical scheme (for flexible tag-set size) in a way that could provide all these desiderata. 

\autoref{tab:schemecompare} provides a by-tag comparison between our annotation scheme and several others. Only direct comparisons are listed in the table; in other cases, our scheme had two tags where another scheme had one, or vice versa. Some of these examples are listed below, by the particular entailment types for each annotation scheme.

Several labels from the \citet{naik-etal-2018-stress}'s concur with ours, but our taxonomy has much wider coverage. In fact, it is a near proper superset of their scheme. Both taxonomies have a \textsc{Negation} tag, an \textsc{Ambiguity} tag, a \textsc{Real World Knowledge}---which is for us is labeled \textsc{Reasoning-Facts}, and a \textsc{Anonymy} tag---which for us is \textsc{Basic-Lexical-Dissimilar}. Additionally, both taxonomies have a tag for Numerical reasoning. We didn't include ``word overlap'' as that is easily automatable and would thus be an inappropriate use of limited hand-annotation time. Instead, we do include a more flexible version of word overlap in our \textsc{Basic-Lexical-Similar} tag, which accounts not only for synonym at the word level, but also for phrase level paraphrases. 

Our scheme can handle nearly all of the suggested reasoning types in \citet{sammons-etal-2010-ask}. For example, their `numerical reasoning' tag maps onto a combination of \textsc{Numerical} tags and \textsc{Reasoning-Facts} to account for external mathematical knowledge for us. A combination of their `kinship' and `parent-sibling' tags is present in our \textsc{Reference-Family} tag. One important difference between our scheme and theirs is that we do not separate negative and positive occurrences of the phenomena; both would appear under the same tag for us. One could imagine performing a further round of annotation on the ANLI data to separate positive from negative occurrences as \citeauthor{sammons-etal-2010-ask} does.

Several of the intuitions of the  \citep{lobue-yates-2011-types} taxonomy are present in our scheme. For example, their `arithmetic' tag would roughly correspond to a combination of our \textsc{Numerical-Cardinal} and \textsc{Reasoning-Facts} (i.e., for mathematical reasoning). Their ``preconditions'' tag would roughly correspond to our \textsc{Tricky-Pragmatic} tag. Interestingly, our \textsc{Tricky-Exhaustification} tag seems to be a combination of their `mutual exclusivity' and `omniscience' and `functionality' tags. Other relationships between our tags and theirs are provided in \autoref{tab:schemecompare}.

Many of our numerical reasoning types were inspired by \citet{ravichander-etal-2019-equate}, which showed that many NLI systems perform very poorly on many types of numerical reasoning. In addition to including cardinal and ordinal tags, as they do, we take their ideas one step further and also tag numerical examples where the numbers are not playing canonical roles as numerical object (e.g., \textsc{Numerical-Nominal} and \textsc{Numerical-Counting}). We also expand on their basic numerical types by specifying whether a number refers to a date or an age. For any of their examples requiring numerical reasoning, we would assign \textsc{Numerical} as a top level tag, as well as a \textsc{Reasoning-Facts} tag, as we described in the paragraph above. A similar set of tags would be present for their ``lexical inference'' examples where, e.g., it is necessary to know that `m' refers to `meters' when it follows a number; in this case, we would additionally include a \textsc{Tricky-Wordplay} tag.

\newcite{rozen-etal-2019-diversify}'s tagset also has some overlap with our tagset, although none directly. They present two automatically generated datasets. One targets comparative reasoning about numbers---i.e., corresponding to a combination of our \textsc{Numerical-Cardinal} and \textsc{Basic-ComparativeSuperlative} tags---and the other targets dative-alternation---which, like \citep{poliak-etal-2018-collecting}'s recasting of VerbNet, would probably correspond to our \textsc{Tricky-Syntactic} tag.

The annotation tagset of \citet{poliak-etal-2018-collecting} overlaps with ours in a few tags. For example, their `pun' tag is a proper subset of our \textsc{Tricky-Wordplay} tag. Their `NER' and `Gendered Anaphora' fall under our \textsc{Reference-Coreference} and \textsc{Reference-Names} tags. Their recasting of the \cite[MegaVeridicality]{white2018} dataset would have some overlap with our \textsc{Tricky-Pragmatic} tag, for example, for the factive pair \textit{Someone knew something happened. $\Rightarrow$ something happened.}. Similarly, their examples recast from \citet[VerbNet]{schuler2005verbnet} would likely recieve our \textsc{Tricky-Syntactic} tag for argument structure alternation, in at least some cases. 

In comparison with \citet{white-etal-2017-inference}, which uses pre-existing semantic annotations to create an RTE/NLI formatted dataset. This approach has several strong benefits, not the least of which is its use of minimal pairs to generate examples that can pinpoint exact failure points. For the first of our goals---understanding the contents of ANLI in particular---it would be interesting to have such annotations, and this could be a potentially fruitful future direction for research. But for the other---understanding current model performance on ANLI---it is not immediately clear to us that annotating ANLI for lexical semantic properties of predicates and their arguments (e.g., volition, awareness, and change of state) would help. Therefore, we leave it for future work for now.

From the above pairwise comparisons between existing annotation schemes (or data creation schemes), it should be clear there are many shared intuitions and many works are attempting to capture similar phenomena. We believe our tags thread the needle in a way that incorporates the best parts of the older annotation schemes while also innovating new phenomena and ways to view phenomena in relation to each other. Specific to the second point, very few of the schemes cited above arrange low level phenomena into a comprehensive multilevel hierarchy. This is one of the main benefits of our scheme. A hierarchy allows us to compare models at multiple levels, and hopefully, as our models improve, it can allow us to explore transfer between different reasoning types.

\input{tables/appendix_most_freq}

\section{Dataset Properties}\label{appendix:properties}

We measure the length of words and sentences in ANLI across all rounds and across all gold labels. We also draw a comparison to SNLI and MultiNLI, as other relevant large scale NLI datasets in \autoref{tab:words}. We also report length of rationales in \autoref{tab:rationales}.

\input{tables/appendix_stats_table_datasets}

As the table shows, the statistics across classification labels are roughly the same within each dataset. It is easy to see that ANLI contains much longer contexts than both MNLI and SNLI. Overall, ANLI and MNLI appear more similar in statistics to each other than to SNLI: both have longer statements and longer words.

We also look at the most frequent words for the contexts, statements and rationales. \autoref{tab:words_round_gold} shows the most frequent words used by round and gold label. \autoref{tab:words_top_annot_tags} shows the most frequent words by annotation tag. We analyzed the top 25 most frequent words (with stopwords removed based on the NLTK\footnote{\url{https://www.nltk.org/}} stopword list) in development set contexts, statements, and rationales, for the entire dataset, by round, and by gold label (see \autoref{tab:words_round_gold} in the Appendix), as well as for by top-level annotation tag (see \autoref{tab:words_top_annot_tags} in the Appendix). The most frequent words in contexts reflect the domains of the original texts. We note that words from Wikipedia contexts predictably figure prominently in the most frequent words lists, including, for example `film', `album', `directed', `football', `band', `television'. References to nations, such as `american', `state', and `national' are also common. On the other hand, statements were written by crowdworkers, and show a preference instead for terms like `born', `died', and `people', suggesting again, that Wikipedia contexts, consisting largely of biographies, have a specific genre effect on constructed statements. Several examples appear in the top 25 most frequent words for both statements and contexts, including `film', `american', `one', `two', `film', `not', `first', `new', `played',`album', and `city'. In particular, words in contexts such as `one', `first', `new', and `best' appear to be opposed by (near) antonyms such as `two', `last', `old', `least', and `less' in statements. This suggests the words present in a context might affect how crowdworkers construct statements. Finally, we observe that the top 25 most frequent words in contexts are generally used roughly 3 times as often as their corresponding versions in statements. This suggests that the vocabulary used in statements is wider and more varied than the vocabulary used in contexts.

\section{Analyzing Annotator Rationales}\label{appendix:rationaleproperties}

\input{tables/appendix_rationales_table}

We observe that the most frequent words in rationales differ from those in contexts and statements.  The original annotators showed is a preference for using `statement' and `context' in their rationales to refer to example pairs, as well as `system' to refer to the model; this last term is likely due to the fact that the name of the Mechanical Turk task used to employ crowdworkers in the original data collection was called ``Beat the System'' \citep[App. E]{nie-etal-2020-adversarial}. The set of most frequent words in rationales also contains, predictably, references to the model performance (e.g., `correct', `incorrect'), and to speech act verbs (e.g., `says', `states'). Interestingly, there is a higher number of verbs denoting mental states (e.g.,  `think', `know', `confused'), which suggests that the annotators could be ascribing theory of mind to the system, or at least using mental-state terms metaphorically---which could an artifact of the \newcite{nie-etal-2020-adversarial} data collection procedure that encourages them to think of the model as an adversary. Moreover, rationales contain more modal qualifiers (e.g., `probably', `may', `could'), which are often used to mark uncertainty, suggesting that the annotators aware of the fact that their rationales might be biased by their human expectations. Finally, we note that the top 25 most frequent words used in rationales are much more commonly used than their counterparts either in contexts (by roughly two times) or in statements (by roughly 5-6 times). This suggests that the genre of text created by crowdworkers in writing rationales is more narrow than the original context texts (from domains such as Wikipedia), and crowdworker annotated statement text.

\input{tables/appendix_incidence_tables}


\section{Tag Breakdowns}\label{appendix:tagbreakdown}

\autoref{tab:fullincidence} shows a breakdown of second-level tag incidence by top-level tag.



\input{tables/appendix_heatmap_pvalues_table}

\section{Model Correlation Significance}

The model comparison p-values are reported in \autoref{tab:heatmapPearsonP}.

\input{tables/appendix_model_preds_table_specifictags}

\input{tables/appendix_model_genre_full}

\section{Model Predictions Breakdown by Tag}\label{appendix:modelpreds}

Full model predictions by top level label are provided in \autoref{tab:modelpredstoplevel}. More detailed model prediction breakdowns by specific tags are provided in \autoref{tab:modelpredsbasic} (Basic), \autoref{tab:modelpredsnumerical} (Numerical), \autoref{tab:modelpredsreasoning} (Reasoning), \autoref{tab:modelpredsreference} (Reference \& Names), \autoref{tab:modelpredstricky} (Tricky), \autoref{tab:modelpredsimperfect} (Imperfections). 

For \textsc{Numerical}, \textsc{Counting} is hardest, which makes sense given that \textsc{Counting} examples are relatively rare, and require that one actually counts phrases in the text, which is a metalinguistic skill. \textsc{Ordinal} is the next most difficult category, perhaps because, like \textsc{Counting} examples, \textsc{Ordinal} examples are relatively rare.\footnote{Additionally, it seems difficult for models to bootstrap their \textsc{Cardinal} number knowledge for \textsc{Ordinal} numbers. One might hope that a model could bootstrap its knowledge of the order of cardinal numbers (e.g., that \textit{one} comes before \textit{two} and \textit{three}) to perform well on their corresponding ordinals, However, numerical order information doesn't seem to be generally applied in these models. Perhaps this is because many common ordinal numbers in English are not morphologically composed of their cardinal counterparts (e.g., \textit{one} and \textit{first}, \textit{two} and \textit{second}.}
For \textsc{Basic},  \textsc{Implication}, \textsc{Idiom} and \textsc{Negation} were more difficult than \textsc{Lexical}, \textsc{Comparative \& Superlative} and \textsc{Coordination} (see \autoref{tab:modelpredsbasic} in the Appendix).
For \textsc{Reference}, there is a lot of variation in the behavior of different models,  particularly for the \textsc{Names} examples, although also for \textsc{Coreference} examples, making it difficult to determine which is more difficult. Finally,  for \textsc{Tricky} examples,  \textsc{Wordplay} examples the most difficult, again because these require complex metalinguistic abilities (i.e., word games, puns, and anagrams), but they are followed closely by \textsc{Exhaustification} examples, which require a complex type of pragmatic reasoning.\footnote{See \newcite{chierchia2004scalar} for a summary of the linguistic theory on exhaustification, although we adopt a wider definition of the phenomenon for the tag here as in \autoref{tab:fullontolog}.}

\section{Model Predictions Breakdown by Domain}

\autoref{tab:modelpredsgenreall} shows the breakdown by genre.

%% file: tables/appendix_annotation_comparisons.tex
\begin{table*}[t]
    \centering
    \scriptsize
    \begin{adjustbox}{max width=\linewidth}
    \begin{tabular}{cr}
    \toprule
    Our Scheme's Tag & Other Scheme's Tag (Citation) \\ \midrule
    \textsc{Basic-Negation} & Negation  \citep{naik-etal-2018-stress} \\
    \textsc{Basic-Lexical-Dissimilar} & Antonymy  \citep{naik-etal-2018-stress},  Contrast  \citep{bejar2012cognitive}; Ch. 3\footnote{The chapter snippet is available \href{https://drive.google.com/file/d/0BzcZKTSeYL8VenY0QkVpZVpxYnc/view}{here}.}\\
    \textsc{Basic-Lexical-Similar} & Overlap \citep{naik-etal-2018-stress},  Similar \citep{bejar2012cognitive}; Ch. 3\\
    \textsc{Basic-CauseEffect} & Cause-Purpose \citep{bejar2012cognitive}; Ch. 3, cause  \citep{sammons-etal-2010-ask},  Cause and Effect \citep{lobue-yates-2011-types}\\
    \textsc{Basic-Coordination} & Conjoined Noun Phrases  \citep{fracas}\\
    \textsc{Basic-ComparativeSuperlative} & Comparatives  \citep{fracas}\\
    \textsc{Numerical} & numeric reasoning, numerical quantity \citep{sammons-etal-2010-ask}\\
    \textsc{Numerical-Cardinal} & cardinal \citep{ravichander-etal-2019-equate}\\
    \textsc{Numerical-Ordinal} & ordinal  \citep{ravichander-etal-2019-equate}\\
    \textsc{Reference-Coreference} & Anaphora (Inter-Sentential, Intra-Sentential) \citep{fracas},  coreference \citep{sammons-etal-2010-ask}\\
    \textsc{Reference-Coreference} with \textsc{Reference-Names} &  Representation \citep{bejar2012cognitive}; Ch. 3\\
    \textsc{Reference-Family} & parent-sibling, kinship \citep{sammons-etal-2010-ask} \\
    \textsc{Reference-Names} & name \citep{sammons-etal-2010-ask} \\
    \textsc{Reasoning-Debatable} & Cultural/Situational \citep{lobue-yates-2011-types}\\
    \textsc{Reasoning-Plausibility-Likely} & Probabilistic Dependency \citep{lobue-yates-2011-types}\\
    \textsc{Reasoning-Containment-Times} & Temporal Adverbials \citep{fracas},  Space-Time \citep{bejar2012cognitive}; Ch. 3,  event chain, temporal \citep{sammons-etal-2010-ask} \\
    \textsc{Reasoning-Containment-Location} & spatial reasoning \citep{sammons-etal-2010-ask}, Geometry \citep{lobue-yates-2011-types} \\
    \textsc{Reasoning-Containment-Parts} & Part-Whole, Class-Inclusions \citep{bejar2012cognitive}; Ch. 3,  has-parts \citep{lobue-yates-2011-types}\\
    \textsc{Reasoning-Facts} & Real World Knowledge  \citealt{naik-etal-2018-stress, clark2018, bernardy2019kind}\\
    \textsc{Tricky-Syntactic} & passive-active, missing argument, missing relation, simple rewrite,  \citep{sammons-etal-2010-ask} \\
    \textsc{Imperfections-Ambiguity} & Ambiguity  \citealt{naik-etal-2018-stress}\\
    \bottomrule
    \end{tabular}
   \end{adjustbox}
    \caption{Comparisons between our tagset and tags from other annotation schemes. }
    \label{tab:schemecompare}
\end{table*}

%% file: tables/appendix_full_ontology_table.tex
\begin{table*}[t]
    \centering
    \tiny
    \begin{tabular}{p{3em}p{3em}p{3em}p{24em}p{12em}ccp{9em}}
    \toprule
 \bf Top Level & \bf Second Level & \bf Third Level & \bf Context & \bf Hypothesis & \bf Round & \bf Label & \bf Other Tags \\\midrule 
\multirow{6}{*}{Num.} & \multirow{2}{*}{Cardinal} & Dates & Otryadyn G\"undegmaa (\textellipsis born \textbf{23 May} 1978), is a Mongolian sports shooter. \textellipsis & Otryadyn G\"undegmaa was born on May 23rd & A1 & E (N) & Ordinal, Dates \\
& & Ages & \textellipsis John Fox probably won’t roam an NFL sideline again\textellipsis \textbf{the 63-year-old} Fox will now move into an analyst role\textellipsis	& John Fox \textbf{is under 60 years old}. & A3 & C (E) & Ref., Coref.\\
& \multirow{2}{*}{Ordinal} 
& Dates & Black Robe\textellipsis is a historical novel by Brian Moore set in New France in \textbf{the 17th century}\textellipsis & Black Robe is a novel set in New France in \textbf{the mid 1600s}  & A2 & N (E) & Reasoning, Plaus., Likely, Cardinal \\
& & Ages & John Barnard (\textbf{6 July 1794} at Chislehurst, Kent; \textbf{died 17 November 1878} at Cambridge, England) was an English amateur cricketer who played first-class cricket from 1815 to 1830. M\textellipsis	& John Barnard died before \textbf{his fifth birthday}. & A1 & C (N) & Cardinal, Dates, Reasoning, Facts \\
& Counting & & \textellipsis The Demand Institute \textbf{was founded} in 2012 \textbf{by Mark Leiter and Jonathan Spector}\textellipsis &	The Demand Institute \textbf{was founded by two men}. & A2 & E (N) & Ref., Names\\
& Nominal & & Ra\'ul Alberto Osella (born 8 June 1984 in Morteros) is an Argentine association footballer \textellipsis He played \textbf{FIFA U-17 World Cup Final} for Argentina national team in \textbf{2001}. \textellipsis	& Raul Alberto Osella no longer plays for \textbf{the FIFA U-17 Argentina team}.  & A2 & E (N) & Reasoning, Facts, Tricky, Exhaust., Cardinal, Age, Dates \\ \midrule
\multirow{6}{*}{Basic} &  Lexical & & \textellipsis The \textbf{dating app Hater}, which matches users based on the things they hate, has compiled all of their data to create a map of \textbf{the foods everyone hates}\textellipsis &	Hater is an \textbf{app designed for foodies in relationships}. & A3 & C (N) & \\
& Comp.\& Super. & & \textellipsis try to hit your shot onto the upslope because they are \textbf{easier} putts to make \textbf{opposed to} downhill putts. & 	Upslope putts are \textbf{simple to do} & A3 & N (E)  \\
& Implic. & & [DANIDA]\textellipsis provides humanitarian aid \textellipsis to developing countries\textellipsis	&	\textbf{Focusing on developing countries}, DANIDA hopes to improve citizens of different countries lives.  & A2 & E (N) \\
& Idioms & & \textellipsis he set to work to hunt for his dear money\textellipsis he found nothing; \textbf{all had been spent}\textellipsis	& The money \textbf{got up and walked away}. & A3 & N (C) & Reasoning, Plaus., Unlikely  \\
& Negation &  & Bernardo Provenzano \textellipsis \textbf{was suspected of having been the head of the Corleonesi} \textellipsis & It was \textbf{never confirmed that Bernardo Provenzano was the leader of the Corleonesi}. & A2 & E (N) & Tricky, Prag.\\
& Coord. &  & \textellipsis Dan went home and started cooking a steak. However, Dan accidentally \textbf{burned the steak}\textellipsis. & 	The steak was \textbf{cooked for too long or on too high a temperature}. & A3 & E (N) & Basic, Lexical, Tricky, Prag. \\ \midrule
\multirow{3}{*}{Ref.} & Coref. & & \textellipsis Tim was a tutor. \textellipsis His latest student really pushed him, though. Tim could not get through to \textbf{him}. He had to give up\textellipsis & Tim gave up on \textbf{her} eventually. & A3 & C (E) \\
& Names & & \textbf{Never Shout Never} is an EP  \textbf{by Never Shout Never} which was released  December 8, 2009.\textellipsis & 	\textbf{Never Shout Never} has a \textbf{self titled} EP. & A1 & E (N)  & \\
& Family & & Sir Hugh Montgomery \textellipsis was the \textbf{son} of Adam Montgomery, the 5th Laird of Braidstane, by his \textbf{wife and cousin}.	& 	Sir Hugh Montgomery had at least one \textbf{sibling}. & A2 & N (E) & Reasoning, Plaus., Likely\\\midrule
\multirow{4}{*}{Tricky} & Syntactic & & Gunby\textellipsis \textbf{is situated close to the borders with Leicestershire and Rutland}, and 9 mi south from \textbf{Grantham}\textellipsis &	Gunby \textbf{borders Rutland an Grantham}. & A1 &  C (E) & Imperfect., Spelling \\
& Prag. & & \textellipsis Singh won \textbf{the award} for Women Leadership in Industry\textellipsis & \textellipsis Singh won \textbf{many awards} for Women in Leadership in Industry. & A3 &  C (N) & \\
& Exhaust. &  & Linguistics \textellipsis \textbf{involves an analysis of language form, language meaning, and language in context}. \textellipsis & 	Form and meaning are \textbf{the only aspects} of language linguistics is concerned with. & A1 & C (N) & \\
& Wordplay & & \textellipsis Brock Lesnar and Braun Strowman will both be under \textellipsis on \textbf{Raw}\textellipsis &	\textbf{Raw is not an anagram of war}  & A3 & C (E) & \\ \midrule
\multirow{7}{*}{Reasoning} & \multirow{3}{*}{Plaus.} & Likely & \textbf{B. Dalton Bookseller\textellipsis founded in 1966 by Bruce Dayton}, a member of the same family that operated the Dayton's department store chain\textellipsis & \textbf{Bruce Dayton founded the Dayton's department store chain}. & A1 & C (E) & Ref., Names\\
& & Unlikely  & The Disenchanted Forest is a 1999 documentary film that follows endangered orphan orangutans \textellipsis returned to their rainforest home. \textellipsis &	The Disenchanted Forest is \textellipsis about \textbf{orangutans trying to learn how to fly by building their own planes}\textellipsis & A2 & C (N) & Reasoning, Facts \\
& & Debatable & The Hitchhiker's Guide to the Galaxy is a 2005 British-American \textbf{comic science fiction film}\textellipsis &	Hitchhiker's Guide to the Galaxy is \textbf{a humorous film}. & A1 & N (E) & Basic, Lexical\\
& Facts &  & \textellipsis [Joey] decided to make [his mom] \textbf{pretend tea}. He got some hot water from the tap and mixed in the herb. But \textbf{to his shock}, his mom really drank the tea! \textbf{She said the herb he'd picked was chamomile}, a delicious tea!	 & \textbf{Joey knew how to make chamomile tea}. & A3 & C (E)\\ 
& \multirow{3}{*}{Contain.} & Parts & Milky Way Farm in Giles County, Tennessee, is \textbf{the former estate of Franklin C. Mars} \textellipsis its manor house is now a venue for special events.	& \textbf{The barn} is occassionaly staged for photo shoots. & A1 & N (C) & Plaus., Unlikely, Imperfect., Spelling\\
& & Loc. & Latin Jam Workout is a Latin Dance Fitness Program\textellipsis [f]ounded in 2007 \textbf{in Los Angeles, California}, Latin Jam Workout combines \textellipsis music with dance\textellipsis	& Latin Jam Workout was not created in \textbf{a latin american country} & A2 & E (C) & Basic, Negation \\
& & Times & Forbidden Heaven is a 1935 American drama film\textellipsis released \textbf{on October 5, 1935} \textellipsis	& Forbidden Heaven is \textellipsis film released \textbf{in the same month as the holiday Halloween}. & A1 & & Facts \\ \midrule
\multirow{5}{*}{Imperfect.} & Error & & Albert Levitt (March 14, 1887 – \textbf{June 18, 1968}) was a judge, law professor, attorney, and candidate for political office. \textellipsis	&	Albert Levitt \textellipsis held several positions in the legal field during his life, (\textbf{which ended in the summer of 1978})\textellipsis &  A2 & N (C) & Num., Cardinal, Dates\\
& Ambig. & & Diablo is a 2015 Canadian-American psychological western \textellipsis starring \textbf{Scott Eastwood}\textellipsis It was the first Western starring \textbf{Eastwood}, the son of Western icon \textbf{Clint Eastwood}. &\textbf{It} was the last western starring \textbf{Eastwood} & A2 & C (N) & Ref., Coref., Label, Basic, Comp.\&Sup., Lexical, Num., Ordinal, Family  \\
& Spelling & & ``Call My Name'' is a song \textbf{recorded by Pietro Lombardi} from his first studio album ``Jackpot''\textellipsis It was \textbf{written and produced by} ``DSDS'' jury member Dieter Bohlen\textellipsis. &	``Call my Name'' was \textbf{written and recorded by Pierrot Lombardi} for his album "Jackpot". & A1 & C (E) & Tricky, Syntactic, Imperfect., Spelling \\
& Translat. & & Club \textbf{Deportivo D\'enia} is a Spanish football team\textellipsis it plays in \textbf{Divisiones Regionales de F\'utbol} \textellipsis holding home games at ``\textbf{Estadio Diego Mena Cuesta}'',\textellipsis	&	Club \textbf{Deportivo D\'enia} plays in the Spanish village ``\textbf{Estadio Diego Mena Cuesta}''. & A2 & C (E) &  Tricky, Syntactic\\
\bottomrule
\end{tabular}
    \caption{Examples from the full scheme.
    }
    \label{tab:fullontolog}
\end{table*}

%% file: tables/appendix_interannotator_agreement.tex
\begin{table*}[t]
    \centering
    \scriptsize
    \begin{adjustbox}{max width=\linewidth}
    \begin{tabular}{lrr}
    \toprule
    Tag & Agreement (\%) & Cohen's Kappa 
    \\ \midrule
    \textsc{Numerical}                        &   \bf 85.9\% &        \bf  0.68 \\ 
    \textsc{Basic}                            &   73.9\% &        \bf  0.45 \\
    \textsc{Reasoning}                        &   60.8\% &          0.16 \\
    \textsc{Reference}                        &   67.3\% &          0.27 \\
    \textsc{Tricky}                           &   68.3\% &          0.25 \\
    \textsc{Imperfections}                    &  78.9\% &          0.31 \\
    \midrule
\textsc{Numerical-Cardinal}               & \bf   92.5\% &      \bf    0.55 \\
\textsc{Numerical-Cardinal-Age}           & \bf   98.5\% &      \bf    0.79 \\
\textsc{Numerical-Cardinal-Counting}      & \bf   99.5\% &      \bf    0.91 \\
\textsc{Numerical-Cardinal-Dates}         & \bf   91.5\% &      \bf    0.74 \\
\textsc{Numerical-Cardinal-Nominal}       & \bf   97.5\% &         -0.01 \\
\textsc{Numerical-Cardinal-Nominal-Dates} & \bf   97.5\% &          0.00 \\
\textsc{Numerical-Ordinal}                & \bf   99.0\% &      \bf    0.50 \\
\textsc{Numerical-Ordinal-Dates}          & \bf   98.5\% &         -0.01 \\
\textsc{Basic-0}                          & \bf   98.5\% &         -0.01 \\
\textsc{Basic-CauseEffect}                & \bf   94.0\% &          0.11 \\
\textsc{Basic-ComparativeSuperlative}     & \bf  95.0\% &          0.36 \\
\textsc{Basic-Conjunction}                & \bf   87.9\% &          0.24 \\
\textsc{Basic-Idiom}                      & \bf   97.5\% &          0.27 \\
\textsc{Basic-Lexical-0}                  &      79.4\% &          0.36 \\
\textsc{Basic-Negation}                   & \bf   95.0\% &      \bf    0.73 \\
\textsc{Reasoning-0}                      & \bf   99.5\% &          0.00 \\
\textsc{Reasoning-Containment-Location}   & \bf   97.0\% &       \bf   0.56 \\
\textsc{Reasoning-Containment-Time}       & \bf   94.5\% &          0.24 \\
\textsc{Reasoning-Debatable}              & \bf   88.4\% &          0.32 \\
\textsc{Reasoning-Facts}                  &         64.3\% &          0.26 \\
\textsc{Reference-Coreference}            &          69.8\% &          0.28 \\
\textsc{Reference-Family}                 & \bf   99.5\% &        \bf  0.85 \\
\textsc{Reference-Names}                  & \bf   88.4\% &          0.02 \\
\textsc{Tricky-Exhaustification}          & \bf   94.0\% &        \bf  0.47 \\
\textsc{Tricky-Presupposition}            &         81.4\% &          0.02 \\
\textsc{Tricky-Syntactic}                 &  \bf  87.4\% &          0.19 \\
\textsc{Tricky-Translation}               & \bf   93.5\% &        \bf  0.40 \\
\textsc{Tricky-Wordplay}                  & \bf   96.0\% &          0.18 \\
\textsc{Imperfections-0}                        &   \bf 93.0\% &         -0.01 \\
\textsc{Imperfections-Ambiguity}                &   \bf 93.0\% &          0.19 \\
\textsc{Imperfections-Error}                    &   \bf 94.5\% &         -0.02 \\
\textsc{Imperfections-NonNative}                &   \bf 94.5\% &      \bf    0.40 \\
\textsc{Imperfections-Spelling}                 &   \bf 94.0\% &          0.22 \\
\textsc{EventCoref}                       & \bf  89.9\% &          0.25 \\
\midrule
    Average & \bf 91.6\% & 0.29 \\
    \bottomrule
    \end{tabular}
   \end{adjustbox}
    \caption{Interannotator Agreement for 200 randomly sampled examples: Percent Agreement, Cohen's Kappa, and Counts for tags. Bolded examples show high inter-annotator agreement (above 85\% or Kappa of 0.4).
    }
    \label{tab:IAAallTags}
\end{table*}

%% file: tables/appendix_most_freq.tex
\begin{table*}[t]
    \centering
    \tiny
    \begin{adjustbox}{max width=\linewidth}
    \begin{tabular}{p{5em}p{16em}p{15em}p{15em}p{15em}}
    \toprule
    \bf Subset & Context & Statement & Rationale & Context$+$Statement \\
    \midrule
ANLI &  film (647),  american (588),  \textbf{known} (377),  first (376),  (born (365),  \textbf{also} (355),  one (342),  new (341),  released (296),  album (275),  \textbf{united} (249),  \textbf{directed} (240),  not (236),  – (218),  \textbf{based} (214),  \textbf{series} (196),  \textbf{best} (191),  may (188),  \textbf{band} (185),  \textbf{state} (182),  \textbf{football} (177),  two (175),  written (175),  \textbf{television} (175),  \textbf{national} (169),  \textbf{south} (165) &  not (252),  born (132),  years (120),  released (107),  one (87),  film (83),  first (82),  only (76),  \textbf{people} (75),  year (61),  \textbf{played} (58),  new (58),  two (54),  \textbf{made} (54),  album (49),  no (46),  \textbf{died} (46),  \textbf{won} (46),  \textbf{less} (44),  \textbf{last} (42),  american (41),  \textbf{years.} (40),  \textbf{three} (40),  written (38),  \textbf{used} (37),  \textbf{john} (37) &  not (1306),  \textbf{system} (753),  \textbf{statement} (494),  \textbf{know} (343),  \textbf{think} (274),  \textbf{definitely} (268),  \textbf{context} (261),  \textbf{correct} (243),  \textbf{difficult} (228),  only (224),  \textbf{doesn't} (223),  may (221),  \textbf{confused} (218),  no (200),  \textbf{says} (198),  \textbf{incorrect} (193),  \textbf{text} (184),  \textbf{could} (181),  \textbf{states} (166),  born (160),  one (155),  \textbf{say} (147),  years (146),  \textbf{don't} (140),  \textbf{would} (130),  \textbf{whether} (129) &  film (730),  american (629),  not (488),  first (458),  one (429),  known (414),  released (403),  new (399),  also (379),  (born (368),  album (324),  united (281),  directed (274),  based (238),  two (229),  born (226),  series (223),  played (221),  – (221),  best (220),  band (219),  only (213),  written (213),  football (208),  may (208),  state (204)\\  \midrule
R1 &   film (299),  american (272),  known (175),  (born (169),  first (158),  also (129),  released (119),  album (115),  directed (106),  based (104),  united (103),  new (97),  – (93),  football (88),  one (84),  band (77),  best (77),  south (73),  \textbf{former} (71),  written (70),  series (67),  played (67),  \textbf{march} (66),  city (65),  located (65),  television (64) &   born (65),  film (47),  not (46),  years (45),  released (43),  first (36),  died (26),  only (25),  american (24),  \textbf{population} (23),  \textbf{old} (23),  album (22),  won (22),  played (21),  directed (21),  new (19),  last (18),  football (18),  \textbf{century.} (18),  year (18),  united (17),  years. (16),  \textbf{world} (16),  written (16),  one (16),  based (16) &  not (392),  system (331),  know (135),  statement (126),  think (111),  context (105),  difficult (93),  definitely (86),  correct (80),  born (80),  only (75),  may (75),  confused (75),  incorrect (63),  could (62),  stated (62),  don't (59),  says (58),  doesn't (57),  \textbf{information} (54),  states (53),  no (53),  first (52),  \textbf{probably} (49),  used (48),  text (47) &  film (346),  american (296),  first (194),  known (188),  (born (170),  released (162),  also (140),  album (137),  directed (127),  united (120),  based (120),  new (116),  born (109),  football (106),  one (100),  – (94),  band (91),  best (89),  played (88),  written (86),  south (81),  world (79),  city (77),  series (77),  population (77),  name (77) \\ 
R2 & film (301),  american (266),  known (166),  (born (159),  also (146),  released (136),  new (128),  album (127),  first (126),  directed (114),  one (112),  series (110),  united (97),  – (95),  television (95),  band (87),  state (86),  based (83),  written (82),  \textbf{song} (79),  national (76),  played (74),  best (69),  located (67),  city (66),  football (66) &  not (75),  years (54),  released (53),  born (51),  one (32),  first (32),  film (31),  year (29),  \textbf{ago.} (24),  only (24),  played (23),  album (23),  known (22),  two (22),  new (21),  band (19),  made (18),  city (16),  no (16),  died (16),  john (15),  less (15),  won (15),  written (14),  people (14), \textbf{lived} (14) &  not (387),  system (198),  statement (125),  know (93),  doesn't (79),  difficult (78),  think (77),  years (74),  context (72),  confused (70),  may (65),  only (63),  born (61),  states (60),  correct (59),  no (56),  \textbf{ai} (55),  definitely (55),  released (52),  text (50),  incorrect (49),  say (48),  year (48),  could (45),  one (44),  says (42) &  film (332),  american (280),  released (189),  known (188),  (born (161),  also (159),  first (158),  album (150),  new (149),  one (144),  series (124),  directed (123),  band (106),  united (105),  television (101),  not (98),  played (97),  – (97),  written (96),  state (96),  song (89),  born (88),  based (87),  national (83),  city (82),  located (80)\\ 
R3 &  not (197),  one (146),  said (122),  new (116),  would (104),  first (92),  some (91),  make (87),  people (83),  may (83),  also (80),  \textbf{time} (77),  no (75),  -- (75),  \textbf{like} (74),  \textbf{get} (74),  last (72),  only (68),  two (68),  \textbf{police} (66),  made (61),  think (55),  \textbf{home} (54),  \textbf{go} (54),  \textbf{way} (53),  \textbf{many} (53) &  not (131),  people (48),  one (39),  only (27),  no (22),  made (21),  years (21),  speaker (19),  two (19),  new (18),  three (17),  used (16),  \textbf{use} (16),  \textbf{person} (16),  less (16),  born (16),  \textbf{good} (15),  make (14),  year (14),  first (14),  played (14),  \textbf{school} (13),  \textbf{government} (13),  didn't (13),  last (13),  some (13) &  not (527),  statement (243),  system (224),  definitely (127),  know (115),  correct (104),  says (98),  no (91),  doesn't (87),  text (87),  think (86),  only (86),  context (84),  incorrect (81),  may (81),  \textbf{model} (75),  could (74),  confused (73),  one (67),  said (66),  say (63),  whether (58),  difficult (57),  neither (57),  incorrect. (56),  would (53) &  not (328),  one (185),  new (134),  people (131),  said (127),  would (115),  first (106),  some (104),  make (101),  no (97),  may (95),  only (95),  two (87),  time (86),  last (85),  like (83),  get (82),  made (82),  also (80),  -- (75),  \textbf{police} (74),  use (67),  many (66),  three (63),  \textbf{home} (62),  \textbf{go} (62) \\\midrule
Contra. & american (219),  film (216),  new (146),  (born (129),  first (124),  also (116),  known (115),  united (110),  one (108),  released (94),  album (86),  – (81),  directed (78),  series (76),  may (72),  best (71),  television (70),  band (69),  not (68),  based (66),  written (65),  south (65),  national (63),  two (62),  \textbf{song} (60),  football (59) &  not (63),  years (55),  born (42),  film (37),  released (36),  first (31),  year (30),  only (28),  one (23),  new (23),  died (21),  people (19),  american (19),  won (19),  years. (19),  \textbf{world} (18),  three (18),  played (18),  album (17),  two (17),  less (17),  directed (17),  \textbf{old} (16),  made (16),  written (15),  \textbf{lived} (15) &  not (471),  system (269),  statement (174),  incorrect (121),  think (104),  definitely (90),  confused (87),  difficult (83),  only (78),  born (71),  says (63),  context (61),  years (57),  states (51),  one (50),  would (49),  incorrect. (47),  know (42),  name (42),  \textbf{probably} (41),  year (41),  \textbf{ai} (41),  could (40),  first (38),  may (38),  model (35) &  film (253),  american (238),  new (169),  first (155),  not (131),  one (131),  (born (130),  released (130),  known (126),  also (125),  united (119),  album (103),  directed (95),  series (88),  – (83),  band (82),  written (80),  two (79),  best (79),  may (78),  television (78),  south (77),  world (75),  based (74),  years (74),  football (72)\\ 
Neut. &  film (224),  american (198),  known (126),  first (118),  one (116),  released (115),  (born (112),  also (107),  album (101),  new (97),  not (95),  directed (93),  based (77),  united (74),  football (67),  may (61),  band (60),  best (60),  – (58),  city (55),  two (55),  national (54),  played (54),  series (53),  state (51),  \textbf{song} (51) &  not (63),  one (37),  born (36),  released (29),  only (28),  never (25),  played (24),  film (22),  people (21),  made (19),  first (18),  no (18),  new (17),  album (17),  won (17),  known (16),  population (15),  john (14),  two (14),  last (14),  name (13),  united (13),  died (12),  best (12),  football (11),  written (11) & not (608),  know (263),  system (236),  doesn't (157),  no (150),  context (147),  statement (146),  may (133),  say (125),  whether (124),  correct (123),  could (119),  neither (117),  don't (117),  only (110),  definitely (109),  text (102),  information (89),  \textbf{nor} (83),  mentioned (80),  think (80),  state (78),  says (71),  difficult (71),  incorrect (69),  confused (67) &  film (246),  american (208),  not (158),  one (153),  released (144),  known (142),  first (136),  album (118),  new (114),  (born (114),  also (112),  directed (101),  united (87),  based (83),  played (78),  football (78),  only (76),  best (72),  band (70),  two (69),  made (69),  city (66),  may (64),  born (63),  name (63),  written (60)\\ 
Entail. &  film (207),  american (171),  known (136),  first (134),  also (132),  (born (124),  one (118),  new (98),  album (88),  released (87),  – (79),  state (73),  not (73),  based (71),  directed (69),  series (67),  united (65),  played (61),  written (61),  best (60),  television (60),  \textbf{former} (60),  two (58),  band (56),  may (55),  \textbf{located} (53) &  not (63),  one (37),  born (36),  released (29),  only (28),  never (25),  played (24),  film (22),  people (21),  made (19),  first (18),  no (18),  new (17),  album (17),  won (17),  known (16),  population (15),  john (14),  two (14),  last (14),  name (13),  united (13),  died (12),  best (12),  football (11),  written (11) &  not (608),  know (263),  system (236),  doesn't (157),  no (150),  context (147),  statement (146),  may (133),  say (125),  whether (124),  correct (123),  could (119),  neither (117),  don't (117),  only (110),  definitely (109),  text (102),  information (89),  nor (83),  mentioned (80),  think (80),  state (78),  says (71),  difficult (71),  incorrect (69),  confused (67) &  film (231),  not (199),  american (183),  first (167),  known (146),  one (145),  also (142),  released (129),  (born (124),  new (116),  album (103),  born (91),  state (84),  years (82),  two (81),  based (81),  – (79),  directed (78),  played (77),  series (76),  united (75),  written (73),  people (71),  best (69),  band (67),  may (66) \\ 
\bottomrule
    \end{tabular}
    \end{adjustbox}
    \caption{Top 25 most common words used by round and gold label. Bolded words are used preferentially in particular subsets.}
    \label{tab:words_round_gold}
\end{table*}

\begin{table*}[t]
    \centering
    \tiny
    \begin{tabular}{p{5em}p{15em}p{14em}p{16em}p{14em}}
    \toprule
    \bf Subset & Context & Statement & Rationale & Context$+$Statement$+$Rationale \\
    \midrule
ANLI &  film (647),  american (588),  \textbf{known} (377),  first (376),  (born (365),  \textbf{also} (355),  one (342),  new (341),  released (296),  album (275),  \textbf{united} (249),  \textbf{directed} (240),  not (236),  – (218),  \textbf{based} (214),  \textbf{series} (196),  \textbf{best} (191),  may (188),  \textbf{band} (185),  \textbf{state} (182),  \textbf{football} (177),  two (175),  written (175),  \textbf{television} (175),  \textbf{national} (169),  \textbf{south} (165) &  not (252),  born (132),  years (120),  released (107),  one (87),  film (83),  first (82),  only (76),  \textbf{people} (75),  year (61),  \textbf{played} (58),  new (58),  two (54),  \textbf{made} (54),  album (49),  no (46),  \textbf{died} (46),  \textbf{won} (46),  \textbf{less} (44),  \textbf{last} (42),  american (41),  \textbf{years.} (40),  \textbf{three} (40),  written (38),  \textbf{used} (37),  \textbf{john} (37) &  not (1306),  \textbf{system} (753),  \textbf{statement} (494),  \textbf{know} (343),  \textbf{think} (274),  \textbf{definitely} (268),  \textbf{context} (261),  \textbf{correct} (243),  \textbf{difficult} (228),  only (224),  \textbf{doesn't} (223),  may (221),  \textbf{confused} (218),  no (200),  \textbf{says} (198),  \textbf{incorrect} (193),  \textbf{text} (184),  \textbf{could} (181),  \textbf{states} (166),  born (160),  one (155),  \textbf{say} (147),  years (146),  \textbf{don't} (140),  \textbf{would} (130),  \textbf{whether} (129) &  not (1794),  film (802),  system (781),  american (659),  one (584),  first (563),  statement (511),  released (504),  known (495),  also (467),  new (452),  only (437),  may (429),  know (387),  born (386),  (born (371),  album (362),  no (337),  think (337),  based (335),  years (332),  two (313),  states (313),  united (308),  state (304),  directed (301)\\  \midrule
Numerical & american (236),  film (211),  (born (162),  first (151),  known (138),  album (136),  new (129),  released (126),  also (117),  united (117),  one (109),  – (101),  band (87),  series (83),  best (82),  television (79),  directed (77),  football (76),  based (75),  state (74),  played (73),  second (72),  south (71),  world (70),  city (69),  states (65) & years (114),  born (79),  released (74),  first (61),  year (52),  not (44),  \textbf{died} (38),  \textbf{less} (37),  two (36),  one (35),  years. (34),  \textbf{three} (32),  population (30),  old (30),  film (28),  \textbf{ago.} (27),  album (26),  only (24),  old. (24),  \textbf{century.} (23),  last (23),  won (20),  \textbf{least} (20),  world (20),  second (18),  played (18) & not (344),  system (291),  statement (166),  years (137),  difficult (125),  born (115),  think (103),  definitely (102),  year (90),  confused (90),  only (88),  correct (84),  know (82),  context (77),  released (72),  may (71),  incorrect (70),  first (61),  text (60),  could (59),  would (57),  one (55),  says (51),  doesn't (50),  \textbf{mentioned} (49),  died (48) &  not (423),  system (297),  years (278),  first (273),  released (272),  film (265),  american (256),  born (231),  one (199),  album (181),  year (170),  statement (169),  (born (166),  known (160),  only (158),  new (158),  two (145),  may (140),  also (137),  united (134),  difficult (125),  based (124),  states (117),  think (112),  band (111),  second (109) \\
Basic  & film (238),  american (193),  one (143),  known (138),  new (135),  first (134),  also (132),  not (125),  released (105),  directed (104),  (born (100),  album (99),  state (97),  united (90),  may (83),  song (80),  based (78),  series (74),  best (74),  two (73),  television (72),  – (69),  south (68),  written (68),  said (65),  would (64) & not (219),  one (51),  people (41),  no (36),  film (31),  new (31),  released (28),  less (28),  never (27),  played (24),  only (24),  born (23),  two (23),  made (22),  album (21),  last (21),  first (21),  used (20),  least (18),  written (18),  three (17),  directed (17),  best (16),  years (16),  \textbf{movie} (16),  \textbf{good} (16) & not (546),  system (290),  statement (248),  know (125),  definitely (120),  think (115),  context (101),  says (101),  doesn't (97),  correct (92),  only (91),  confused (89),  may (88),  incorrect (83),  states (78),  no (76),  text (75),  could (69),  one (65),  difficult (61),  whether (58),  would (58),  say (56),  neither (54),  said (52),  model (50) &  not (890),  system (303),  film (298),  one (259),  statement (254),  american (227),  new (196),  first (191),  known (182),  also (181),  may (180),  only (176),  released (165),  no (154),  think (149),  know (146),  state (140),  would (137),  two (134),  directed (133),  album (132),  states (128),  based (127),  says (127),  people (126),  said (123) \\
Reference  & film (188),  american (163),  known (139),  (born (128),  also (112),  first (98),  one (85),  new (83),  directed (72),  – (71),  not (71),  released (70),  best (66),  united (61),  album (57),  television (56),  south (54),  world (54),  based (53),  may (52),  written (52),  series (50),  band (49),  ) (45),  two (45),  national (44) & not (70),  born (39),  years (33),  name (23),  film (21),  made (20),  won (19),  one (19),  people (19),  first (19),  only (17),  year (17),  played (16),  released (16),  died (16),  known (15),  band (15),  speaker (14),  new (14),  written (14),  three (13),  two (12),  no (12),  man (12),  directed (11),  album (10) & not (358),  system (199),  statement (112),  know (91),  think (71),  doesn't (70),  confused (67),  may (66),  context (60),  model (60),  only (57),  says (52),  correct (52),  could (51),  definitely (50),  name (50),  difficult (49),  born (46),  one (42),  probably (41),  would (41),  incorrect (40),  states (39),  don't (38),  no (35),  understand (34) &  not (499),  film (230),  system (207),  known (186),  american (171),  first (147),  one (146),  also (139),  (born (129),  may (126),  statement (122),  born (122),  new (112),  only (109),  name (105),  released (105),  know (104),  think (100),  directed (93),  years (89),  would (88),  written (84),  two (83),  states (82),  based (82),  best (80) \\
Tricky  & film (227),  american (142),  first (110),  known (104),  one (102),  also (99),  new (93),  (born (88),  album (83),  released (81),  directed (77),  based (75),  song (71),  not (68),  series (65),  written (61),  united (60),  band (59),  ) (55),  may (51),  – (50),  south (48),  only (48),  two (48),  television (46),  located (44) & not (82),  only (58),  born (33),  film (32),  released (27),  one (26),  two (22),  first (21),  made (19),  years (19),  new (18),  three (18),  played (16),  album (16),  american (16),  used (16),  people (14),  series (14),  wrote (13),  directed (13),  written (13),  also (13),  band (13),  known (13),  won (13),  \textbf{starts} (12) & not (386),  system (204),  statement (129),  only (88),  know (75),  think (73),  difficult (69),  context (67),  confused (66),  incorrect (63),  definitely (63),  may (57),  correct (54),  says (51),  states (49),  doesn't (48),  one (43),  name (42),  used (41),  text (41),  no (40),  \textbf{ai} (38),  don't (37),  \textbf{words} (36),  first (36),  could (35) &  not (536),  film (281),  system (208),  only (194),  one (171),  first (167),  american (166),  also (146),  known (141),  statement (133),  new (124),  released (123),  album (111),  may (110),  based (110),  directed (99),  two (92),  (born (89),  written (89),  series (88),  know (87),  song (86),  used (86),  made (86),  name (86),  think (85) \\
Reasoning & film (390),  american (363),  (born (245),  first (229),  also (227),  known (226),  new (219),  one (203),  released (173),  album (159),  united (154),  directed (151),  not (147),  based (138),  – (125),  football (124),  state (117),  national (116),  played (111),  best (110),  band (109),  television (108),  may (108),  series (106),  former (105),  south (104) & not (131),  born (92),  released (66),  years (60),  people (50),  first (49),  one (49),  film (43),  played (39),  year (36),  only (35),  new (35),  made (30),  never (30),  two (29),  died (27),  album (27),  won (26),  no (26),  known (25),  last (25),  american (24),  used (24),  united (22),  \textbf{john} (22),  city (22) & not (919),  system (466),  know (291),  statement (279),  context (188),  definitely (173),  correct (172),  doesn't (171),  think (164),  no (162),  may (162),  could (147),  difficult (144),  only (126),  say (126),  whether (123),  says (119),  confused (119),  text (118),  don't (114),  neither (110),  incorrect (110),  born (101),  one (96),  information (95),  states (92) &  not (1197),  system (483),  film (481),  american (411),  one (348),  first (335),  know (312),  released (307),  known (306),  also (292),  new (290),  statement (288),  may (281),  (born (250),  only (249),  born (249),  no (239),  state (218),  based (213),  album (206),  think (200),  played (196),  united (196),  context (191),  could (184),  doesn't (182) \\
Imperfections & film (87),  american (76),  also (54),  one (52),  first (47),  known (45),  released (45),  new (44),  album (42),  not (36),  based (35),  directed (35),  (born (35),  city (34),  united (33),  written (31),  two (30),  song (29),  – (26),  series (25),  band (25),  people (25),  television (24),  population (24),  name (24),  national (24) & not (38),  film (18),  people (14),  born (12),  written (12),  one (12),  only (11),  first (11),  made (10),  released (10),  new (10),  american (8),  city (8),  two (7),  years (7),  \textbf{popular} (7),  many (6),  different (6),  united (6),  album (6),  \textbf{street} (6),  show (6),  also (6),  population (6),  three (6),  \textbf{life} (5) & not (168),  system (82),  statement (70),  know (50),  correct (38),  context (35),  think (34),  says (32),  no (30),  definitely (29),  doesn't (28),  confused (26),  could (26),  incorrect (26),  one (24),  states (23),  only (23),  stated (22),  neither (22),  may (21),  model (21),  say (21),  text (20),  don't (20),  difficult (19),  state (19) &  not (242),  film (116),  american (94),  system (89),  one (88),  statement (72),  also (72),  first (71),  known (65),  released (64),  know (63),  new (58),  written (55),  based (54),  album (53),  only (52),  no (50),  two (49),  people (47),  think (46),  city (45),  may (44),  states (44),  made (43),  directed (42),  united (42) \\
\bottomrule
    \end{tabular}
    \caption{Top 25 most common words used by annotation tag. Bolded words are used preferentially in particular subsets.}
    \label{tab:words_top_annot_tags}
\end{table*}

%% file: tables/appendix_stats_table_datasets.tex
\begin{table*}[t]
    \centering
    \scriptsize
    \begin{adjustbox}{max width=\linewidth}
    \begin{tabular}{ccrrrrr}
    \toprule
   \multicolumn{2}{c}{\textbf{Dataset}} & \multicolumn{2}{c}{\textbf{Contexts}} & \multicolumn{2}{c}{\textbf{Statements}} \\
    & & Word$_{Len.}$  & Sent.$_{Len.}$ & Word$_{Len.}$  & Sent.$_{Len.}$ \\
    \midrule
    \multirow{7}{*}{\textbf{ANLI}} & All & 4.98 (0.60) & 55.6 (13.7) & 4.78 (0.76) & 10.3 (5.28) \\
    & A1 & 5.09 (\textbf{0.69})   & 54.1 (8.35) & 4.91 (0.74) & 11.0 (5.36) \\
    & A2 & 5.09 (0.47)   & 54.2 (8.24) & 4.80 (0.77) & 10.1 (4.95)\\
    & A3 & \textbf{4.73} (0.50)   & \textbf{59.2} (\textbf{21.5}) & 4.59 (0.76) & 9.5 (5.38)\\
    & C  & 5.00 (\textbf{0.79}) & 55.8 (13.8) & 4.76 (0.73) & \textbf{11.4 (6.51)}  \\  
    & N  & 4.97 (0.47) & 55.4 (13.8) & 4.83 (0.78) & 9.4 (4.49) \\
    & E  & 5.00 (0.49) & 55.7 (13.6) & 4.75 (0.78) & 10.3 (4.44)   \\ \midrule
    \multirow{6}{*}{\textbf{MNLI}} & All & 4.90 (0.97) & 19.5 (13.6)  & 4.82 (0.90) & 10.4 (4.43)  \\
    & M & \bf 4.88 (1.10)   & 19.3 (14.2) & 4.78 (0.92) & 9.9 (4.28) \\
    & MM & \bf 4.93 (0.87) & 19.7 (13.0) & 4.86 (0.89) & 10.8 (4.53) \\
    & C & 4.90 (0.97)  & 19.4 (13.6) & 4.79 (0.90) & 9.7 (3.99)  \\
    & N & 4.90 (0.98)  & 19.4 (13.8) & 4.79 (\textbf{0.85}) & 10.9 (4.46)  \\
    & E & 4.91 (0.96)  & 19.6 (13.5) & 4.86 (\textbf{0.95}) & 10.4 (4.71)  \\ \midrule
    \multirow{4}{*}{\textbf{SNLI}} & All & 4.31 (0.65) & 14.0 (6.32) & 4.23 (0.75) & 7.5 (3.14)  \\
    & C & 4.31 (0.64) & 14.0 (6.35) & 4.16 (0.71) & 7.4 (2.90)  \\
    & N & 4.31 (0.66) & 13.8 (6.28) & 4.26 (0.72) & 8.3 (\textbf{3.36})  \\
    & E & 4.31 (0.64) & 14.0 (6.31) & 4.26 (\textbf{0.81}) & 6.8 (2.90)  \\
    \bottomrule
    \end{tabular}
   \end{adjustbox}
    \caption{Average length of words and sentences in contexts, statements, and reasons for ANLI, MultiNLI, SNLI. Average and (standard deviation) are provided.}
    \label{tab:words}
\end{table*}

%% file: tables/appendix_rationales_table.tex
\begin{table*}[t]
    \centering
    \small
    \begin{tabular}{ccrr}
    \toprule
  \textbf{Dataset} & Word$_{Len.}$  & Sent.$_{Len.}$ & Count \\
    \midrule
     All &  4.54 (0.69) & 21.05 (13.63) & 3198 \\ \midrule
     R1  &  \textbf{4.57} (0.65) & 22.4 (13.80) & 1000\\
     R2 &  4.51 (\textbf{0.71})  &20.14 (12.96) & 1000\\
     R3 &  4.55 (0.70) & \bf 20.81 (14.11) & 1198\\ \midrule
     C  &  4.53 (0.70) & 19.46 (12.64) & 1062 \\  
     N  &  4.52 (0.64) & \bf 23.81 (15.05) & 1066 \\
     E  &  \bf 4.58 (0.72) & 19.87 (12.66) & 1070  \\ \midrule
     Numerical & 4.44 (0.65) & 21.79 (13.21) & 1036\\
     Basic & \textbf{4.63} (0.69) & 21.31 (13.92) & 1327 \\
     Reference & 4.53 (0.70) & 20.04 (13.01) & 868\\
     Tricky & 4.56 (\textbf{0.71}) & 20.58 (13.22) & 893\\
     Reasoning & 4.52 (0.66) & \bf 21.82 (14.08) & 1197\\
     Imperfection & 4.53 (\textbf{0.71}) & 19.26 (13.06) & 452 \\
    \bottomrule
    \end{tabular}
    \caption{Average length of words and sentences in rationales for ANLI. Average and (standard deviation) are provided.}
    \label{tab:rationales}
\end{table*}

%% file: tables/appendix_incidence_tables.tex
\begin{table*}[t]
    \centering
    \scriptsize
    \begin{tabular}{ccrrrrrrrrrr}
    \toprule
    \bf & Round & Overall & Cardinal & Ordinal & Counting & Nominal & Dates & Age \\
    \midrule
    \multirow{4}{*}{\bf Numerical} & A1  &  40.8\% &  37.8\% &  6.2\% &  1.9\% &  4.2\% &  27.4\% &  5.9\% \\
& A2 &  38.5\% &  34.7\% &  6.7\% &  2.8\% &  3.5\% &  24.3\% &  6.7\% \\
& A3 &  20.3\% &  18.6\% &  2.8\% &  2.3\% &  0.4\% &  7.1\% &  3.2\% \\
& All &  32.4\% &  29.6\% &  5.1\% &  2.3\% &  2.6\% &  18.8\% &  5.1\% \\
    \toprule
    & Round & Overall & Lexical & Compr. Supr. & Implic. & Idioms & Negation & Coord. \\
    \midrule
    \multirow{4}{*}{\bf Basic} 
& A1 &  31.4\% &  16.0\% &  5.3\% &  1.5\% &  0.3\% & 5.6\% &  5.5\% \\
& A2 &  41.2\% &  20.2\% &  7.6\% &  2.4\% & 1.7\% & 9.8\% &  4.5\% \\
& A3 &  50.2\% &  26.4\% &  4.9\% &  4.2\% & 2.2\% & 15.8\% &  6.1\% \\
& All &  41.5\% &  21.2\% &  5.9\% &  2.8\% &  1.4\% &  10.7\% &  5.4\% \\
    \toprule
    &  Round & Overall & Coreference & Names & Family \\
    \midrule
    \multirow{4}{*}{\bf Ref. \& Names} 
& A1 &  24.5\% &  15.8\% &  12.5\%  & 1.0\% \\
& A2 &  29.4\% &  22.7\% &  11.2\%  & 1.7\% \\
& A3 &  27.5\% &  25.5\% &   1.9\%  & 1.3\% \\
& All &  27.1\% &  21.6\% &  8.1\% &  1.3\% \\
    \toprule
    & Round & Overall & Syntactic & Prag. & Exhaustif. & Wordplay \\
    \midrule
    \multirow{4}{*}{\bf Tricky} 
& A1 &  29.5\% &  14.5\% & 4.7\% &  5.5\% &  2.0\% \\
& A2 &  29.1\% &  8.0\% &  2.8\% & 8.6\% &  5.7\% \\
& A3 &  25.6\% &  9.3\% &  6.7\% &  4.8\% &  5.5\% \\
    & All &  27.9\% &  10.5\% &  4.8\% &  6.2\% &  4.5\% \\
    \toprule
    & Round & Overall & Likely & Unlikely & Debatable & Facts & Containment \\
    \midrule
    \multirow{4}{*}{\bf Reasoning} 
& A1 &  58.4\% &  25.7\% &  6.2\% &  3.1\% &  19.6\% &  11.0\% \\
& A2 &  62.7\% &  23.9\% &  6.9\% & 6.5\% &  25.6\% &  10.3\% \\
& A3 &  63.9\% &  22.7\% &  10.9\% &  10.8\% &  26.5\% &  5.3\% \\
& All &  61.8\% &  24.0\% &  8.2\% &  7.0\% &  24.0\% &  8.7\% \\
    \toprule
    & Round & Overall & Error & Ambiguous & EventCoref & Translation & Spelling  \\
    \midrule
    \multirow{4}{*}{\bf Imperfections} 
& A1 &  12.4\% &  3.3\% &  2.8\% &  0.9\% &  5.7\% &  5.8\% \\
& A2 &  13.5\% &  2.5\% &  4.0\% &  3.4\% &  6.2\% &  6.5\% \\
& A3 &  16.1\% &  2.2\% &  7.6\% &  1.9\% &  0.8\% &  5.5\% \\
& All &  14.1\% &  2.6\% &  5.0\% &  2.1\% &  4.0\% &  5.9\% \\
    \bottomrule
    \end{tabular}
    \caption{Analysis of development set. Percent examples with particular tag, per round, on average.
    }
    \label{tab:fullincidence}
\end{table*}

%% file: tables/appendix_heatmap_pvalues_table.tex
\begin{table*}[t]
    \centering
    \small
    \begin{adjustbox}{max width=\linewidth}
\begin{tabular}{lrrrrrrrrrrrrr}
\toprule
{} & gold label & BERT (R1) & RoBERTa (R2) & RoBERTa (R3) &  ALBERT &    BERT & distilBERT &   XLNet & XLNet-Large &     XLM & BART & RoBERTa-base & distilRoBERTa \\
\midrule
gold label    &           &        \textbf{0}&           \textbf{0}&           \textbf{0}&  \textbf{0.0127} &   0.085 &         \textbf{0}&      \textbf{0}&          \textbf{0}&      \textbf{0}&   \textbf{0}&       \textbf{0.0039} &   \bf     0.0001 \\
BERT (R1)     &         \textbf{0}&          &           \textbf{0}&           \textbf{0}&      \textbf{0}&      \textbf{0}&      0.095 &  0.8655 &   \bf   0.0031 &  0.9842 &   \textbf{0}&       \bf 0.0001 &   \bf     0.0003 \\
RoBERTa (R2)  &         \textbf{0}&        \textbf{0}&            &           \textbf{0}&   0.946 &  0.1298 &         \textbf{0}&      \textbf{0}&          \textbf{0}&      \textbf{0}&   \textbf{0}&           \textbf{0}&            \textbf{0}\\
RoBERTa (R3)  &         \textbf{0}&        \textbf{0}&           \textbf{0}&             &  0.4057 &  0.8928 &         \textbf{0}&      \textbf{0}&          \textbf{0}&      \textbf{0}&   \textbf{0}&           \textbf{0}&    \bf    0.0002 \\
ALBERT        &   \bf  0.0127 &        \textbf{0}&        0.946 &       0.4057 &        &      \textbf{0}&         \textbf{0}&      \textbf{0}&          \textbf{0}&      \textbf{0}&   \textbf{0}&       0.2016 &         0.416 \\
BERT          &      0.085 &        \textbf{0}&       0.1298 &       0.8928 &      \textbf{0}&        &         \textbf{0}&      \textbf{0}&          \textbf{0}&      \textbf{0}&   \textbf{0}&     \bf  0.0039 &        0.4133 \\
distilBERT    &         \textbf{0}&     0.095 &           \textbf{0}&           \textbf{0}&      \textbf{0}&      \textbf{0}&          &      \textbf{0}&          \textbf{0}&      \textbf{0}&   \textbf{0}&           \textbf{0}&            \textbf{0}\\
XLNet         &         \textbf{0}&    0.8655 &           \textbf{0}&           \textbf{0}&      \textbf{0}&      \textbf{0}&         \textbf{0}&        &          \textbf{0}&      \textbf{0}&   \textbf{0}&           \textbf{0}&            \textbf{0}\\
XLNet-Large   &         \textbf{0}&  \bf  0.0031 &           \textbf{0}&           \textbf{0}&      \textbf{0}&      \textbf{0}&         \textbf{0}&      \textbf{0}&            &      \textbf{0}&   \textbf{0}&           \textbf{0}&            \textbf{0}\\
XLM           &         \textbf{0}&    0.9842 &           \textbf{0}&           \textbf{0}&      \textbf{0}&      \textbf{0}&         \textbf{0}&      \textbf{0}&          \textbf{0}&       &   \textbf{0}&           \textbf{0}&            \textbf{0}\\
BART          &         \textbf{0}&        \textbf{0}&           \textbf{0}&           \textbf{0}&      \textbf{0}&      \textbf{0}&         \textbf{0}&      \textbf{0}&          \textbf{0}&      \textbf{0}&     &           \textbf{0}&            \textbf{0}\\
RoBERTa-base  &   \bf  0.0039 &   \bf 0.0001 &           \textbf{0}&           \textbf{0}&  0.2016 & \bf 0.0039 &         \textbf{0}&      \textbf{0}&          \textbf{0}&      \textbf{0}&   \textbf{0}&          &            \textbf{0}\\
distilRoBERTa &   \bf  0.0001 & \bf   0.0003 &           \textbf{0}&   \bf    0.0002 &   0.416 &  0.4133 &         \textbf{0}&      \textbf{0}&          \textbf{0}&      \textbf{0}&   \textbf{0}&           \textbf{0}&              \\
\bottomrule
\end{tabular}  
\end{adjustbox}
\caption{Pearson correlation p-values for heatmap. Any p-value $<0.05$ is significant (bold).
    }
    \label{tab:heatmapPearsonP}
\end{table*}

%% file: tables/appendix_model_preds_table_specifictags.tex
\begin{table*}[t]
    \centering
    \small
    \begin{adjustbox}{max width=\linewidth}
    \begin{tabular}{crccccccc}
    \toprule
    \bf Round & \bf Model & \bf Numerical & \bf Basic & \bf Ref. \& Names & \bf Tricky & \bf Reasoning & \bf Imperfections\\ \midrule
\multirow{12}{*}{\bf A1}  &                BERT (R1) &  0.10 (0.57) &  0.13 (0.60) &  0.11 (0.56) &  0.10 (0.56) &  0.12 (0.59) &  0.13 (0.57) \\
 &    RoBERTa Ensemble (R2) &  0.68 (0.13) &  0.67 (0.13) &  0.69 (0.15) &  0.6 (0.18) &  0.66 (0.15) &  0.61 (0.14) \\
 &    RoBERTa Ensemble (R3) &  0.72 (0.07) &  0.73 (0.08) &  0.72 (0.08) &  0.65 (0.09) &  0.7 (0.08) &  0.68 (0.07) \\
 &        BERT-base-uncased &  0.24 (0.92) &  0.39 (0.92) &  0.28 (0.88) &  0.26 (0.86) &  0.3 (0.92) &  0.3 (0.87) \\
 &              ALBERT-base &  0.23 (0.95) &  0.44 (0.98) &  0.3 (0.97) &  0.24 (0.95) &  0.27 (0.96) &  0.32 (0.95) \\
 &  distilBERT-base-uncased &  0.19 (0.35) &  0.21 (0.34) &  0.21 (0.31) &  0.22 (0.31) &  0.17 (0.34) &  0.24 (0.31) \\
 &             RoBERTa-base &  0.32 (0.40) &  0.47 (0.33) &  0.31 (0.34) &  0.34 (0.40) &  0.38 (0.34) &  0.37 (0.36) \\
 &       distilRoBERTa-base &  0.34 (0.39) &  0.42 (0.34) &  0.31 (0.31) &  0.37 (0.38) &  0.39 (0.36) &  0.4 (0.39) \\
 &         XLnet-base-cased &  0.17 (0.54) &  0.21 (0.48) &  0.21 (0.46) &  0.22 (0.54) &  0.14 (0.48) &  0.21 (0.52) \\
 &        XLnet-large-cased &  0.16 (0.52) &  0.18 (0.45) &  0.16 (0.47) &  0.19 (0.54) &  0.13 (0.52) &  0.17 (0.50) \\
 &                      XLM &  0.15 (0.61) &  0.19 (0.57) &  0.17 (0.59) &  0.19 (0.58) &  0.13 (0.56) &  0.2 (0.61) \\
 &               BART-Large &  0.13 (0.32) &  0.12 (0.25) &  0.12 (0.27) &  0.15 (0.30) &  0.10 (0.34) &  0.13 (0.32) \\ \midrule
 \multirow{12}{*}{\bf A2} &                BERT (R1) &  0.29 (0.53) &  0.3 (0.47) &  0.29 (0.44) &  0.25 (0.48) &  0.31 (0.47) &  0.33 (0.48) \\
 &    RoBERTa Ensemble (R2) &  0.19 (0.28) &  0.21 (0.26) &  0.20 (0.25) &  0.16 (0.23) &  0.19 (0.24) &  0.19 (0.27) \\
 &    RoBERTa Ensemble (R3) &  0.50 (0.18) &  0.43 (0.16) &  0.41 (0.14) &  0.44 (0.14) &  0.45 (0.14) &  0.33 (0.14) \\
 &        BERT-base-uncased &  0.25 (0.91) &  0.39 (0.88) &  0.30 (0.84) &  0.25 (0.86) &  0.31 (0.94) &  0.39 (0.91) \\
 &              ALBERT-base &  0.25 (0.98) &  0.41 (0.99) &  0.30 (0.99) &  0.28 (0.96) &  0.30 (1.00) &  0.35 (1.01) \\
 &  distilBERT-base-uncased &  0.22 (0.36) &  0.27 (0.33) &  0.24 (0.34) &  0.25 (0.34) &  0.23 (0.38) &  0.25 (0.33) \\
 &             RoBERTa-base &  0.39 (0.48) &  0.40 (0.41) &  0.35 (0.38) &  0.39 (0.41) &  0.36 (0.41) &  0.42 (0.38) \\
 &       distilRoBERTa-base &  0.42 (0.44) &  0.40 (0.38) &  0.36 (0.38) &  0.41 (0.37) &  0.39 (0.41) &  0.43 (0.34) \\
 &         XLnet-base-cased &  0.24 (0.63) &  0.27 (0.57) &  0.24 (0.55) &  0.26 (0.57) &  0.25 (0.58) &  0.27 (0.49) \\
 &        XLnet-large-cased &  0.22 (0.62) &  0.26 (0.58) &  0.22 (0.58) &  0.25 (0.59) &  0.22 (0.58) &  0.25 (0.57) \\
 &                      XLM &  0.23 (0.70) &  0.25 (0.64) &  0.24 (0.63) &  0.25 (0.64) &  0.22 (0.66) &  0.23 (0.62) \\
 &               BART-Large &  0.20 (0.43) &  0.23 (0.38) &  0.21 (0.39) &  0.24 (0.37) &  0.21 (0.39) &  0.28 (0.35) \\ \midrule
  \multirow{12}{*}{\bf A3} &                BERT (R1) &  0.34 (0.53) &  0.34 (0.51) &  0.32 (0.50) &  0.29 (0.55) &  0.32 (0.49) &  0.31 (0.54) \\
 &    RoBERTa Ensemble (R2) &  0.29 (0.47) &  0.26 (0.54) &  0.26 (0.57) &  0.24 (0.58) &  0.27 (0.55) &  0.23 (0.58) \\
 &    RoBERTa Ensemble (R3) &  0.20 (0.43) &  0.23 (0.50) &  0.24 (0.53) &  0.25 (0.54) &  0.25 (0.54) &  0.23 (0.52) \\
 &        BERT-base-uncased &  0.28 (0.80) &  0.42 (0.66) &  0.26 (0.64) &  0.21 (0.60) &  0.30 (0.65) &  0.37 (0.64) \\
 &              ALBERT-base &  0.29 (1.10) &  0.39 (1.08) &  0.27 (1.08) &  0.24 (1.02) &  0.30 (1.10) &  0.35 (1.09) \\
 &  distilBERT-base-uncased &  0.23 (0.41) &  0.25 (0.35) &  0.26 (0.36) &  0.24 (0.35) &  0.22 (0.34) &  0.22 (0.35) \\
 &             RoBERTa-base &  0.41 (0.48) &  0.36 (0.40) &  0.29 (0.38) &  0.29 (0.43) &  0.34 (0.43) &  0.34 (0.43) \\
 &       distilRoBERTa-base &  0.39 (0.42) &  0.33 (0.37) &  0.30 (0.37) &  0.33 (0.36) &  0.35 (0.37) &  0.32 (0.37) \\
 &         XLnet-base-cased &  0.22 (0.61) &  0.25 (0.53) &  0.24 (0.52) &  0.27 (0.55) &  0.21 (0.50) &  0.24 (0.57) \\
 &        XLnet-large-cased &  0.21 (0.60) &  0.23 (0.60) &  0.22 (0.59) &  0.23 (0.59) &  0.20 (0.60) &  0.25 (0.57) \\
 &                      XLM &  0.23 (0.73) &  0.25 (0.65) &  0.23 (0.66) &  0.23 (0.63) &  0.21 (0.66) &  0.25 (0.70) \\
 &               BART-Large &  0.20 (0.44) &  0.22 (0.36) &  0.21 (0.36) &  0.22 (0.41) &  0.19 (0.37) &  0.26 (0.36) \\ \midrule
   \multirow{12}{*}{\bf ANLI} &               BERT (R1) &  0.22 (0.54) &  0.26 (0.52) &  0.26 (0.50) &  0.21 (0.53) &  0.26 (0.51) &  0.27 (0.53) \\
 &    RoBERTa Ensemble (R2) &  0.41 (0.26) &  0.37 (0.33) &  0.34 (0.37) &  0.33 (0.34) &  0.35 (0.33) &  0.32 (0.37) \\
 &    RoBERTa Ensemble (R3) &  0.52 (0.20) &  0.44 (0.27) &  0.41 (0.30) &  0.45 (0.26) &  0.45 (0.28) &  0.39 (0.28) \\
 &        BERT-base-uncased &  0.25 (0.89) &  0.40 (0.80) &  0.28 (0.76) &  0.24 (0.77) &  0.31 (0.83) &  0.36 (0.78) \\
 &              ALBERT-base &  0.25 (1.00) &  0.41 (1.02) &  0.29 (1.03) &  0.25 (0.98) &  0.29 (1.03) &  0.34 (1.03) \\
 &  distilBERT-base-uncased &  0.21 (0.37) &  0.25 (0.34) &  0.24 (0.34) &  0.24 (0.33) &  0.21 (0.36) &  0.23 (0.33) \\
 &             RoBERTa-base &  0.37 (0.45) &  0.40 (0.39) &  0.31 (0.37) &  0.34 (0.41) &  0.36 (0.40) &  0.37 (0.39) \\
 &       distilRoBERTa-base &  0.38 (0.41) &  0.38 (0.36) &  0.32 (0.36) &  0.37 (0.37) &  0.37 (0.38) &  0.37 (0.37) \\
 &         XLnet-base-cased &  0.21 (0.59) &  0.25 (0.53) &  0.23 (0.52) &  0.25 (0.55) &  0.20 (0.52) &  0.24 (0.53) \\
 &        XLnet-large-cased &  0.20 (0.58) &  0.23 (0.55) &  0.20 (0.56) &  0.22 (0.57) &  0.19 (0.57) &  0.23 (0.55) \\
 &                      XLM &  0.20 (0.67) &  0.23 (0.62) &  0.22 (0.64) &  0.23 (0.62) &  0.19 (0.63) &  0.23 (0.65) \\
 &               BART-Large &  0.17 (0.39) &  0.19 (0.33) &  0.19 (0.35) &  0.20 (0.36) &  0.17 (0.37) &  0.23 (0.35) \\
    \bottomrule
    \end{tabular}
    \end{adjustbox}
    \caption{Correct label probability and entropy of label predictions for each model on each round's development set: mean probability (mean entropy). BERT (R1) has zero accuracy, by construction, on A1 because it was used to collect A1, whereas RoBERTas (R2) and (R3) were part of an ensemble of several identical architectures with different random seeds, so they have low, but non-zero, accuracy on their respective rounds. Recall that the entropy for three equiprobable outcomes (i.e., random chance of three NLI labels) is upper bounded by $\approx 1.58$.}
    \label{tab:modelpredstoplevel}
\end{table*}

\begin{table*}[t]
    \centering
    \small
    \begin{adjustbox}{max width=\linewidth}
\begin{tabular}{crcccccccc}
\toprule
\bf BASIC & & & & & & & & & \\
\bf Round & \bf Model & \bf Basic &  Lexical & Comp.Sup. &  ModusPonens &  CauseEffect &        Idiom &     Negation &  Coordination \\\midrule
\multirow{12}{*}{\bf A1} &                 BERT (R1) &  0.11 (0.56) &  0.12 (0.59) &            0.13 (0.66) &  0.07 (0.31) &  0.15 (0.55) &  0.01 (0.45) &  0.07 (0.40) &  0.10 (0.52) \\
&    RoBERTa Ensemble (R2) &  0.69 (0.15) &  0.73 (0.14) &            0.63 (0.24) &  0.43 (0.06) &  0.75 (0.02) &  0.35 (0.12) &  0.66 (0.17) &  0.67 (0.13) \\
&    RoBERTa Ensemble (R3) &  0.72 (0.08) &  0.78 (0.08) &            0.72 (0.15) &  0.32 (0.19) &  0.75 (0.01) &  0.67 (0.02) &  0.67 (0.06) &  0.65 (0.08) \\
&        BERT-base-uncased &  0.28 (0.88) &  0.27 (0.90) &            0.26 (0.94) &  0.07 (0.63) &  0.43 (0.74) &  0.04 (0.66) &  0.25 (0.76) &  0.32 (0.97) \\
&              ALBERT-base &  0.30 (0.97) &  0.31 (0.97) &            0.33 (0.98) &  0.07 (0.80) &  0.35 (0.96) &  0.14 (1.18) &  0.27 (1.06) &  0.26 (0.90) \\
&  distilBERT-base-uncased &  0.21 (0.31) &  0.23 (0.32) &            0.23 (0.26) &  0.15 (0.19) &  0.01 (0.09) &  0.30 (0.22) &  0.21 (0.37) &  0.18 (0.29) \\
&             RoBERTa-base &  0.31 (0.34) &  0.30 (0.33) &            0.35 (0.51) &  0.14 (0.39) &  0.49 (0.09) &  0.33 (0.08) &  0.27 (0.27) &  0.28 (0.30) \\
&       distilRoBERTa-base &  0.31 (0.31) &  0.29 (0.31) &            0.36 (0.36) &  0.15 (0.27) &  0.36 (0.32) &  0.33 (0.22) &  0.26 (0.36) &  0.30 (0.24) \\
&         XLnet-base-cased &  0.21 (0.46) &  0.22 (0.45) &            0.19 (0.58) &  0.12 (0.44) &  0.01 (0.19) &  0.01 (0.26) &  0.22 (0.50) &  0.21 (0.37) \\
&        XLnet-large-cased &  0.16 (0.47) &  0.16 (0.46) &            0.23 (0.51) &  0.07 (0.65) &  0.05 (0.69) &  0.01 (0.46) &  0.15 (0.48) &  0.14 (0.40) \\
&                      XLM &  0.17 (0.59) &  0.18 (0.61) &            0.21 (0.63) &  0.07 (0.55) &  0.12 (0.42) &  0.06 (0.90) &  0.17 (0.65) &  0.14 (0.47) \\
&               BART-Large &  0.12 (0.27) &  0.12 (0.29) &            0.14 (0.34) &  0.12 (0.23) &  0.03 (0.23) &  0.00 (0.08) &  0.12 (0.25) &  0.12 (0.24) \\ \midrule
\multirow{12}{*}{\bf A2} &                 BERT (R1) &  0.29 (0.44) &  0.31 (0.46) &            0.31 (0.56) &  0.24 (0.31) &  0.29 (0.40) &  0.35 (0.44) &  0.24 (0.41) &  0.20 (0.38) \\
&    RoBERTa Ensemble (R2) &  0.20 (0.25) &  0.24 (0.23) &            0.19 (0.33) &  0.33 (0.32) &  0.21 (0.35) &  0.19 (0.21) &  0.17 (0.26) &  0.15 (0.29) \\
&    RoBERTa Ensemble (R3) &  0.41 (0.14) &  0.43 (0.15) &            0.49 (0.16) &  0.55 (0.18) &  0.15 (0.17) &  0.28 (0.10) &  0.42 (0.09) &  0.41 (0.21) \\
&        BERT-base-uncased &  0.30 (0.84) &  0.26 (0.84) &            0.32 (0.86) &  0.21 (0.92) &  0.37 (0.90) &  0.33 (0.83) &  0.31 (0.76) &  0.36 (0.79) \\
&              ALBERT-base &  0.30 (0.99) &  0.29 (0.99) &            0.35 (1.06) &  0.30 (1.02) &  0.32 (0.97) &  0.28 (1.00) &  0.28 (1.00) &  0.29 (1.02) \\
&  distilBERT-base-uncased &  0.24 (0.34) &  0.20 (0.35) &            0.29 (0.38) &  0.03 (0.24) &  0.50 (0.23) &  0.29 (0.35) &  0.24 (0.36) &  0.20 (0.34) \\
&             RoBERTa-base &  0.35 (0.38) &  0.33 (0.36) &            0.37 (0.48) &  0.21 (0.20) &  0.14 (0.33) &  0.31 (0.33) &  0.41 (0.33) &  0.33 (0.43) \\
&       distilRoBERTa-base &  0.36 (0.38) &  0.36 (0.36) &            0.36 (0.48) &  0.20 (0.20) &  0.19 (0.21) &  0.28 (0.34) &  0.41 (0.42) &  0.39 (0.36) \\
&         XLnet-base-cased &  0.24 (0.55) &  0.24 (0.55) &            0.26 (0.64) &  0.18 (0.41) &  0.44 (0.35) &  0.31 (0.60) &  0.20 (0.52) &  0.18 (0.51) \\
&        XLnet-large-cased &  0.22 (0.58) &  0.19 (0.58) &            0.28 (0.63) &  0.20 (0.43) &  0.42 (0.43) &  0.23 (0.58) &  0.18 (0.54) &  0.22 (0.60) \\
&                      XLM &  0.24 (0.63) &  0.22 (0.64) &            0.23 (0.72) &  0.20 (0.41) &  0.41 (0.50) &  0.33 (0.74) &  0.21 (0.57) &  0.23 (0.72) \\
&               BART-Large &  0.21 (0.39) &  0.19 (0.38) &            0.26 (0.43) &  0.20 (0.13) &  0.47 (0.39) &  0.18 (0.45) &  0.19 (0.37) &  0.20 (0.34) \\ \midrule
\multirow{12}{*}{\bf A3} &                  BERT (R1) &  0.32 (0.50) &  0.33 (0.51) &            0.36 (0.59) &  0.29 (0.72) &  0.25 (0.57) &  0.22 (0.47) &  0.32 (0.46) &  0.34 (0.50) \\
&    RoBERTa Ensemble (R2) &  0.26 (0.57) &  0.26 (0.57) &            0.29 (0.55) &  0.25 (0.81) &  0.16 (0.58) &  0.24 (0.68) &  0.25 (0.62) &  0.26 (0.56) \\
&    RoBERTa Ensemble (R3) &  0.24 (0.53) &  0.23 (0.53) &            0.21 (0.53) &  0.24 (0.57) &  0.17 (0.51) &  0.19 (0.57) &  0.23 (0.57) &  0.28 (0.50) \\
&        BERT-base-uncased &  0.26 (0.64) &  0.22 (0.63) &            0.28 (0.73) &  0.19 (0.34) &  0.14 (0.55) &  0.28 (0.50) &  0.31 (0.63) &  0.22 (0.69) \\
&              ALBERT-base &  0.27 (1.08) &  0.25 (1.07) &            0.29 (1.07) &  0.22 (1.00) &  0.19 (0.99) &  0.28 (1.16) &  0.30 (1.14) &  0.23 (1.03) \\
&  distilBERT-base-uncased &  0.26 (0.36) &  0.27 (0.36) &            0.33 (0.32) &  0.30 (0.59) &  0.20 (0.43) &  0.22 (0.37) &  0.24 (0.38) &  0.21 (0.36) \\
&             RoBERTa-base &  0.29 (0.38) &  0.24 (0.38) &            0.44 (0.41) &  0.17 (0.45) &  0.14 (0.32) &  0.18 (0.52) &  0.36 (0.39) &  0.31 (0.43) \\
&       distilRoBERTa-base &  0.30 (0.37) &  0.26 (0.36) &            0.37 (0.41) &  0.36 (0.56) &  0.17 (0.41) &  0.30 (0.31) &  0.33 (0.41) &  0.36 (0.31) \\
&         XLnet-base-cased &  0.24 (0.52) &  0.23 (0.51) &            0.29 (0.55) &  0.25 (0.68) &  0.28 (0.55) &  0.21 (0.55) &  0.24 (0.57) &  0.22 (0.51) \\
&        XLnet-large-cased &  0.22 (0.59) &  0.21 (0.61) &            0.25 (0.52) &  0.18 (0.88) &  0.25 (0.52) &  0.22 (0.72) &  0.21 (0.59) &  0.16 (0.60) \\
&                      XLM &  0.23 (0.66) &  0.22 (0.66) &            0.27 (0.70) &  0.34 (0.94) &  0.23 (0.64) &  0.15 (0.57) &  0.21 (0.66) &  0.22 (0.67) \\
&               BART-Large &  0.21 (0.36) &  0.20 (0.37) &            0.30 (0.44) &  0.20 (0.18) &  0.19 (0.26) &  0.14 (0.37) &  0.19 (0.36) &  0.17 (0.40) \\ \midrule
\multirow{12}{*}{\bf ANLI} &               BERT (R1) &  0.26 (0.50) &  0.27 (0.51) &            0.27 (0.60) &  0.21 (0.50) &  0.25 (0.52) &  0.26 (0.46) &  0.26 (0.44) &  0.23 (0.48) \\
&    RoBERTa Ensemble (R2) &  0.34 (0.37) &  0.36 (0.37) &            0.35 (0.37) &  0.33 (0.46) &  0.25 (0.45) &  0.23 (0.47) &  0.29 (0.44) &  0.36 (0.36) \\
&    RoBERTa Ensemble (R3) &  0.41 (0.30) &  0.42 (0.31) &            0.46 (0.27) &  0.34 (0.36) &  0.23 (0.35) &  0.25 (0.36) &  0.36 (0.35) &  0.43 (0.29) \\
&        BERT-base-uncased &  0.28 (0.76) &  0.24 (0.76) &            0.29 (0.84) &  0.16 (0.56) &  0.24 (0.67) &  0.28 (0.63) &  0.30 (0.69) &  0.29 (0.80) \\
&              ALBERT-base &  0.29 (1.03) &  0.28 (1.02) &            0.33 (1.04) &  0.19 (0.94) &  0.24 (0.98) &  0.27 (1.10) &  0.29 (1.08) &  0.25 (0.99) \\
&  distilBERT-base-uncased &  0.24 (0.34) &  0.24 (0.35) &            0.29 (0.32) &  0.19 (0.38) &  0.26 (0.33) &  0.25 (0.35) &  0.24 (0.37) &  0.20 (0.33) \\
&             RoBERTa-base &  0.31 (0.37) &  0.28 (0.36) &            0.39 (0.46) &  0.17 (0.38) &  0.18 (0.30) &  0.23 (0.42) &  0.36 (0.35) &  0.30 (0.39) \\
&       distilRoBERTa-base &  0.32 (0.36) &  0.30 (0.35) &            0.36 (0.42) &  0.26 (0.38) &  0.20 (0.35) &  0.29 (0.31) &  0.34 (0.41) &  0.35 (0.30) \\
&         XLnet-base-cased &  0.23 (0.52) &  0.23 (0.51) &            0.25 (0.59) &  0.19 (0.54) &  0.29 (0.45) &  0.23 (0.55) &  0.23 (0.54) &  0.20 (0.46) \\
&        XLnet-large-cased &  0.20 (0.56) &  0.19 (0.56) &            0.25 (0.56) &  0.15 (0.70) &  0.27 (0.52) &  0.21 (0.65) &  0.19 (0.56) &  0.17 (0.54) \\
&                      XLM &  0.22 (0.64) &  0.21 (0.64) &            0.24 (0.69) &  0.22 (0.69) &  0.27 (0.58) &  0.21 (0.65) &  0.20 (0.63) &  0.20 (0.62) \\
&               BART-Large &  0.19 (0.35) &  0.18 (0.35) &            0.24 (0.41) &  0.17 (0.19) &  0.25 (0.29) &  0.15 (0.38) &  0.18 (0.35) &  0.16 (0.33) \\
\bottomrule
    \end{tabular}
    \end{adjustbox}
    \caption{Correct label probability and entropy of label predictions for the \textsc{Basic} subset: mean probability (mean entropy). BERT (R1) has zero accuracy, by construction, on A1 because it was used to collect A1, whereas RoBERTas (R2) and (R3) were part of an ensemble of several identical architectures with different random seeds, so they have low, but non-zero, accuracy on their respective rounds. Recall that the entropy for three equiprobable outcomes (i.e., random chance of three NLI labels) is upper bounded by $\approx 1.58$.}
    \label{tab:modelpredsbasic}
\end{table*}

\begin{table*}[t]
    \centering
    \small
    \begin{adjustbox}{max width=\linewidth}
\begin{tabular}{crccccccc}
\toprule
\bf NUMERICAL & & & & & & & &  \\
\bf Round &                    Model &    Numerical &     Cardinal &      Ordinal &     Counting &      Nominal &        Dates &          Age \\
\midrule
\multirow{12}{*}{\bf A1 } &                BERT (R1) &  0.10 (0.57) &  0.10 (0.57) &  0.11 (0.60) &  0.09 (0.64) &  0.07 (0.46) &  0.10 (0.58) &  0.07 (0.41) \\
&    RoBERTa Ensemble (R2) &  0.68 (0.13) &  0.68 (0.13) &  0.71 (0.18) &  0.51 (0.23) &  0.72 (0.11) &  0.69 (0.13) &  0.64 (0.11) \\
&    RoBERTa Ensemble (R3) &  0.72 (0.07) &  0.72 (0.07) &  0.77 (0.05) &  0.51 (0.23) &  0.69 (0.06) &  0.75 (0.07) &  0.64 (0.08) \\
&        BERT-base-uncased &  0.24 (0.92) &  0.24 (0.93) &  0.26 (0.84) &  0.21 (0.95) &  0.30 (0.67) &  0.24 (0.94) &  0.24 (1.12) \\
&              ALBERT-base &  0.23 (0.95) &  0.22 (0.96) &  0.24 (0.96) &  0.15 (0.97) &  0.32 (0.91) &  0.21 (0.97) &  0.15 (0.99) \\
&  distilBERT-base-uncased &  0.19 (0.35) &  0.19 (0.36) &  0.21 (0.30) &  0.23 (0.32) &  0.24 (0.32) &  0.16 (0.34) &  0.23 (0.44) \\
&             RoBERTa-base &  0.32 (0.40) &  0.34 (0.40) &  0.24 (0.40) &  0.36 (0.40) &  0.42 (0.29) &  0.33 (0.41) &  0.35 (0.58) \\
&       distilRoBERTa-base &  0.34 (0.39) &  0.35 (0.39) &  0.31 (0.36) &  0.30 (0.39) &  0.40 (0.35) &  0.36 (0.37) &  0.38 (0.51) \\
&         XLnet-base-cased &  0.17 (0.54) &  0.16 (0.54) &  0.17 (0.56) &  0.15 (0.63) &  0.22 (0.44) &  0.15 (0.54) &  0.21 (0.77) \\
&        XLnet-large-cased &  0.16 (0.52) &  0.16 (0.52) &  0.20 (0.48) &  0.20 (0.49) &  0.19 (0.49) &  0.14 (0.52) &  0.22 (0.81) \\
&                      XLM &  0.15 (0.61) &  0.15 (0.60) &  0.18 (0.67) &  0.15 (0.64) &  0.18 (0.59) &  0.14 (0.59) &  0.15 (0.77) \\
&               BART-Large &  0.13 (0.32) &  0.13 (0.32) &  0.12 (0.32) &  0.24 (0.45) &  0.11 (0.25) &  0.12 (0.30) &  0.13 (0.59) \\ \midrule
\multirow{12}{*}{\bf A2 } &                BERT (R1) &  0.29 (0.53) &  0.28 (0.53) &  0.33 (0.53) &  0.43 (0.49) &  0.31 (0.53) &  0.25 (0.53) &  0.18 (0.48) \\
&    RoBERTa Ensemble (R2) &  0.19 (0.28) &  0.20 (0.28) &  0.19 (0.24) &  0.14 (0.30) &  0.20 (0.34) &  0.19 (0.26) &  0.22 (0.25) \\
&    RoBERTa Ensemble (R3) &  0.50 (0.18) &  0.51 (0.18) &  0.50 (0.13) &  0.36 (0.20) &  0.44 (0.19) &  0.55 (0.17) &  0.51 (0.15) \\
&        BERT-base-uncased &  0.25 (0.91) &  0.25 (0.92) &  0.28 (0.92) &  0.36 (0.92) &  0.19 (0.83) &  0.22 (0.92) &  0.22 (1.02) \\
&              ALBERT-base &  0.25 (0.98) &  0.24 (0.98) &  0.30 (0.97) &  0.29 (1.03) &  0.22 (0.93) &  0.22 (0.97) &  0.18 (1.01) \\
&  distilBERT-base-uncased &  0.22 (0.36) &  0.22 (0.36) &  0.24 (0.33) &  0.17 (0.49) &  0.30 (0.38) &  0.21 (0.35) &  0.13 (0.39) \\
&             RoBERTa-base &  0.39 (0.48) &  0.39 (0.48) &  0.34 (0.42) &  0.28 (0.64) &  0.42 (0.48) &  0.37 (0.48) &  0.38 (0.61) \\
&       distilRoBERTa-base &  0.42 (0.44) &  0.42 (0.44) &  0.39 (0.35) &  0.33 (0.53) &  0.42 (0.50) &  0.41 (0.42) &  0.43 (0.57) \\
&         XLnet-base-cased &  0.24 (0.63) &  0.23 (0.63) &  0.33 (0.62) &  0.26 (0.59) &  0.27 (0.64) &  0.21 (0.62) &  0.16 (0.66) \\
&        XLnet-large-cased &  0.22 (0.62) &  0.22 (0.62) &  0.26 (0.60) &  0.17 (0.59) &  0.25 (0.64) &  0.19 (0.63) &  0.18 (0.73) \\
&                      XLM &  0.23 (0.70) &  0.22 (0.71) &  0.25 (0.62) &  0.20 (0.69) &  0.26 (0.69) &  0.21 (0.72) &  0.18 (0.85) \\
&               BART-Large &  0.20 (0.43) &  0.19 (0.44) &  0.28 (0.40) &  0.21 (0.42) &  0.16 (0.44) &  0.17 (0.45) &  0.18 (0.53) \\ \midrule
\multirow{12}{*}{\bf A3 } &                BERT (R1) &  0.34 (0.53) &  0.34 (0.53) &  0.43 (0.49) &  0.34 (0.34) &  0.41 (0.48) &  0.31 (0.48) &  0.28 (0.45) \\
&    RoBERTa Ensemble (R2) &  0.29 (0.47) &  0.29 (0.46) &  0.25 (0.47) &  0.17 (0.48) &  0.35 (0.41) &  0.30 (0.34) &  0.32 (0.36) \\
&    RoBERTa Ensemble (R3) &  0.20 (0.43) &  0.20 (0.42) &  0.25 (0.52) &  0.11 (0.37) &  0.20 (0.77) &  0.22 (0.30) &  0.26 (0.44) \\
&        BERT-base-uncased &  0.28 (0.80) &  0.28 (0.80) &  0.30 (0.80) &  0.20 (0.60) &  0.30 (0.95) &  0.34 (0.88) &  0.32 (0.83) \\
&              ALBERT-base &  0.29 (1.10) &  0.29 (1.10) &  0.32 (1.07) &  0.23 (0.96) &  0.31 (1.23) &  0.31 (1.10) &  0.27 (1.14) \\
&  distilBERT-base-uncased &  0.23 (0.41) &  0.22 (0.41) &  0.23 (0.32) &  0.22 (0.45) &  0.17 (0.28) &  0.24 (0.42) &  0.17 (0.38) \\
&             RoBERTa-base &  0.41 (0.48) &  0.43 (0.48) &  0.28 (0.42) &  0.41 (0.44) &  0.37 (0.24) &  0.48 (0.50) &  0.53 (0.48) \\
&       distilRoBERTa-base &  0.39 (0.42) &  0.40 (0.40) &  0.38 (0.53) &  0.55 (0.29) &  0.23 (0.36) &  0.44 (0.45) &  0.42 (0.59) \\
&         XLnet-base-cased &  0.22 (0.61) &  0.22 (0.60) &  0.19 (0.64) &  0.22 (0.61) &  0.17 (0.56) &  0.23 (0.71) &  0.21 (0.63) \\
&        XLnet-large-cased &  0.21 (0.60) &  0.22 (0.61) &  0.20 (0.55) &  0.25 (0.43) &  0.14 (0.68) &  0.22 (0.63) &  0.23 (0.67) \\
&                      XLM &  0.23 (0.73) &  0.23 (0.74) &  0.24 (0.64) &  0.23 (0.61) &  0.19 (0.70) &  0.23 (0.76) &  0.23 (0.84) \\
&               BART-Large &  0.20 (0.44) &  0.20 (0.44) &  0.14 (0.43) &  0.28 (0.49) &  0.18 (0.40) &  0.20 (0.49) &  0.21 (0.48) \\ \midrule
\multirow{12}{*}{\bf A3 } &                BERT (R1) &  0.22 (0.54) &  0.22 (0.55) &  0.27 (0.54) &  0.31 (0.48) &  0.19 (0.49) &  0.19 (0.54) &  0.16 (0.45) \\
&    RoBERTa Ensemble (R2) &  0.41 (0.26) &  0.41 (0.26) &  0.40 (0.26) &  0.25 (0.35) &  0.48 (0.22) &  0.44 (0.21) &  0.39 (0.23) \\
&    RoBERTa Ensemble (R3) &  0.52 (0.20) &  0.52 (0.19) &  0.55 (0.18) &  0.30 (0.27) &  0.56 (0.16) &  0.59 (0.14) &  0.50 (0.19) \\
&        BERT-base-uncased &  0.25 (0.89) &  0.25 (0.89) &  0.28 (0.86) &  0.26 (0.81) &  0.25 (0.75) &  0.25 (0.92) &  0.25 (1.01) \\
&              ALBERT-base &  0.25 (1.00) &  0.25 (1.00) &  0.28 (0.99) &  0.23 (0.99) &  0.28 (0.94) &  0.23 (0.98) &  0.19 (1.03) \\
&  distilBERT-base-uncased &  0.21 (0.37) &  0.21 (0.37) &  0.23 (0.32) &  0.20 (0.43) &  0.26 (0.34) &  0.19 (0.35) &  0.18 (0.41) \\
&             RoBERTa-base &  0.37 (0.45) &  0.38 (0.45) &  0.29 (0.41) &  0.35 (0.51) &  0.42 (0.37) &  0.36 (0.45) &  0.40 (0.57) \\
&       distilRoBERTa-base &  0.38 (0.41) &  0.39 (0.41) &  0.36 (0.39) &  0.40 (0.41) &  0.40 (0.41) &  0.39 (0.40) &  0.41 (0.55) \\
&         XLnet-base-cased &  0.21 (0.59) &  0.20 (0.59) &  0.24 (0.60) &  0.22 (0.61) &  0.24 (0.54) &  0.19 (0.60) &  0.19 (0.69) \\
&        XLnet-large-cased &  0.20 (0.58) &  0.19 (0.58) &  0.22 (0.55) &  0.21 (0.51) &  0.21 (0.57) &  0.17 (0.58) &  0.21 (0.75) \\
&                      XLM &  0.20 (0.67) &  0.20 (0.67) &  0.22 (0.64) &  0.20 (0.65) &  0.22 (0.64) &  0.18 (0.67) &  0.18 (0.82) \\
&               BART-Large &  0.17 (0.39) &  0.17 (0.39) &  0.19 (0.38) &  0.24 (0.45) &  0.14 (0.34) &  0.15 (0.39) &  0.17 (0.54) \\
\bottomrule
    \end{tabular}
    \end{adjustbox}
    \caption{Correct label probability and entropy of label predictions for the \textsc{Numerical} subset: mean probability (mean entropy). BERT (R1) has zero accuracy, by construction, on A1 because it was used to collect A1, whereas RoBERTas (R2) and (R3) were part of an ensemble of several identical architectures with different random seeds, so they have low, but non-zero, accuracy on their respective rounds. Recall that the entropy for three equiprobable outcomes (i.e., random chance of three NLI labels) is upper bounded by $\approx 1.58$.}
    \label{tab:modelpredsnumerical}
\end{table*}

\begin{table*}[t]
    \centering
    \small
    \begin{adjustbox}{max width=\linewidth}
\begin{tabular}{crcccccc}
\toprule
\bf REASONING & & & & & & &  \\
\bf Round &  \bf Model &  \bf Reasoning &       Likely &     Unlikely &    Debatable &        Facts &  Containment \\
\midrule
\multirow{12}{*}{\bf A1 } &                BERT (R1) &  0.13 (0.60) &  0.14 (0.57) &  0.15 (0.54) &  0.16 (0.52) &  0.11 (0.64) &  0.11 (0.62) \\
&    RoBERTa Ensemble (R2) &  0.67 (0.13) &  0.64 (0.16) &  0.78 (0.13) &  0.61 (0.05) &  0.65 (0.12) &  0.71 (0.14) \\
&    RoBERTa Ensemble (R3) &  0.73 (0.08) &  0.72 (0.09) &  0.78 (0.04) &  0.68 (0.00) &  0.71 (0.08) &  0.75 (0.11) \\
&        BERT-base-uncased &  0.39 (0.92) &  0.56 (0.95) &  0.52 (0.96) &  0.35 (0.78) &  0.25 (0.89) &  0.19 (0.91) \\
&              ALBERT-base &  0.44 (0.98) &  0.67 (1.02) &  0.59 (1.04) &  0.33 (1.00) &  0.20 (0.92) &  0.22 (0.93) \\
&  distilBERT-base-uncased &  0.21 (0.34) &  0.23 (0.29) &  0.13 (0.21) &  0.21 (0.23) &  0.24 (0.39) &  0.18 (0.47) \\
&             RoBERTa-base &  0.47 (0.33) &  0.61 (0.32) &  0.71 (0.21) &  0.34 (0.23) &  0.28 (0.40) &  0.34 (0.34) \\
&       distilRoBERTa-base &  0.42 (0.34) &  0.52 (0.32) &  0.62 (0.25) &  0.22 (0.38) &  0.26 (0.37) &  0.33 (0.33) \\
&         XLnet-base-cased &  0.21 (0.48) &  0.25 (0.47) &  0.09 (0.33) &  0.20 (0.40) &  0.24 (0.56) &  0.19 (0.48) \\
&        XLnet-large-cased &  0.18 (0.45) &  0.22 (0.43) &  0.12 (0.35) &  0.19 (0.41) &  0.17 (0.55) &  0.18 (0.43) \\
&                      XLM &  0.19 (0.57) &  0.23 (0.54) &  0.08 (0.55) &  0.18 (0.49) &  0.18 (0.63) &  0.17 (0.58) \\
&               BART-Large &  0.12 (0.25) &  0.13 (0.25) &  0.04 (0.19) &  0.05 (0.27) &  0.13 (0.28) &  0.13 (0.24) \\ \midrule
\multirow{12}{*}{\bf A2 } &                BERT (R1) &  0.30 (0.47) &  0.34 (0.44) &  0.31 (0.42) &  0.36 (0.44) &  0.23 (0.49) &  0.33 (0.54) \\
&    RoBERTa Ensemble (R2) &  0.21 (0.26) &  0.27 (0.28) &  0.21 (0.33) &  0.16 (0.27) &  0.18 (0.22) &  0.17 (0.19) \\
&    RoBERTa Ensemble (R3) &  0.43 (0.16) &  0.43 (0.14) &  0.45 (0.18) &  0.43 (0.16) &  0.40 (0.13) &  0.38 (0.17) \\
&        BERT-base-uncased &  0.39 (0.88) &  0.58 (0.89) &  0.54 (0.93) &  0.44 (0.92) &  0.23 (0.86) &  0.20 (0.89) \\
&              ALBERT-base &  0.41 (0.99) &  0.66 (1.02) &  0.57 (1.03) &  0.41 (0.96) &  0.20 (0.96) &  0.19 (0.95) \\
&  distilBERT-base-uncased &  0.27 (0.33) &  0.31 (0.33) &  0.23 (0.28) &  0.27 (0.32) &  0.27 (0.34) &  0.25 (0.42) \\
&             RoBERTa-base &  0.40 (0.41) &  0.49 (0.46) &  0.47 (0.35) &  0.38 (0.40) &  0.30 (0.39) &  0.39 (0.41) \\
&       distilRoBERTa-base &  0.40 (0.38) &  0.45 (0.40) &  0.49 (0.36) &  0.41 (0.35) &  0.33 (0.37) &  0.39 (0.36) \\
&         XLnet-base-cased &  0.27 (0.57) &  0.32 (0.62) &  0.29 (0.44) &  0.29 (0.54) &  0.27 (0.54) &  0.21 (0.59) \\
&        XLnet-large-cased &  0.26 (0.58) &  0.29 (0.60) &  0.27 (0.46) &  0.26 (0.53) &  0.26 (0.58) &  0.25 (0.57) \\
&                      XLM &  0.25 (0.64) &  0.28 (0.67) &  0.28 (0.54) &  0.22 (0.60) &  0.25 (0.64) &  0.25 (0.64) \\
&               BART-Large &  0.23 (0.38) &  0.22 (0.36) &  0.27 (0.28) &  0.25 (0.35) &  0.25 (0.40) &  0.24 (0.42) \\ \midrule
\multirow{12}{*}{\bf A3 } &                BERT (R1) &  0.34 (0.51) &  0.37 (0.47) &  0.38 (0.48) &  0.35 (0.51) &  0.29 (0.54) &  0.35 (0.46) \\
&    RoBERTa Ensemble (R2) &  0.26 (0.54) &  0.25 (0.51) &  0.28 (0.58) &  0.25 (0.62) &  0.25 (0.51) &  0.28 (0.38) \\
&    RoBERTa Ensemble (R3) &  0.23 (0.50) &  0.23 (0.47) &  0.25 (0.52) &  0.21 (0.56) &  0.22 (0.48) &  0.20 (0.38) \\
&        BERT-base-uncased &  0.42 (0.66) &  0.59 (0.65) &  0.56 (0.76) &  0.55 (0.59) &  0.21 (0.64) &  0.28 (0.74) \\
&              ALBERT-base &  0.39 (1.08) &  0.53 (1.07) &  0.51 (1.14) &  0.47 (1.08) &  0.23 (1.06) &  0.27 (1.03) \\
&  distilBERT-base-uncased &  0.25 (0.35) &  0.26 (0.34) &  0.25 (0.29) &  0.23 (0.30) &  0.26 (0.36) &  0.25 (0.45) \\
&             RoBERTa-base &  0.36 (0.40) &  0.38 (0.43) &  0.51 (0.35) &  0.40 (0.41) &  0.27 (0.38) &  0.33 (0.42) \\
&       distilRoBERTa-base &  0.33 (0.37) &  0.35 (0.35) &  0.46 (0.34) &  0.38 (0.31) &  0.26 (0.38) &  0.30 (0.33) \\
&         XLnet-base-cased &  0.25 (0.53) &  0.24 (0.52) &  0.24 (0.46) &  0.26 (0.57) &  0.28 (0.53) &  0.27 (0.58) \\
&        XLnet-large-cased &  0.23 (0.60) &  0.24 (0.62) &  0.24 (0.57) &  0.26 (0.60) &  0.25 (0.61) &  0.23 (0.53) \\
&                      XLM &  0.25 (0.65) &  0.27 (0.68) &  0.23 (0.55) &  0.26 (0.67) &  0.27 (0.66) &  0.25 (0.66) \\
&               BART-Large &  0.22 (0.36) &  0.20 (0.34) &  0.21 (0.32) &  0.25 (0.36) &  0.26 (0.36) &  0.25 (0.39) \\ \midrule
\multirow{12}{*}{\bf A3 } &                  BERT (R1) &  0.26 (0.52) &  0.29 (0.49) &  0.31 (0.48) &  0.33 (0.49) &  0.23 (0.55) &  0.25 (0.56) \\
&    RoBERTa Ensemble (R2) &  0.37 (0.33) &  0.39 (0.32) &  0.38 (0.41) &  0.28 (0.44) &  0.33 (0.32) &  0.41 (0.21) \\
&    RoBERTa Ensemble (R3) &  0.44 (0.27) &  0.46 (0.24) &  0.43 (0.32) &  0.34 (0.37) &  0.41 (0.26) &  0.48 (0.19) \\
&        BERT-base-uncased &  0.40 (0.80) &  0.58 (0.82) &  0.54 (0.85) &  0.49 (0.71) &  0.23 (0.78) &  0.21 (0.86) \\
&              ALBERT-base &  0.41 (1.02) &  0.62 (1.04) &  0.54 (1.09) &  0.43 (1.04) &  0.21 (0.99) &  0.22 (0.96) \\
&  distilBERT-base-uncased &  0.25 (0.34) &  0.27 (0.32) &  0.22 (0.27) &  0.24 (0.30) &  0.26 (0.36) &  0.22 (0.45) \\
&             RoBERTa-base &  0.40 (0.39) &  0.49 (0.40) &  0.55 (0.32) &  0.39 (0.38) &  0.28 (0.39) &  0.36 (0.38) \\
&       distilRoBERTa-base &  0.38 (0.36) &  0.44 (0.36) &  0.50 (0.32) &  0.37 (0.33) &  0.28 (0.38) &  0.34 (0.34) \\
&         XLnet-base-cased &  0.25 (0.53) &  0.27 (0.53) &  0.22 (0.43) &  0.26 (0.54) &  0.27 (0.54) &  0.22 (0.55) \\
&        XLnet-large-cased &  0.23 (0.55) &  0.25 (0.55) &  0.22 (0.49) &  0.25 (0.56) &  0.23 (0.58) &  0.22 (0.51) \\
&                      XLM &  0.23 (0.62) &  0.26 (0.63) &  0.21 (0.55) &  0.23 (0.62) &  0.24 (0.65) &  0.22 (0.62) \\
&               BART-Large &  0.19 (0.33) &  0.18 (0.32) &  0.19 (0.28) &  0.23 (0.34) &  0.22 (0.36) &  0.20 (0.34) \\
\bottomrule
    \end{tabular}
    \end{adjustbox}
    \caption{Correct label probability and entropy of label predictions for the \textsc{Reasoning} subset: mean probability (mean entropy). BERT (R1) has zero accuracy, by construction, on A1 because it was used to collect A1, whereas RoBERTas (R2) and (R3) were part of an ensemble of several identical architectures with different random seeds, so they have low, but non-zero, accuracy on their respective rounds. Recall that the entropy for three equiprobable outcomes (i.e., random chance of three NLI labels) is upper bounded by $\approx 1.58$.}
    \label{tab:modelpredsreasoning}
\end{table*}

\begin{table*}[t]
    \centering
    \small
    \begin{adjustbox}{max width=\linewidth}
\begin{tabular}{crcccccc}
\toprule
\bf REFERENCE & & & & &  \\
\bf Round &  \bf Model & \bf Reference &  Coreference &        Names &    Family \\ 
\midrule
\multirow{12}{*}{\bf A1 } &                BERT (R1) &  0.12 (0.59) &  0.11 (0.56) &  0.12 (0.60) &  0.12 (0.56) \\
&    RoBERTa Ensemble (R2) &  0.66 (0.15) &  0.67 (0.15) &  0.68 (0.15) &  0.29 (0.19) \\
&    RoBERTa Ensemble (R3) &  0.70 (0.08) &  0.70 (0.08) &  0.75 (0.06) &  0.44 (0.17) \\
&        BERT-base-uncased &  0.30 (0.92) &  0.30 (0.91) &  0.30 (0.93) &  0.33 (1.01) \\
&              ALBERT-base &  0.27 (0.96) &  0.27 (0.95) &  0.26 (0.97) &  0.21 (1.03) \\
&  distilBERT-base-uncased &  0.17 (0.34) &  0.16 (0.33) &  0.14 (0.33) &  0.42 (0.26) \\
&             RoBERTa-base &  0.38 (0.34) &  0.37 (0.32) &  0.36 (0.34) &  0.39 (0.50) \\
&       distilRoBERTa-base &  0.39 (0.36) &  0.38 (0.35) &  0.38 (0.33) &  0.39 (0.57) \\
&         XLnet-base-cased &  0.14 (0.48) &  0.11 (0.45) &  0.14 (0.49) &  0.41 (0.70) \\
&        XLnet-large-cased &  0.13 (0.52) &  0.10 (0.49) &  0.13 (0.53) &  0.35 (0.64) \\
&                      XLM &  0.13 (0.56) &  0.10 (0.54) &  0.14 (0.56) &  0.35 (0.82) \\
&               BART-Large &  0.10 (0.34) &  0.09 (0.34) &  0.10 (0.34) &  0.46 (0.27) \\ \midrule
\multirow{12}{*}{\bf A2 } &                BERT (R1) &  0.31 (0.47) &  0.29 (0.47) &  0.33 (0.48) &  0.34 (0.41) \\
&    RoBERTa Ensemble (R2) &  0.19 (0.24) &  0.20 (0.24) &  0.16 (0.24) &  0.18 (0.24) \\
&    RoBERTa Ensemble (R3) &  0.45 (0.14) &  0.46 (0.16) &  0.42 (0.14) &  0.45 (0.17) \\
&        BERT-base-uncased &  0.31 (0.94) &  0.32 (0.95) &  0.30 (0.98) &  0.29 (0.93) \\
&              ALBERT-base &  0.30 (1.00) &  0.30 (1.02) &  0.30 (0.97) &  0.22 (1.07) \\
&  distilBERT-base-uncased &  0.23 (0.38) &  0.23 (0.40) &  0.26 (0.41) &  0.22 (0.36) \\
&             RoBERTa-base &  0.36 (0.41) &  0.35 (0.43) &  0.39 (0.41) &  0.32 (0.44) \\
&       distilRoBERTa-base &  0.39 (0.41) &  0.37 (0.44) &  0.45 (0.36) &  0.41 (0.55) \\
&         XLnet-base-cased &  0.25 (0.58) &  0.26 (0.59) &  0.23 (0.60) &  0.21 (0.58) \\
&        XLnet-large-cased &  0.22 (0.58) &  0.22 (0.61) &  0.22 (0.55) &  0.28 (0.66) \\
&                      XLM &  0.22 (0.66) &  0.23 (0.69) &  0.20 (0.63) &  0.17 (0.69) \\
&               BART-Large &  0.21 (0.39) &  0.22 (0.40) &  0.18 (0.39) &  0.23 (0.52) \\ \midrule
\multirow{12}{*}{\bf A3 } &                BERT (R1) &  0.32 (0.49) &  0.33 (0.48) &  0.27 (0.51) &  0.25 (0.59) \\
&    RoBERTa Ensemble (R2) &  0.27 (0.55) &  0.27 (0.53) &  0.26 (0.76) &  0.39 (0.39) \\
&    RoBERTa Ensemble (R3) &  0.25 (0.54) &  0.24 (0.54) &  0.26 (0.46) &  0.47 (0.41) \\
&        BERT-base-uncased &  0.30 (0.65) &  0.30 (0.65) &  0.27 (0.65) &  0.35 (0.71) \\
&              ALBERT-base &  0.30 (1.10) &  0.30 (1.10) &  0.24 (1.10) &  0.33 (1.12) \\
&  distilBERT-base-uncased &  0.22 (0.34) &  0.23 (0.35) &  0.22 (0.37) &  0.32 (0.41) \\
&             RoBERTa-base &  0.34 (0.43) &  0.34 (0.44) &  0.37 (0.46) &  0.41 (0.39) \\
&       distilRoBERTa-base &  0.35 (0.37) &  0.36 (0.37) &  0.40 (0.39) &  0.39 (0.50) \\
&         XLnet-base-cased &  0.21 (0.50) &  0.22 (0.50) &  0.16 (0.54) &  0.30 (0.52) \\
&        XLnet-large-cased &  0.20 (0.60) &  0.21 (0.61) &  0.14 (0.53) &  0.19 (0.56) \\
&                      XLM &  0.21 (0.66) &  0.21 (0.66) &  0.18 (0.70) &  0.27 (0.71) \\
&               BART-Large &  0.19 (0.37) &  0.20 (0.37) &  0.17 (0.34) &  0.12 (0.38) \\ \midrule
\multirow{12}{*}{\bf ANLI } &                BERT (R1) &  0.26 (0.51) &  0.27 (0.49) &  0.22 (0.54) &  0.25 (0.51) \\
&    RoBERTa Ensemble (R2) &  0.35 (0.33) &  0.34 (0.35) &  0.42 (0.24) &  0.29 (0.28) \\
&    RoBERTa Ensemble (R3) &  0.45 (0.28) &  0.42 (0.31) &  0.56 (0.13) &  0.46 (0.26) \\
&        BERT-base-uncased &  0.31 (0.83) &  0.30 (0.81) &  0.30 (0.93) &  0.32 (0.87) \\
&              ALBERT-base &  0.29 (1.03) &  0.29 (1.04) &  0.27 (0.98) &  0.26 (1.08) \\
&  distilBERT-base-uncased &  0.21 (0.36) &  0.21 (0.36) &  0.20 (0.37) &  0.30 (0.35) \\
&             RoBERTa-base &  0.36 (0.40) &  0.35 (0.41) &  0.37 (0.38) &  0.37 (0.43) \\
&       distilRoBERTa-base &  0.37 (0.38) &  0.37 (0.39) &  0.41 (0.35) &  0.40 (0.54) \\
&         XLnet-base-cased &  0.20 (0.52) &  0.21 (0.52) &  0.18 (0.54) &  0.29 (0.58) \\
&        XLnet-large-cased &  0.19 (0.57) &  0.19 (0.58) &  0.17 (0.54) &  0.26 (0.62) \\
&                      XLM &  0.19 (0.63) &  0.19 (0.64) &  0.17 (0.60) &  0.25 (0.73) \\
&               BART-Large &  0.17 (0.37) &  0.18 (0.37) &  0.14 (0.36) &  0.25 (0.41) \\\bottomrule
    \end{tabular}
    \end{adjustbox}
    \caption{Correct label probability and entropy of label predictions for the \textsc{Reference} subset: mean probability (mean entropy). BERT (R1) has zero accuracy, by construction, on A1 because it was used to collect A1, whereas RoBERTas (R2) and (R3) were part of an ensemble of several identical architectures with different random seeds, so they have low, but non-zero, accuracy on their respective rounds. Recall that the entropy for three equiprobable outcomes (i.e., random chance of three NLI labels) is upper bounded by $\approx 1.58$.}
    \label{tab:modelpredsreference}
\end{table*}

\begin{table*}[t]
    \centering
    \small
    \begin{adjustbox}{max width=\linewidth}
\begin{tabular}{crcccccc}
\toprule
\bf TRICKY & & & & & & \\
 \bf Round & \bf Model &  \bf   Tricky &    Syntactic & Pragmatic & Exhaustification &     Wordplay \\
\midrule
\multirow{12}{*}{\bf A1 } &                BERT (R1) &  0.10 (0.56) &  0.10 (0.54) &  0.09 (0.56) &      0.11 (0.56) &  0.13 (0.72) \\
 &    RoBERTa Ensemble (R2) &  0.60 (0.18) &  0.60 (0.17) &  0.60 (0.23) &      0.59 (0.17) &  0.52 (0.15) \\
 &    RoBERTa Ensemble (R3) &  0.65 (0.09) &  0.67 (0.09) &  0.72 (0.08) &      0.54 (0.11) &  0.51 (0.06) \\
 &        BERT-base-uncased &  0.26 (0.86) &  0.27 (0.82) &  0.22 (0.82) &      0.25 (0.86) &  0.16 (0.82) \\
 &              ALBERT-base &  0.24 (0.95) &  0.24 (0.90) &  0.19 (0.92) &      0.23 (1.01) &  0.18 (0.85) \\
 &  distilBERT-base-uncased &  0.22 (0.31) &  0.17 (0.28) &  0.27 (0.29) &      0.39 (0.30) &  0.18 (0.29) \\
 &             RoBERTa-base &  0.34 (0.40) &  0.37 (0.40) &  0.23 (0.34) &      0.33 (0.44) &  0.47 (0.38) \\
 &       distilRoBERTa-base &  0.37 (0.38) &  0.43 (0.35) &  0.28 (0.42) &      0.36 (0.48) &  0.44 (0.25) \\
 &         XLnet-base-cased &  0.22 (0.54) &  0.19 (0.53) &  0.24 (0.51) &      0.39 (0.62) &  0.12 (0.57) \\
 &        XLnet-large-cased &  0.19 (0.54) &  0.16 (0.55) &  0.20 (0.47) &      0.31 (0.56) &  0.11 (0.49) \\
 &                      XLM &  0.19 (0.58) &  0.19 (0.54) &  0.20 (0.59) &      0.27 (0.77) &  0.12 (0.59) \\
 &               BART-Large &  0.15 (0.30) &  0.12 (0.31) &  0.18 (0.21) &      0.25 (0.29) &  0.10 (0.23) \\ \midrule
\multirow{12}{*}{\bf A2 } &                BERT (R1) &  0.25 (0.48) &  0.22 (0.53) &  0.20 (0.35) &      0.29 (0.47) &  0.21 (0.47) \\
 &    RoBERTa Ensemble (R2) &  0.16 (0.23) &  0.19 (0.25) &  0.10 (0.13) &      0.20 (0.21) &  0.09 (0.30) \\
 &    RoBERTa Ensemble (R3) &  0.44 (0.14) &  0.40 (0.13) &  0.33 (0.10) &      0.37 (0.16) &  0.59 (0.14) \\
 &        BERT-base-uncased &  0.25 (0.86) &  0.24 (0.85) &  0.22 (0.79) &      0.24 (0.85) &  0.18 (0.89) \\
 &              ALBERT-base &  0.28 (0.96) &  0.26 (0.93) &  0.25 (0.93) &      0.31 (1.06) &  0.18 (0.91) \\
 &  distilBERT-base-uncased &  0.25 (0.34) &  0.25 (0.32) &  0.20 (0.40) &      0.29 (0.47) &  0.13 (0.22) \\
 &             RoBERTa-base &  0.39 (0.41) &  0.45 (0.42) &  0.34 (0.43) &      0.31 (0.44) &  0.40 (0.36) \\
 &       distilRoBERTa-base &  0.41 (0.37) &  0.45 (0.38) &  0.32 (0.49) &      0.33 (0.39) &  0.43 (0.26) \\
 &         XLnet-base-cased &  0.26 (0.57) &  0.26 (0.53) &  0.23 (0.51) &      0.29 (0.59) &  0.16 (0.58) \\
 &        XLnet-large-cased &  0.25 (0.59) &  0.30 (0.60) &  0.25 (0.64) &      0.31 (0.64) &  0.16 (0.57) \\
 &                      XLM &  0.25 (0.64) &  0.26 (0.63) &  0.22 (0.64) &      0.37 (0.67) &  0.15 (0.59) \\
 &               BART-Large &  0.24 (0.37) &  0.25 (0.42) &  0.34 (0.44) &      0.35 (0.38) &  0.10 (0.29) \\ \midrule
\multirow{12}{*}{\bf A3 } &                BERT (R1) &  0.29 (0.55) &  0.29 (0.50) &  0.29 (0.64) &      0.28 (0.48) &  0.25 (0.58) \\
 &    RoBERTa Ensemble (R2) &  0.24 (0.58) &  0.26 (0.51) &  0.24 (0.62) &      0.18 (0.53) &  0.24 (0.72) \\
 &    RoBERTa Ensemble (R3) &  0.25 (0.54) &  0.29 (0.53) &  0.20 (0.57) &      0.23 (0.58) &  0.24 (0.50) \\
 &        BERT-base-uncased &  0.21 (0.60) &  0.24 (0.63) &  0.19 (0.58) &      0.22 (0.63) &  0.16 (0.54) \\
 &              ALBERT-base &  0.24 (1.02) &  0.26 (0.99) &  0.24 (1.02) &      0.25 (1.03) &  0.18 (1.03) \\
 &  distilBERT-base-uncased &  0.24 (0.35) &  0.27 (0.40) &  0.32 (0.42) &      0.23 (0.31) &  0.11 (0.22) \\
 &             RoBERTa-base &  0.29 (0.43) &  0.32 (0.46) &  0.24 (0.46) &      0.21 (0.41) &  0.34 (0.40) \\
 &       distilRoBERTa-base &  0.33 (0.36) &  0.39 (0.42) &  0.26 (0.38) &      0.32 (0.44) &  0.34 (0.21) \\
 &         XLnet-base-cased &  0.27 (0.55) &  0.27 (0.51) &  0.33 (0.66) &      0.33 (0.53) &  0.13 (0.50) \\
 &        XLnet-large-cased &  0.23 (0.59) &  0.23 (0.65) &  0.29 (0.57) &      0.33 (0.61) &  0.09 (0.50) \\
 &                      XLM &  0.23 (0.63) &  0.25 (0.68) &  0.27 (0.63) &      0.30 (0.68) &  0.12 (0.57) \\
 &               BART-Large &  0.22 (0.41) &  0.19 (0.37) &  0.31 (0.44) &      0.31 (0.42) &  0.10 (0.44) \\  \midrule
\multirow{12}{*}{\bf ANLI } &                BERT (R1) &  0.21 (0.53) &  0.19 (0.52) &  0.22 (0.56) &      0.24 (0.50) &  0.22 (0.55) \\
 &    RoBERTa Ensemble (R2) &  0.33 (0.34) &  0.39 (0.30) &  0.32 (0.41) &      0.30 (0.29) &  0.22 (0.47) \\
 &    RoBERTa Ensemble (R3) &  0.45 (0.26) &  0.48 (0.24) &  0.38 (0.34) &      0.38 (0.27) &  0.42 (0.29) \\
 &        BERT-base-uncased &  0.24 (0.77) &  0.25 (0.76) &  0.21 (0.69) &      0.24 (0.79) &  0.17 (0.72) \\
 &              ALBERT-base &  0.25 (0.98) &  0.25 (0.94) &  0.23 (0.98) &      0.27 (1.04) &  0.18 (0.96) \\
 &  distilBERT-base-uncased &  0.24 (0.33) &  0.22 (0.33) &  0.28 (0.38) &      0.30 (0.38) &  0.13 (0.23) \\
 &             RoBERTa-base &  0.34 (0.41) &  0.37 (0.43) &  0.26 (0.42) &      0.29 (0.43) &  0.38 (0.38) \\
 &       distilRoBERTa-base &  0.37 (0.37) &  0.42 (0.38) &  0.28 (0.41) &      0.34 (0.43) &  0.39 (0.23) \\
 &         XLnet-base-cased &  0.25 (0.55) &  0.23 (0.52) &  0.28 (0.59) &      0.33 (0.58) &  0.14 (0.54) \\
 &        XLnet-large-cased &  0.22 (0.57) &  0.22 (0.60) &  0.26 (0.55) &      0.32 (0.61) &  0.12 (0.53) \\
 &                      XLM &  0.23 (0.62) &  0.22 (0.60) &  0.24 (0.62) &      0.32 (0.70) &  0.13 (0.58) \\
 &               BART-Large &  0.20 (0.36) &  0.17 (0.36) &  0.27 (0.37) &      0.31 (0.37) &  0.10 (0.35) \\ \bottomrule
    \end{tabular}
    \end{adjustbox}
    \caption{Correct label probability and entropy of label predictions for the \textsc{Tricky} subset: mean probability (mean entropy). BERT (R1) has zero accuracy, by construction, on A1 because it was used to collect A1, whereas RoBERTas (R2) and (R3) were part of an ensemble of several identical architectures with different random seeds, so they have low, but non-zero, accuracy on their respective rounds. Recall that the entropy for three equiprobable outcomes (i.e., random chance of three NLI labels) is upper bounded by $\approx 1.58$.}
    \label{tab:modelpredstricky}
\end{table*}

\begin{table*}[t]
    \centering
    \small
    \begin{adjustbox}{max width=\linewidth}
\begin{tabular}{crcccccc}
\toprule
\bf IMPERFECTIONS & & & & & & \\ \midrule
\bf Round &         \bf            Model &      \bf Imperfections &      \bf   Errors & \bf    Ambiguity & \bf   EventCoref & \bf  Translation &   \bf   Spelling \\
  \multirow{12}{*}{\bf A1 } &                BERT (R1) &  0.13 (0.57) &  0.07 (0.38) &  0.17 (0.73) &  0.12 (0.77) &  0.11 (0.59) &  0.14 (0.64) \\
&    RoBERTa Ensemble (R2) &  0.61 (0.14) &  0.38 (0.11) &  0.53 (0.19) &  0.82 (0.25) &  0.67 (0.17) &  0.77 (0.12) \\
&    RoBERTa Ensemble (R3) &  0.68 (0.07) &  0.49 (0.12) &  0.57 (0.02) &  0.89 (0.00) &  0.71 (0.06) &  0.81 (0.07) \\
&        BERT-base-uncased &  0.30 (0.87) &  0.26 (0.85) &  0.33 (0.87) &  0.30 (1.01) &  0.29 (0.96) &  0.29 (0.88) \\
&              ALBERT-base &  0.32 (0.95) &  0.32 (1.06) &  0.41 (0.88) &  0.45 (0.97) &  0.27 (1.03) &  0.27 (0.88) \\
&  distilBERT-base-uncased &  0.24 (0.31) &  0.35 (0.32) &  0.27 (0.36) &  0.24 (0.26) &  0.17 (0.40) &  0.17 (0.33) \\
&             RoBERTa-base &  0.37 (0.36) &  0.26 (0.37) &  0.34 (0.53) &  0.55 (0.16) &  0.29 (0.35) &  0.44 (0.32) \\
&       distilRoBERTa-base &  0.40 (0.39) &  0.27 (0.47) &  0.41 (0.46) &  0.55 (0.25) &  0.27 (0.32) &  0.47 (0.40) \\
&         XLnet-base-cased &  0.21 (0.52) &  0.27 (0.52) &  0.31 (0.67) &  0.15 (0.55) &  0.22 (0.49) &  0.14 (0.48) \\
&        XLnet-large-cased &  0.17 (0.50) &  0.22 (0.43) &  0.26 (0.51) &  0.13 (0.25) &  0.20 (0.57) &  0.09 (0.54) \\
&                      XLM &  0.20 (0.61) &  0.24 (0.62) &  0.30 (0.54) &  0.27 (0.52) &  0.19 (0.51) &  0.12 (0.63) \\
&               BART-Large &  0.13 (0.32) &  0.18 (0.25) &  0.14 (0.45) &  0.11 (0.20) &  0.12 (0.30) &  0.11 (0.39) \\ \midrule
  \multirow{12}{*}{\bf A2 } &                 BERT (R1) &  0.33 (0.48) &  0.42 (0.39) &  0.32 (0.47) &  0.27 (0.43) &  0.29 (0.51) &  0.34 (0.45) \\
&    RoBERTa Ensemble (R2) &  0.19 (0.27) &  0.22 (0.22) &  0.19 (0.23) &  0.21 (0.33) &  0.16 (0.23) &  0.21 (0.28) \\
&    RoBERTa Ensemble (R3) &  0.33 (0.14) &  0.34 (0.17) &  0.43 (0.11) &  0.40 (0.11) &  0.46 (0.13) &  0.32 (0.12) \\
&        BERT-base-uncased &  0.39 (0.91) &  0.47 (1.07) &  0.42 (0.87) &  0.32 (0.91) &  0.35 (0.96) &  0.33 (0.86) \\
&              ALBERT-base &  0.35 (1.01) &  0.44 (1.00) &  0.39 (1.04) &  0.51 (0.94) &  0.31 (0.98) &  0.29 (1.01) \\
&  distilBERT-base-uncased &  0.25 (0.33) &  0.33 (0.27) &  0.23 (0.37) &  0.09 (0.30) &  0.26 (0.38) &  0.22 (0.30) \\
&             RoBERTa-base &  0.42 (0.38) &  0.43 (0.34) &  0.41 (0.47) &  0.38 (0.36) &  0.40 (0.37) &  0.43 (0.34) \\
&       distilRoBERTa-base &  0.43 (0.34) &  0.50 (0.28) &  0.41 (0.29) &  0.42 (0.19) &  0.50 (0.34) &  0.43 (0.33) \\
&         XLnet-base-cased &  0.27 (0.49) &  0.32 (0.47) &  0.29 (0.46) &  0.17 (0.46) &  0.29 (0.60) &  0.25 (0.49) \\
&        XLnet-large-cased &  0.25 (0.57) &  0.31 (0.61) &  0.25 (0.54) &  0.23 (0.65) &  0.18 (0.49) &  0.21 (0.55) \\
&                      XLM &  0.23 (0.62) &  0.23 (0.61) &  0.22 (0.66) &  0.24 (0.68) &  0.20 (0.64) &  0.19 (0.62) \\
&               BART-Large &  0.28 (0.35) &  0.28 (0.41) &  0.36 (0.38) &  0.14 (0.36) &  0.25 (0.32) &  0.21 (0.32) \\ \midrule
  \multirow{12}{*}{\bf A3 } &                 BERT (R1) &  0.31 (0.54) &  0.30 (0.57) &  0.28 (0.58) &  0.24 (0.29) &  0.42 (0.76) &  0.36 (0.52) \\
&    RoBERTa Ensemble (R2) &  0.23 (0.58) &  0.22 (0.65) &  0.23 (0.58) &  0.36 (0.52) &  0.26 (0.21) &  0.19 (0.46) \\
&    RoBERTa Ensemble (R3) &  0.23 (0.52) &  0.23 (0.55) &  0.17 (0.52) &  0.32 (0.48) &  0.16 (0.26) &  0.22 (0.46) \\
&        BERT-base-uncased &  0.37 (0.64) &  0.41 (0.58) &  0.38 (0.61) &  0.43 (0.54) &  0.19 (0.67) &  0.33 (0.66) \\
&              ALBERT-base &  0.35 (1.09) &  0.40 (1.08) &  0.37 (1.10) &  0.38 (1.15) &  0.23 (0.97) &  0.32 (1.02) \\
&  distilBERT-base-uncased &  0.22 (0.35) &  0.27 (0.39) &  0.21 (0.36) &  0.13 (0.35) &  0.30 (0.23) &  0.24 (0.40) \\
&             RoBERTa-base &  0.34 (0.43) &  0.19 (0.28) &  0.35 (0.51) &  0.29 (0.36) &  0.30 (0.46) &  0.30 (0.39) \\
&       distilRoBERTa-base &  0.32 (0.37) &  0.30 (0.32) &  0.33 (0.44) &  0.35 (0.30) &  0.21 (0.50) &  0.27 (0.32) \\
&         XLnet-base-cased &  0.24 (0.57) &  0.31 (0.52) &  0.27 (0.65) &  0.17 (0.51) &  0.40 (0.40) &  0.26 (0.48) \\
&        XLnet-large-cased &  0.25 (0.57) &  0.34 (0.43) &  0.26 (0.63) &  0.23 (0.58) &  0.36 (0.60) &  0.25 (0.58) \\
&                      XLM &  0.25 (0.70) &  0.36 (0.73) &  0.26 (0.77) &  0.19 (0.48) &  0.38 (0.54) &  0.26 (0.67) \\
&               BART-Large &  0.26 (0.36) &  0.38 (0.28) &  0.25 (0.38) &  0.16 (0.44) &  0.29 (0.45) &  0.26 (0.32) \\ \midrule
  \multirow{12}{*}{\bf ANLI } &                 BERT (R1) &  0.27 (0.53) &  0.24 (0.44) &  0.27 (0.58) &  0.24 (0.43) &  0.22 (0.57) &  0.28 (0.53) \\
&    RoBERTa Ensemble (R2) &  0.32 (0.37) &  0.28 (0.31) &  0.27 (0.42) &  0.35 (0.39) &  0.39 (0.20) &  0.38 (0.29) \\
&    RoBERTa Ensemble (R3) &  0.39 (0.28) &  0.36 (0.27) &  0.31 (0.33) &  0.44 (0.22) &  0.55 (0.11) &  0.44 (0.22) \\
&        BERT-base-uncased &  0.36 (0.78) &  0.37 (0.83) &  0.38 (0.72) &  0.35 (0.79) &  0.31 (0.93) &  0.32 (0.80) \\
&              ALBERT-base &  0.34 (1.03) &  0.38 (1.05) &  0.38 (1.05) &  0.46 (1.02) &  0.29 (1.00) &  0.30 (0.98) \\
&  distilBERT-base-uncased &  0.23 (0.33) &  0.32 (0.33) &  0.23 (0.36) &  0.12 (0.31) &  0.23 (0.38) &  0.21 (0.35) \\
&             RoBERTa-base &  0.37 (0.39) &  0.29 (0.33) &  0.36 (0.51) &  0.37 (0.33) &  0.34 (0.37) &  0.39 (0.35) \\
&       distilRoBERTa-base &  0.37 (0.37) &  0.35 (0.37) &  0.36 (0.41) &  0.42 (0.24) &  0.38 (0.34) &  0.39 (0.35) \\
&         XLnet-base-cased &  0.24 (0.53) &  0.30 (0.51) &  0.28 (0.60) &  0.17 (0.49) &  0.27 (0.53) &  0.22 (0.48) \\
&        XLnet-large-cased &  0.23 (0.55) &  0.28 (0.48) &  0.26 (0.58) &  0.22 (0.57) &  0.21 (0.54) &  0.19 (0.56) \\
&                      XLM &  0.23 (0.65) &  0.27 (0.65) &  0.26 (0.70) &  0.23 (0.59) &  0.21 (0.58) &  0.19 (0.64) \\
&               BART-Large &  0.23 (0.35) &  0.27 (0.31) &  0.26 (0.39) &  0.15 (0.37) &  0.20 (0.32) &  0.20 (0.34) \\ \bottomrule
    \end{tabular}
    \end{adjustbox}
    \caption{Correct label probability and entropy of label predictions for the \textsc{Imperfections} subset: mean probability (mean entropy). BERT (R1) has zero accuracy, by construction, on A1 because it was used to collect A1, whereas RoBERTas (R2) and (R3) were part of an ensemble of several identical architectures with different random seeds, so they have low, but non-zero, accuracy on their respective rounds. Recall that the entropy for three equiprobable outcomes (i.e., random chance of three NLI labels) is upper bounded by $\approx 1.58$. A3 had no examples of \textsc{Translation}, so no numbers can be reported.}
    \label{tab:modelpredsimperfect}
\end{table*}

%% file: tables/appendix_model_genre_full.tex
\begin{table*}[t]
    \centering
    \small
    \begin{adjustbox}{max width=\linewidth}
\begin{tabular}{lrrrrrrr}
\toprule
\bf Genre &     \bf               Model &  \bf  Numerical &   \bf     Basic & \bf   Reference &    \bf   Tricky & \bf   Reasoning &   \bf   Imperfections \\
\midrule
\multirow{12}{*}{\bf Wikipedia} &     BERT-Large (R1) &  0.20 (0.55) &  0.23 (0.49) &  0.24 (0.51) &  0.18 (0.52) &  0.23 (0.53) &  0.24 (0.52) \\
&  RoBERTa-Large (R2) &  0.43 (0.21) &  0.40 (0.21) &  0.40 (0.21) &  0.37 (0.22) &  0.42 (0.21) &  0.37 (0.21) \\
&  RoBERTa-Large (R3) &  \bf 0.58 (0.13) & \bf 0.51 (0.12) & \bf 0.54 (0.12) & \bf 0.52 (0.12) & \bf 0.53 (0.13) & \bf 0.46 (0.12) \\
&           BERT-Base &  0.26 (0.91) &  0.29 (0.85) &  0.31 (0.93) &  0.25 (0.86) &  0.40 (0.89) &  0.35 (0.88) \\
&         ALBERT-Base &  0.25 (0.97) &  0.30 (0.98) &  0.28 (0.98) &  0.25 (0.95) &  0.42 (0.98) &  0.34 (0.98) \\
&     distilBERT-Base &  0.22 (0.36) &  0.23 (0.33) &  0.21 (0.36) &  0.23 (0.33) &  0.25 (0.34) &  0.25 (0.34) \\
&        RoBERTa-Base &  0.36 (0.45) &  0.33 (0.36) &  0.36 (0.39) &  0.36 (0.40) &  0.42 (0.38) &  0.39 (0.39) \\
&  distilRoBERTa-Base &  0.38 (0.41) &  0.34 (0.35) &  0.39 (0.38) &  0.40 (0.37) &  0.40 (0.36) &  0.40 (0.38) \\
&          XLnet-Base &  0.21 (0.59) &  0.23 (0.52) &  0.20 (0.53) &  0.25 (0.56) &  0.25 (0.53) &  0.25 (0.53) \\
&         XLnet-Large &  0.20 (0.58) &  0.20 (0.54) &  0.19 (0.57) &  0.22 (0.57) &  0.23 (0.53) &  0.23 (0.56) \\
&                 XLM &  0.20 (0.67) &  0.22 (0.62) &  0.19 (0.63) &  0.23 (0.61) &  0.23 (0.61) &  0.23 (0.64) \\
&          BART-Large &  0.17 (0.38) &  0.18 (0.35) &  0.17 (0.37) &  0.20 (0.34) &  0.18 (0.33) &  0.22 (0.33) \\
 \midrule
 
 \multirow{12}{*}{ \bf Fiction}&     BERT-Large (R1) &  \textbf{0.49} (0.35) &  \textbf{0.28} (0.54) &  0.29 (0.52) &  \textbf{0.35} (0.60) &  0.29 (0.51) &  0.30 (0.62) \\
&  RoBERTa-Large (R2) &  0.32 (0.73) &  0.25 (0.68) &  0.26 (0.70) &  0.24 (0.71) &  0.26 (0.63) &  0.24 (0.73) \\
&  RoBERTa-Large (R3) &  0.35 (0.55) &  0.26 (0.70) &  0.29 (0.73) &  0.26 (0.72) &  0.27 (0.64) &  0.28 (0.73) \\
&           BERT-Base &  0.11 (0.46) &  0.17 (0.38) &  0.28 (0.39) &  0.21 (0.45) &  \textbf{0.44} (0.40) &  0.25 (0.40) \\
&         ALBERT-Base &  0.25 (1.03) &  0.22 (1.02) &  0.31 (1.04) &  0.24 (1.00) &  0.39 (1.04) &  0.29 (1.08) \\
&     distilBERT-Base &  0.02 (\textbf{0.20}) &  0.30 (0.41) &  0.19 (0.43) &  0.22 (0.39) &  0.23 (0.45) &  0.25 (0.38) \\
&        RoBERTa-Base &  0.43 (0.22) &  0.24 (\textbf{0.36}) &  0.30 (0.38) &  0.24 (0.40) &  0.33 (0.39) &  0.24 (\textbf{0.34}) \\
&  distilRoBERTa-Base &  0.24 (0.41) &  0.26 (0.47) &  \textbf{0.36} (0.48) &  0.24 (\textbf{0.41}) &  0.33 (0.42) &  0.24 (0.37) \\
&          XLnet-Base &  0.07 (0.52) &  \textbf{0.28} (0.59) &  0.23 (0.48) &  0.27 (0.58) &  0.28 (0.53) &  0.31 (0.64) \\
&         XLnet-Large &  0.30 (0.59) &  0.25 (0.54) &  0.18 (0.55) &  0.24 (0.54) &  0.25 (0.58) &  0.29 (0.54) \\
&                 XLM &  0.32 (0.70) &  0.27 (0.70) &  0.19 (0.62) &  0.25 (0.55) &  0.29 (0.66) &  0.31 (0.69) \\
&          BART-Large &  0.39 (0.44) &  0.23 (0.38) &  0.16 (\textbf{0.34}) &  0.22 (0.39) &  0.24 (\textbf{0.36}) &  \textbf{0.38} (0.36) \\
 \midrule
 
 \multirow{12}{*}{\bf News} &     BERT-Large (R1) &  0.38 (0.47) &  \textbf{0.32} (0.53) &  \textbf{0.26} (0.48) &  0.25 (0.61) &  0.40 (0.49) &  \textbf{0.39} (0.46) \\
&  RoBERTa-Large (R2) &  0.23 (0.40) &  0.24 (0.43) &  0.16 (\textbf{0.32}) &  0.23 (0.49) &  0.26 (0.41) &  0.14 (0.64) \\
&  RoBERTa-Large (R3) &  0.19 (\textbf{0.30}) &  0.22 (0.37) &  0.21 (0.34) &  0.26 (0.40) &  0.22 (0.39) &  0.23 (0.41) \\
&           BERT-Base &  0.18 (0.68) &  0.26 (0.59) &  0.26 (0.52) &  0.17 (0.55) &  \textbf{0.46} (0.59) &  0.28 (0.71) \\
&         ALBERT-Base &  0.21 (1.01) &  0.26 (1.01) &  \textbf{0.26} (1.03) &  0.23 (1.00) &  0.43 (1.05) &  0.32 (1.07) \\
&     distilBERT-Base &  0.16 (0.38) &  0.27 (0.34) &  0.25 (0.28) &  0.24 (\textbf{0.26}) &  0.27 (0.29) &  0.15 (\textbf{0.21}) \\
&        RoBERTa-Base &  0.44 (0.51) &  0.24 (\textbf{0.32}) &  \textbf{0.26} (0.42) &  0.24 (0.46) &  0.37 (0.36) &  0.24 (0.40) \\
&  distilRoBERTa-Base &  \textbf{0.45} (0.42) &  0.27 (0.35) &  0.24 (0.29) &  0.22 (0.31) &  0.35 (\textbf{0.31}) &  0.17 (0.23) \\
&          XLnet-Base &  0.12 (0.59) &  0.25 (0.43) &  0.22 (0.45) &  \textbf{0.34} (0.54) &  0.28 (0.49) &  0.15 (0.51) \\
&         XLnet-Large &  0.13 (0.56) &  0.21 (0.52) &  0.18 (0.57) &  0.24 (0.55) &  0.23 (0.59) &  0.13 (0.52) \\
&                 XLM &  0.20 (0.76) &  0.26 (0.61) &  0.20 (0.58) &  0.23 (0.58) &  0.28 (0.66) &  0.17 (0.56) \\
&          BART-Large &  0.10 (0.49) &  0.23 (0.37) &  0.20 (0.33) &  0.21 (0.32) &  0.25 (0.38) &  0.20 (0.46) \\
 \midrule
 
 \multirow{12}{*}{\bf Procedural}   &     BERT-Large (R1) &  0.37 (0.43) &  \textbf{0.30} (0.57) &  \textbf{0.38} (0.48) &  0.19 (0.46) &  0.34 (0.56) &  0.30 (0.58) \\
&  RoBERTa-Large (R2) &  0.28 (0.65) &  0.24 (0.67) &  0.22 (0.69) &  0.21 (0.70) &  0.26 (0.70) &  0.23 (0.60) \\
&  RoBERTa-Large (R3) &  0.21 (0.63) &  0.24 (0.59) &  0.21 (0.68) &  0.27 (0.64) &  0.25 (0.63) &  0.25 (0.51) \\
&           BERT-Base &  0.22 (0.51) &  0.29 (0.42) &  0.35 (0.47) &  0.20 (\textbf{0.38}) &  \textbf{0.46} (0.46) &  \textbf{0.63} (0.51) \\
&         ALBERT-Base &  0.27 (0.97) &  0.28 (0.95) &  0.36 (0.96) &  0.23 (0.89) &  0.44 (0.96) &  0.56 (0.96) \\
&     distilBERT-Base &  0.21 (0.35) &  0.26 (0.37) &  0.20 (0.34) &  \textbf{0.30} (0.42) &  0.27 (\textbf{0.28}) &  0.22 (0.37) \\
&        RoBERTa-Base &  0.31 (0.60) &  0.27 (0.43) &  0.45 (0.48) &  0.20 (0.45) &  0.32 (0.41) &  0.29 (0.49) \\
&  distilRoBERTa-Base &  \bf 0.42 (0.33) &  \textbf{0.30} (0.38) &  0.37 (\textbf{0.26}) &  0.22 (0.32) &  0.32 (0.34) &  0.32 (\textbf{0.35}) \\
&          XLnet-Base &  0.27 (0.67) &  0.24 (0.51) &  0.17 (0.45) &  0.24 (0.46) &  0.25 (0.49) &  0.21 (0.51) \\
&         XLnet-Large &  0.22 (0.53) &  0.21 (0.57) &  0.18 (0.51) &  0.26 (0.56) &  0.26 (0.56) &  0.18 (0.54) \\
&                 XLM &  0.21 (0.65) &  0.24 (0.59) &  0.18 (0.60) &  0.23 (0.64) &  0.26 (0.56) &  0.19 (0.70) \\
&          BART-Large &  0.21 (0.38) &  0.17 (\textbf{0.33}) &  0.10 (0.40) &  0.22 (0.45) &  0.22 (0.34) &  0.17 (0.39) \\
\bottomrule
\end{tabular}
\end{adjustbox}
\caption{Probability of the correct label (entropy of label predictions) for each model on each top level annotation tag. BERT (R1) has zero accuracy, by construction, on A1 because it was used to collect A1, whereas RoBERTas (R2) and (R3) were part of an ensemble of several identical architectures with different random seeds, so they have low, but non-zero, accuracy on their respective rounds. Recall that the entropy for three equiprobable outcomes (i.e., random chance of three NLI labels) is upper bounded by $\approx 1.58$.}
    \label{tab:modelpredsgenreall}
\end{table*}